\documentclass[examplefnt,biber,plain]{nowfnt} 
\def\part{Section}
\def\parts{Sections}
\def\rv{\color{black}}
\date{}

\usepackage[counterclockwise]{rotating}

\usepackage[utf8]{inputenc}

\def\GD{\texttt{GD}}
\def\SGD{\texttt{SGD}}
\def\mSGD{\texttt{mb-SGD}}
\def\ASGD{\texttt{ASGD}}
\def\DSGD{\texttt{DSGD}}
\def\CSGD{\texttt{CSGD}}
\def\DS{\texttt{EC-SGD}}

\usepackage{amsthm}
\usepackage{amsmath}
\usepackage{amssymb}
\usepackage{amsfonts}
\usepackage{color}
\usepackage{fullpage}
\usepackage{enumitem}
\usepackage{algorithm}
\usepackage{algorithmic}
\usepackage{bm}
\usepackage{adjustbox}

\usepackage{tikz,lipsum,lmodern}
\usepackage[most]{tcolorbox}

\def\bBoxBass{\begin{tcolorbox}[colback=blue!5!white,colframe=black!75!black] \begin{assumption}}
\def\bBoxEass{\end{assumption}\end{tcolorbox}}

\DeclareMathOperator*{\argmin}{argmin}

\def\1{{\bf 1}}
\def\0{{\bf 0}}
\renewcommand{\a}{{\bf a}}
\renewcommand{\b}{{\bf b}}

\newcommand{\f}{{\bf f}}
\newcommand{\g}{{\bf g}}

\renewcommand{\v}{{\bf v}}

\newcommand{\x}{{\bf x}}
\def\X{{\bf X}}
\newcommand{\y}{{\bf y}}
\newcommand{\z}{{\bf z}}

\newcommand{\G}{{\bf G}}

\newcommand{\R}{{\bf R}}

\def\dd{\rm D}

\def\E{\mathbb{E}}

\def\({\left(}
\def\){\right)}
\def\1{{\bf 1}}

\newcounter{ass_counter}

\newtheorem{assumption}[ass_counter]{Assumption}
\allowdisplaybreaks
\newcommand\numberthis{\addtocounter{equation}{1}\tag{\theequation}}

\title{Distributed Learning Systems with
First-order Methods}
\subtitle{Systems and Theory}

\maintitleauthorlist{
Ji Liu \\
University of Rochester \\
ji.liu.uwisc@gmail.com \\
\and
Ce Zhang \\
ETH Zurich \\
ce.zhang@inf.ethz.ch
}

\issuesetup
{%
 copyrightowner={A.~Heezemans and M.~Casey},
 volume        = xx,
 issue         = xx,
 pubyear       = 2018,
 isbn          = xxx-x-xxxxx-xxx-x,
 eisbn         = xxx-x-xxxxx-xxx-x,
 doi           = 10.1561/XXXXXXXXX,
 firstpage     = 1, %Explain
 lastpage      = 18
 }

\usepackage{mwe}

\author[1]{Liu,Ji}
\author[2]{Zhang,Ce}

\affil[1]{University of Rochester, Kuaishou Inc.; ji.liu.uwisc@gmail.com}
\affil[2]{ETH Zurich; ce.zhang@inf.ethz.ch}

\articledatabox{\nowfntstandardcitation}

\begin{document}

\makeabstracttitle

\begin{abstract}
Scalable and efficient distributed learning
is one of the main driving forces behind
the recent rapid advancement of machine learning and artificial intelligence. One prominent feature of this
topic is that recent progresses have been
made by researchers in {\em two} communities:
(1) {\em the system community} such as database,
data management, and distributed systems, and
(2) {\em the machine learning and
mathematical optimization community}. 
The interaction and knowledge sharing
between these two communities has led to the
rapid development of new distributed learning
systems and theory.

In this work, we hope to provide a 
brief introduction of some distributed
learning techniques that have recently
been developed, namely {\em lossy communication compression} (e.g., quantization and
sparsification), \textit{asynchronous communication},
and \textit{decentralized communication}. One special focus in this
work is on making sure that it can
be easily understood by researchers
in {\em both} communities --- On the system
side, we rely on a simplified system model
hiding many system details that are 
not necessary for the intuition behind the system speedups;
while, on the theory side, we rely on minimal 
assumptions and significantly 
simplify the proof of some recent work
to achieve comparable results. 
\end{abstract}

\chapter*{Notations and Definitions}
Throughout this article, we make the following definitions.
\begin{itemize}
\item All vectors are assumed to be column vectors by default;
\item $\alpha$, $\beta$, and $\gamma$ usually denote constants;
\item Bold small letters usually denote vectors, such as $\x$, $\y$, and $\v$;
\item $\langle \x, \y \rangle$ denotes the dot product between two vectors $\x$ and $\y$;
\item Capital letters usually denote matrices, such as $W$;
\item $\lesssim$ means ``small and equal to up to a constant factor'', for example, $a_t \lesssim b_t$ means that there exists a constant $\alpha > 0$ independent of $t$ such that $a_t \leq \alpha b_t$;
\item $\1$ denotes a vector with $1$ at everywhere and its dimension depends on the context;
\item $f'(\cdot)$ denotes the gradient or differential of the function $f(\cdot)$;
\item $[M]:=\{1,2,\cdots, M\}$ denotes a set containing integers from $1$ to $M$. 
\end{itemize}

\chapter{Introduction}

Real-world distributed learning systems, especially
those relying on first-order methods, are
often constructed in two ``phases'' --- first comes the textbook (stochastic) gradient
descent (\SGD) algorithm, and then certain aspects of the system design are ``relaxed'' to remove
the system bottleneck, be that communication
bandwidth, latency, synchronization cost, etc.
Throughout this work, we will describe multiple
popular ways of system relaxation developed in
recent years and analyze their system behaviors and
theoretical guarantees.

In this \part, we provide the background for both
the theory and the system. On the theory side, we describe the
intuition and theoretical properties of standard
gradient descent (\GD) and stochastic gradient descent (\SGD)
algorithms (we refer the readers to
\cite{Bottou2016-ax} for
more details). On the system side, we introduce a
simplified performance model that hides 
many details but is just sophisticated enough  
for us to reason about the performance impact
of the different system relaxation techniques that we
will introduce in the later \parts.

%\begin{table}[]
\begin{sidewaystable}[]
%\scriptsize
\centering
%\begin{minipage}[20cm]
\begin{tabular}{cc|l l}
\hline
Algorithm & System Optimization & \# Iterations to $\epsilon$ & Communication Cost \\
\hline
\GD    & /     &  $O\left({1\over \epsilon}\right)$  & N/A   \\
\SGD   & /    & $O\left({1\over \epsilon} + {\sigma^2 \over \epsilon^2}\right)$  & N/A  \\
\hline
\mSGD & Distributed Baseline &  $O\left({1\over \epsilon} + {\sigma^2\over N\epsilon^2}\right)$   & $O(N\alpha + \beta)$  \\
% \hline
\CSGD & Compression &   $O\left({1\over \epsilon} + {\sigma^2\over N\epsilon^2} + {\sigma'^2 \over \epsilon^2} \right)$   & $O(N\alpha + \beta\eta)$ \\
\DS & Compression & $ O\left({1\over \epsilon} + {\sigma^2\over N\epsilon^2}+ {\sigma' \over \epsilon^{2/3}}\right)$    &  $O(N\alpha + \beta\eta)$ \\
\ASGD & Asynchronization &  $ O\left({N\over \epsilon} + {\sigma^2\over N\epsilon^2} \right)$    & $O(N\alpha + \beta)$ \\
\DSGD & Decentralization &   $ O\left({1\over \epsilon} + {\sigma^2\over N\epsilon^2}+ {\rho\varsigma \over (1-\rho)\epsilon^{2/3}}\right)$  & $O(\text{deg}(G)(\alpha + \beta))$ \\
\hline
\end{tabular}
\caption{Summary of results covered in this work.
For distributed settings, we assume that there are
$N$ workers and the latency and the bandwidth of
the network are $\alpha$ and $\beta$, respectively.
The lossy compression scheme 
has a compression ratio of 
$\eta (<1)$ (which introduces
additional variance $\sigma'$
to the gradient estimator) and the decentralized communication 
scheme uses a communication graph $g$ 
of degree $\text{deg}(G)$. $\varsigma$ measures the data variation among workers in the decentralized scenario -- $\varsigma=0$ if all workers have the same dataset.
We assume the
simplified communication model and communication pattern
as described in Section~\ref{sec:commmodel}.}
\label{tab:summary}
%\end{minipage}
\end{sidewaystable}
%\end{table}

{\rv
\paragraph*{Summary of Results}
In this work, we focus on three different
system relaxation techniques, namely {\em lossy communication compression}, \textit{asynchronous communication}, and \textit{decentralized communication}. For each system relaxation
technique, we study their convergence behavior
(i.e., \# iterations we need to achieve 
$\epsilon$ precision) and the communication
cost per iteration. Table~\ref{tab:summary}
summarizes the results we will cover in
this work.
}

\section{Gradient Descent}

Let us consider the generic machine learning objective that can be summarized by the following form
\begin{align}
\min_{\x \in \mathbb{R}^d}\quad \left\{f(\x):={1\over M}\sum_{m=1}^M F_m(\x)\right\}.
\label{eq:obj-gd}
\end{align}
Let $f^\star := \min_{\x} f(\x)$ and assume that it exists by default.
{\rv Each $F_m$ corresponds to a \textit{data sample} in 
the context of machine learning.}

The gradient descent ($\GD$) can be described as
\begin{align}
(\GD)\quad\x_{t+1} = \x_t - \gamma f'(\x_t)
\label{eq:gd}
\end{align}
where $t$ is the iteration index
and {\rv $f'(\x_t)$ is the gradient of 
$f$ at $\x_t$}.

\subsection{Intuitions}
We provide two intuitions about the gradient descent $\GD$ algorithm to indicate why it will work:
%\begin{itemize}
%\item ({\bf Steepest descent direction}) 
\paragraph{Steepest descent direction}
The gradient (or a differential) of a function is the steepest direction to increase the function value given an infinitely small step, which can be seen from the property of the function gradient $\forall \|\v\| = 1$
\begin{align*}
\langle f'(\x), \v\rangle = f'_{\v}(\x) := \lim_{\delta \rightarrow 0}\frac{f(\x+\v\delta) - f(\x)}{\delta}.
\end{align*}
$f'_{\v}(\x)$ is the directional gradient, which indicates how much increment there is on function value along the direction $\v$ by a tiny unit step. To find the steepest unit descent direction is to maximize 
\[
\max_{\|\v\|=1} \quad f'_{\v}(\x).
\]
Since $f'_\v(\x) = \langle f'(\x), \v\rangle$, it is easy to verify that the steepest direction is $\v^\star = {f'(\x) \over \|f'_(\x)\|}$. Note that our goal is to minimize the function value. Therefore, $\GD$ is a natural idea via moving the model $\x_t$ along the steepest ``descent'' direction $-f'(\x_t)$. 
%\item ({\bf Minimizing a model function}) 

\begin{figure}
\centering
\includegraphics[width=0.75\textwidth]{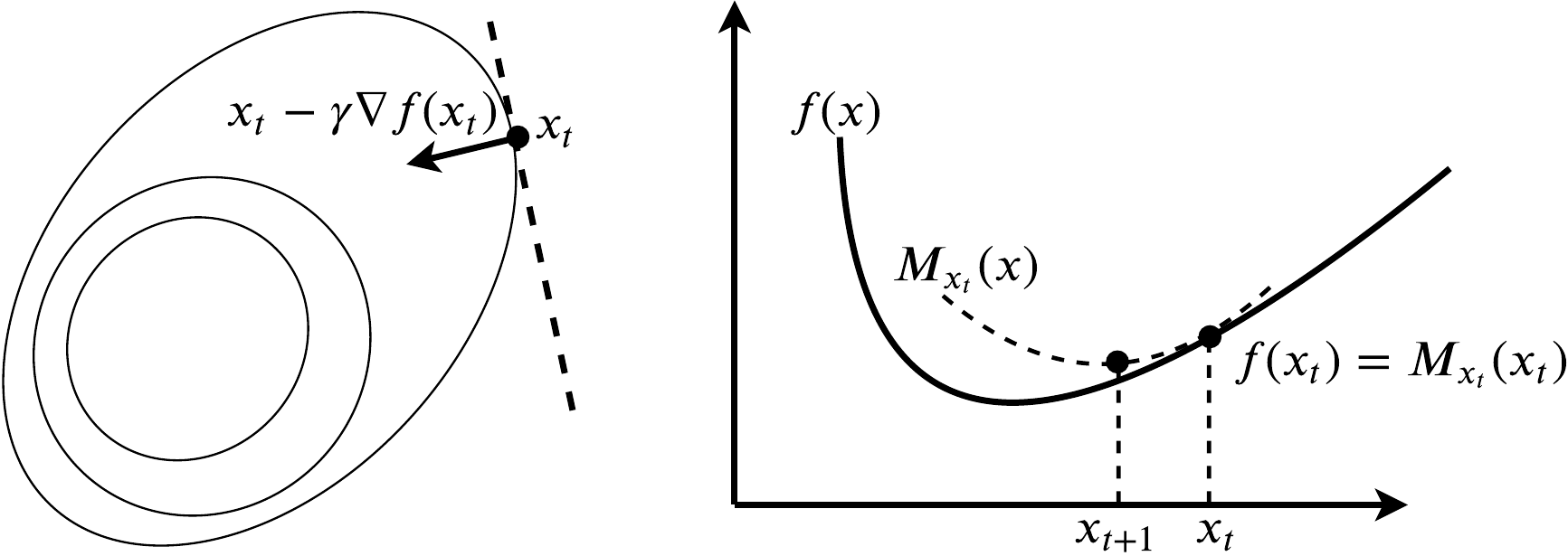}
\caption{(Left) Illustration of gradient and steepest descent
direction; (Right) Illustration of model function.}
\label{fig:grad}
\end{figure}

\paragraph{Minimizing a model function}
Another perspective from which to view gradient descent is based on the model function. Since the original objective function $f(\x)$ is usually very complicated, it is very hard to minimize the objective function directly. A straightforward idea is to construct a model function to locally approximate (at $\x_t$) the original objective in each iteration. The model function needs to be simple and to approximate the original function well enough. Therefore, the most natural idea is to choose a quadratic function (that is usually simple to solve)
\[
M_{\x_t,\gamma}(\x):= f(\x_t) + \langle f'(\x_t), \x - \x \rangle + {1 \over 2\gamma}\|\x - \x_t\|^2.
\]
This model function is a good approximation in the sense that
\begin{itemize}
\item $f(\x_t) = M_{\x_t, \gamma}(\x_t)$
\item $f'(\x_t) = M'_{\x_t, \gamma}(\x_t)$
\item $f(\cdot) \leq M_{\x_t, \gamma}(\cdot)$ if the learning rate $\gamma$ is sufficiently small.
\end{itemize}
For the first two, it is easy to understand why they are important. The last one is important to the convergence, which will be seen soon. Figure~\ref{fig:grad} illustrates the geometry of the model function. One can verify that the $\GD$ algorithm is nothing but iteratively update the optimization variable $\x$ via minimizing the model function at the current point $\x_t$:
\begin{align*}
\x_{t+1} = &\argmin_{\x}~M_{\x_t, \gamma}(\x) 
\\= &\argmin_{\x}~{1\over 2\gamma}\left\|\x - \(\x_t- \gamma f'(\x_t)\)\right\|^2 + \text{constant}
\\ = & \x_t- \gamma f'(\x_t).
\end{align*}
The convergence of $\GD$ can also be revealed by this intuition --- $\x_{t+1}$ always improves $\x_t$ unless the gradient is zero
\[
f(\x_{t+1}) \leq M_{\x_t, \gamma}(\x_{t+1}) \leq M_{\x_t, \gamma}(\x_{t}) = f(\x_t),
\]
where $f(\x_{t=1}) = f(\x_t)$ holds if and only if $f'(\x_t)=0$.
%\end{itemize}

\subsection{Convergence rate}
From the intuition of $\GD$, the convergence of $\GD$ is automatically implied. This section provides the convergence \emph{rate} via rigorous analysis. To show the convergence rate, let us first make some commonly used assumptions in the following.
\begin{tcolorbox}[colback=pink!5!white,colframe=black!75!black]
\begin{assumption} \label{ass:gd}
We assume:
\begin{itemize} [leftmargin=*]
\item ({\bf Smoothness}) All functions $F_m(\cdot)$'s are {\rv differentiable}.
\item ({\bf $L$-Lipschitz gradient}) The objective function is assumed to have a Lipschitz gradient, that is, there exists a constant $L$ satisfying $ \forall \x, \; \forall \y$
\begin{align}
\|f'(\x) - f'(\y)\| \leq & L\|\x- \y\| \quad \label{eq:ass:gd_1}
\\
f(\y) - f(\x) \leq & \langle f'(\x), \y-\x \rangle + {L\over 2} \|\y - \x\|^2  \label{eq:ass:gd_2}
\end{align}
\end{itemize}
%\item ({\bf Bounded $f(\x)$ from below})
\end{assumption}
\end{tcolorbox}
The smoothness assumption on $F_m(\cdot)$'s implies that the overall objective function $f(\cdot)$ is {\rv differentiable or smooth} too. The assumption~\eqref{eq:ass:gd_2} can be deduced from \eqref{eq:ass:gd_1}, {\rv and we refer readers to the textbook by \citet{boyd2004convex} or their course link\footnote{\url{http://www.seas.ucla.edu/~vandenbe/236C/lectures/gradient.pdf}}}. The Lipschitz gradient assumption essentially assumes that the curvature of the objective function is bounded by $L$.
We make the assumption of \eqref{eq:ass:gd_2} just for convenience of use later. 

We apply the Lipschitz gradient assumption and immediately obtain the following golden inequality: 
\begin{align*}
f(\x_{t+1}) - f(\x_{t}) \leq & \langle f'(\x_{t}), \x_{t+1} - \x_t  \rangle + {L\over 2}\|\x_{t+1} - \x_t\|^2
\\ = &
- \gamma \| f'(\x_t) \|^2 + {\gamma^2 L \over 2} \|f'(\x_t)\|^2
\\ = &
-\gamma\left(1- {\gamma L\over 2}\right) \|f'(\x_t)\|^2
\numberthis
\label{eq:gd_proof_1}
\end{align*}
We can see that as long as the learning rate $\gamma$ is small enough such that $1-{\gamma L}/2 > 0$, $f(\x_{t+1})$ can improve $f(\x_t)$. Therefore, the learning rate cannot be too large to guarantee the progress in each step. However, it is also a bad idea if the learning rate is too small, since the progress is proportional to $\gamma (1-\gamma L /2)$. The optimal learning rate can be obtained by simply maximizing 
\[
\gamma (1-\gamma L /2)
\]
over $\gamma$, which gives the optimal learning rate for the gradient descent method as $\gamma^{\star} = 1/L$. Substituting $\gamma = \gamma^\star$ into \eqref{eq:gd_proof_1} yields
\begin{align*}
f(\x_{t+1}) - f(\x_t) \leq -{1 \over 2L} \|f'(\x_t)\|^2
\end{align*}
or equivalently 
\begin{align}
f(\x_{t}) - f(\x_{t+1}) \geq {1 \over 2L} \|f'(\x_t)\|^2.
\label{eq:gd_proof_2}
\end{align}
% Acute readers may ask: How do I know the value of the Lipschitz constant $L$? Unfortunately, it is usually unknown; however, in most practical scenarios, it is not hard to find an upper for $L$.

Summarizing Eq.~\eqref{eq:gd_proof_2} over $t$ from $t=1$ to $t=T$ yields
\begin{align*}
{1\over 2L} \sum_{t=1}^T \|f'(\x_t)\|^2 \leq & \sum_{t=1}^T\left(f(\x_{t}) - f(\x_{t+1})\right)  
\\ = &
f(\x_1) - f(\x_{t+1})
\\ \leq &
f(\x_1) - f^\star.
\end{align*}
Rearranging the inequality yields the following convergence rate for gradient descent:
\begin{tcolorbox}[colback=blue!5!white,colframe=black!75!black]
\begin{theorem}\label{thm:gd}
Under Assumption~\ref{ass:gd}, the gradient descent method admits the following convergence rate
\begin{align}
{1\over T}\sum_{t=1}^T\|f'(\x_t)\|^2 \lesssim {L \over T}
\end{align}
by choosing the learning rate $\gamma = {1\over L}$. Here, we treat $f(\x_1) - f^{\star}$ as a constant.
\end{theorem}
\end{tcolorbox}
This result indicates that the averaged gradient norm converges in the rate of $1/T$. {\rv It is worth noting that, unlike the convex case, we are unable to use the commonly used criterion $f(\x_t) - f^{\star}$ to evaluate the convergence (efficiency). That is to say, the algorithm guarantees the convergence only to a stationary point ($\|f'(\x_t)\|^2 \rightarrow 0$) because of the nonconvexity. The connection between two criteria $f(\x_t) - f^\star$ and $\|f'(\x_t)\|^2$ can be seen from
\begin{align*}
    {1\over L}\|f'(\x_t)\|^2 \leq f(\x_t) - f^\star.
\end{align*}
The proof can be found in the standard textbook or the course link\footnote{https://www.cs.rochester.edu/~jliu/CSC-576/class-note-6.pdf}.}

There are two major disadvantages for the $\GD$ method:
\begin{itemize}
\item The computational complexity and system overhead can be 
too high in each iteration to compute a single gradient;
\item For nonconvex objectives, the gradient descent often sticks on a bad (shallow) local optimum.
\end{itemize}

\subsection{Iteration / query / computation complexity} \label{subsec:gd_complexity}
The convergence rate is the key to analyzing the overall complexity. People usually consider three types of overall complexity: iteration complexity, query complexity, and computation complexity. To evaluate the overall complexity to solve the optimization problem in~\eqref{eq:obj-gd}, we need first to specify a precision of our solution, since in practice it is difficult (also not really necessary) to exactly solve the optimization problem. In particular, in our case the overall complexity must take into account how many iterations / queries / computations are required to ensure the average gradient norm ${1\over T}\sum_{t=1}^T \|f'(\x_t)\|^2 \leq \epsilon$.

\paragraph{Iteration complexity.} From Theorem~\ref{thm:gd}, it is straightforward to verify that the iteration complexity is
\begin{align}
O\({L\over \epsilon}\).
\label{eq:gd_ic}
\end{align}

\paragraph{Query complexity.} Here ``query'' refers to the number of queries of the data samples. $\GD$ needs to query all $M$ samples in each iteration. Therefore, the query complexity can be computed from the iteration complexity by multiplying the number of queries in each iteration
\begin{align}
O\({LM \over \epsilon}\).
\label{eq:gd_qc}
\end{align}

\paragraph{Computation complexity.} Similarly, the computation complexity can be computed from the query complexity by multiplying the complexity of computing one sample gradient $F'_{m}(\x)$. The typical complexity of computing one sample gradient is proportional to the dimension of the variable, which is $d$ in our notation. To see the reason, let us imagine a naive linear regression with $F_m:= {1\over 2}(\a_m^\top \x -b)^2$ and a sample gradient of $f'_{m}(\x):=\a_m(\a_m^\top \x - b)$. Therefore, the computation complexity of $\GD$ is
\[
O\({LMd \over \epsilon}\).
\]
It is worth pointing out that the computation complexity is usually proportional to the query complexity (no matter for what kinds of objective) if we consider and compare only first-order (or sample-gradient-based) methods. Therefore, in the remainder of this work, we compare only the query complexity and the iteration complexity.

%The major computational burden in $\GD$ is the iterative gradient computation. To compute the gradient of $f'(\x):={1\over M} F'_m(\x)$, the typical computational complexity is $O(Md)$ where $M$ is the number of samples and $d$ the dimension of the model $\x$. 
%The typical computational complexity to compute the gradient is $O(Md)$. To see the reason, let us imagine a naive linear regression with $F_m:= {1\over 2}(\a_m^\top \x -b)^2$ and a gradient $f'(\x):={1\over M}\sum\a_m(\a_m^\top \x - b)$.
%
%The typical way is to take account of the overall computational complexity to achieve a solution with $\epsilon$ precision, which is
%\begin{align}
%O\(L\over \epsilon}\),
%\end{align}
%and the overall computational complexity is 
%\begin{align}
%O\({LMd \over \epsilon}\)
%\end{align}. 

%%%%%%%%%%%%%%%%%%%%%%%%%%%%%%%%%%%%%%%%%%%%%%%%
%%                                                              									    %%
%%%%%%%%%%%%%%%%%%%%%%%%%%%%%%%%%%%%%%%%%%%%%%%%
\section{Stochastic Gradient Descent}

One disadvantage of $\GD$ is that it requires one to query all samples in an iteration, which could be overly expensive. To overcome this shortcoming, the stochastic gradient method $\SGD$ is widely used in machine learning training. Instead of computing a full gradient in each iteration, it is usual to compute only the gradient on a batch (or minibatch) of sampled data. In particular, people randomly sample an $m_t\in [M]$ independently each time and update the model by
\begin{align}
(\SGD)\quad \x_{t+1} = \x_t - \gamma F'_{m_t}(\x_t),
\label{eq:sgd}
\end{align}
%{\rc They then need to adjust the notation $\nabla F_{m_t}(\x_t)$ to make them consistent globally.}
% CZ: Done
where $m_t \in [M]$ denotes the index randomly selected at the $t$th iteration. $F'_{m}(\x)$ (or $F'_{m_t}(\x_t)$) is called the stochastic gradient (at the $t$th iteration). We use $\g(\cdot):=F'_m(\cdot)$ (or $\g_t(\cdot):=F'_{m_t}(\cdot)$) to denote the stochastic gradient (or at the $t$th iteration) for short. An important property for the stochastic gradient is that its expectation is equal to the true gradient, that is,
\begin{align*}
\E[\g(\x)] = \E_{m} [F'_{m}(\x)] = f'(\x) \quad \forall \x.
\end{align*}
%<<<<<<< HEAD
An immediate advantage of $\SGD$ is that the computational complexity reduces to $O(d)$ per iteration. It is worth pointing out that the $\SGD$ algorithm is NOT a descent algorithm\footnote{\rv A descent algorithm means $f(\x_{t+1})\leq f(\x_t)$, that is, $\x_{t+1}$ is always not worse than $\x_t$ for any iterate $t$.} due to the randomness.

\subsection{Convergence rate}\label{sec:SGD:convergence}
The next questions are whether it converges and, if it does, how quickly. We first make a typical assumption:
\begin{tcolorbox}[colback=pink!5!white,colframe=black!75!black]
\begin{assumption} \label{ass:sgd}
%=======
%A clear advantage of $\SGD$ is that the computational complexity reduces to $O(d)$ per iteration. It is worth pointing out that the $\SGD$ algorithm is NOT a descent algorithm due to the randomness.
%
%The next question is whether it converges and how fast if yes. We first make some typical assumption
%\begin{tcolorbox}[colback=pink!5!white,colframe=black!75!black]
%\begin{assumption}
%>>>>>>> b94b8679f310fa566039184f25fe4da4b1ce56ff
We make the following assumption:
\begin{itemize}
\item ({\bf Unbiased gradient}) The stochastic gradient is unbiased, that is, 
\[
\E_{m}[F'_m(\x)] = f'(\x)\quad \forall \x;
\]
\item ({\bf Bounded stochastic variance}) The stochastic gradient is with bounded variance, that is, there exists a constant $\sigma$ satisfying
\[
\E_m[\|F'_m(\x) - f'(\x)\|^2] \leq \sigma^2\quad \forall \x
\]
\end{itemize}
\end{assumption}
\end{tcolorbox}
We first apply the Lipschitzian gradient property in Assumption~\ref{ass:gd}:
\begin{align*}
f(\x_{t+1}) - f(\x_{t}) \leq & \langle f'(\x_{t}), \x_{t+1} - \x_t  \rangle + {L\over 2}\|\x_{t+1} - \x_t\|^2
\\ = &
-\gamma \langle f'(\x_t), \g_t(\x_t) \rangle + {L\gamma^2\over 2} \|\g_t(\x_t)\|^2. \numberthis 
\label{eq:sgd_proof_1}
\end{align*}
Note two important properties:
\begin{itemize}
\item $\E[\langle f'(\x_t), \g_t(\x_t) \rangle] = \langle f'(\x_t), \E[\g_t(\x_t)] \rangle =  \|f'(\x_t)\|^2$
\item $\E[\|\g_t(\x_t)\|^2] = \|f'(\x_t)\|^2 + \E[\|\g_{t}(\x_t) - f'(\x_t)\|^2] \leq \|f'(\x_t)\|^2 + \sigma^2$,
\end{itemize}
where the second property uses the property of variance, that is, any random variable vector $\xi$ satisfies 
\begin{align}
\E[\|\xi\|^2] = \|\E[\xi]\|^2 + \E[\|\xi- \E[\xi]\|]^2.
\label{eq:mean-var}
\end{align}
Apply these two properties to \eqref{eq:sgd_proof_1} and take expectation on both sides:
\begin{align*}
& \E[f(\x_{t+1})] - \E[f(\x_t)] 
\\ \leq & 
-\gamma \E[\|f'(\x_t)\|^2] + {L\gamma^2 \over 2} \left(\E[\|f'(\x_t)\|^2] + \sigma^2 \right) \numberthis \label{eq:sgd_proof_2}
\\ \leq &
-\gamma\left(1 - {\gamma L \over 2}\right)  \E[\|f'(\x_t)\|^2] + {\gamma^2 \over 2} L\sigma^2. \numberthis %\label{eq:sgd_proof_2}
\end{align*}
From \eqref{eq:sgd_proof_2}, we can see that $\SGD$ does not guarantee ``descent'' in each iteration, unlike $\GD$, but it does guarantee ``descent'' in the expectation sense in each iteration as long as $\gamma$ is small enough and $\|f'(\x_t)\|^2 > 0$. This is because the first term in \eqref{eq:sgd_proof_2} is in the order of $O(\gamma)$ while the second term is in the order of $O(\gamma^2)$.

Next we summarize \eqref{eq:sgd_proof_2} from $t=1$ to $t=T$ and obtain
\begin{align}
\E[f(\x_{T+1})] - f(\x_1) \leq -\gamma\left(1-{\gamma L \over 2} \right)\sum_{t=1}^T\E[\|f'(\x_t)\|^2] + {\gamma^2 \over 2} TL\sigma^2.
\label{eq:sgd_proof_3}
\end{align}
We choose the learning rate $\gamma = {1\over L+ \sigma\sqrt{TL}}$ which implies that $(1-\gamma L/2)> 1/2$. It follows
%to simplify the right-hand side of \eqref{eq:sgd_proof_3}
\begin{align*}
&{1\over T} \sum_{t=1}^T\E[\|f'(\x_t)\|^2] 
\\ \lesssim & 
\frac{f(\x_1)- \E[f(\x_{T+1})]}{T\gamma} + \gamma L\sigma^2
\\  \lesssim &
\frac{f(\x_1)- f^\star}{T\gamma} + \gamma L\sigma^2
\\ \lesssim &
{(f(\x_1) - f^\star)L \over T} + {(f(\x_1) - f^\star)\sqrt{L}\sigma \over \sqrt{T}}. 
\end{align*}
Therefore the convergence rate of $\SGD$ can be summarized into the following theorem
\begin{tcolorbox}[colback=blue!5!white,colframe=black!75!black]
\begin{theorem}\label{thm:sgd}
Under Assumptions~\ref{ass:gd} and \ref{ass:sgd}, the $\SGD$ method admits the following convergence rate
\begin{align*}
{1\over T}\sum_{t=1}^T\E[\|f'(\x_t)\|^2] \lesssim {L \over T} + {\sqrt{L}\sigma \over \sqrt{T}}. 
\end{align*}
by choosing the learning rate $\gamma = {1\over L+\sigma \sqrt{TL}}$. Here we treat $f(\x_1) - f^{\star}$ as a constant.
\end{theorem}
\end{tcolorbox}
%{\rc need to fix the learning rate.}
We highlight the following observations from Theorem~\ref{thm:sgd}
\begin{itemize}
\item ({\bf Consistent with $\GD$}) If $\sigma=0$, the $\SGD$ algorithm reduces to $\GD$ and the convergence rate becomes ${L(f(\x_1)-f^\star)/L}$, which is consistent with the convergence rate for $\GD$ proven in Theorem~\ref{thm:gd}.
\item ({\bf Asymptotic convergence rate}) The convergence rate of $\SGD$ achieves $O(1/\sqrt{T})$. 
\end{itemize}

\subsection{Iteration / query complexity}
Using a similar analysis as Section~\ref{subsec:gd_complexity}, we can obtain the iteration complexity of $\SGD$, which is also the query complexity (since there is only one query per one sample gradient)
\begin{align*}
O\({L\over \epsilon} + {L\sigma^2 \over \epsilon^2}\). 
\end{align*}
It is worse than $\GD$ in terms of the iteration complexity in \eqref{eq:gd_ic}, which is not a surprising result. The comparison of query complexity makes more sense since it is more related to the physical running time or the computation complexity. From the detailed comparison in Table~\ref{tab:cc}, we can see that %{\rc need to fix the wrong number!}
\begin{itemize}
\item $\SGD$ is superior to $\GD$, if ${\sigma^2 \over M} \ll \epsilon$;
\item $\SGD$ is inferior to $\GD$, if ${\sigma^2 \over M } \gg \epsilon$.
\end{itemize}
It is worth pointing out that when the number of samples $M$ is huge and a low precision solution is satisfactory\footnote{A low precision solution is satisfactory in many application scenarios, since a high precision solution may cause an unwanted overfitting issue.}, ${\sigma \over M\sqrt{L}} \ll \epsilon$ usually holds. As a result, $\SGD$ is favored for solving big data problems. 
\begin{table} 
\centering
\begin{tabular}{r c c}
\hline
algorithms & iteration complexity & query complexity
\\ \hline
$\GD$ & $O\({L\over \epsilon}\)$ & $O\({ML\over \epsilon}\)$
\\ 
$\SGD$ & $O\({L\over \epsilon} + {L\sigma^2 \over \epsilon^2}\)$ & $O\({L\over \epsilon} + {L\sigma^2 \over \epsilon^2}\)$
\\ 
$\mSGD$ & $O\({L\over \epsilon} + {L\sigma^2 \over B\epsilon^2}\)$ & $O\({LB\over \epsilon} + {L\sigma^2 \over \epsilon^2}\)$
\\ \hline
\end{tabular}
\caption{Complexity comparison among $\GD$, $\SGD$, and $\mSGD$.}\label{tab:cc}
\end{table}

\subsection{Minibatch stochastic gradient descent ($\mSGD$)} \label{Chap:sec:msgd}
A straightforward variant of the $\GD$ algorithm is to compute the gradient of a minibatch of samples (instead of a single sample) in each iteration, that is, 
\begin{align}
\g^{\mathcal{B}}(\x) = {1\over B}\sum_{m\in \mathcal{B}}F'_{m}(\x),
\label{eq:mSGD_mb}
\end{align}
where $B:=|\mathcal{B}|$. The minibatch $\mathcal{B}$ is obtained by using i.i.d samples with (or without) replacement. One can easily verify that
\begin{align*}
\E[\g^{\mathcal{B}}(\x)] = & f'(\x).
\end{align*}
\paragraph{\rv Sample ``with'' replacement.} The stochastic variance (for the ``with'' replacement case) can be bounded by
\begin{align*}
& \E[\left\|\g^{\mathcal{B}}(\x) - f'(\x)]\right\|^2] 
%\\ = &
%\E\left[\left\|g^{\mathcal{B}}(\x) - \E\left[g^{\mathcal{B}}(\x)\right]\right\|^2\right] 
\\ = &
 \E\left[ \left\| {1\over B} \sum_{m\in \mathcal{B}}\(F'_m(\x) - f'(\x)\) \right\|^2\right] \numberthis \label{eq:wsample:1}
\\ = &
{1\over B} \sum_{m\in \mathcal{B}} \E\left[ \left\|F'_m(\x) - f'(\x) \right\|^2\right]
\\ \leq &
{\sigma^2 \over B}\quad \(\text{from Assumption~\ref{ass:sgd}}\).
\end{align*}

\paragraph{\rv Sample ``without'' replacement}
{\rv
The stochastic variance for the ``without'' replacement is even smaller, but it involves a bit more complicated derivation. We essentially need the following key lemma
\begin{tcolorbox}[colback=yellow!5!white,colframe=black!75!black]
\begin{lemma} \label{lemma:wosample}
Give a set including $M\geq 2$ real numbers $\{a_1, a_2, \cdots, a_M\}$. Define a random variable 
\begin{align*}
    \bar{\xi}_{[B]} := &{1\over B}\sum_{m=1}^B \xi_m,
    %\\
    %\tilde{z} := & {1\over B}\sum_{m=1}^B Z_m
\end{align*} 
where $\xi_1, \cdots, \xi_B$ are uniformly randomly sampled from the set ``without'' replacement, and $B(1\leq B \leq M)$ is the batch size. Then the following equality holds
\begin{align*}
    {\bf Var}[\bar{\xi}] = \left({M-B \over M-1}\right){{\bf Var}[\xi_1]\over B}.
\end{align*}
\end{lemma}
\end{tcolorbox}
\begin{proof}
First, it is not hard to see that the marginal distributions of $\xi_m$'s are identical. For simplicity of notation, we assume that $\E[\xi_m]=0$ without the loss of generality. Therefore, we have ${\bf Var}[\xi_m] = \E [\xi_m^2]$ for all $k$.

Next we have the following derivation:
\begin{align*}
    {\bf Var}[B\bar{\xi}_{[B]}] = & \E [(B\bar{\xi}_{[B]})^2]\quad \text{(due to $\E[\bar{\xi}_{[B]}]=0$)}
    \\ = &
    \sum_{m=1}^B \E[\xi_m^2] + \sum_{k\neq l}\E[\xi_m \xi_l]
    \\ = &
    B{\bf Var}[\xi_1] + B(B-1)\E[\xi_1 \xi_2],
    \numberthis \label{eq:lemma_proof:wrsample:1}
\end{align*}
where the last equality uses the fact $\E[\xi_m^2] = {\bf Var}[\xi_k] = {\bf Var}[\xi_1]$ for any $k$ and $E[\xi_k\xi_l] = E[\xi_1\xi_2]$ for any $k\neq l$. Note that ${\bf Var}[M\bar{\xi}_{[M]}]=0$, since it has only one possible combination for $\{\xi_1, \xi_2, \cdots, \xi_M\}$. Then letting $B=M$ obtains the following dependence from \eqref{eq:lemma_proof:wrsample:1}
\begin{align*}
    \E[\xi_1 \xi_2] = {-1\over M-1}{\bf Var}[\xi_1].
\end{align*}
Plug this result into \eqref{eq:lemma_proof:wrsample:1}
\begin{align*}
    {\bf Var}[B\bar{\xi}_{[B]}] = B\left({M-B\over M-1}\right){\bf Var}[\xi_1],
\end{align*}
which implies the claimed result.
\end{proof}
If $a_m$'s are vectors and satisfy ${1\over M}\sum_{m=1}^M \xi_m = 0$, from Lemma~\ref{lemma:wosample} one can easily verify 
\begin{align}
    \E\left[\|\bar{\xi}\|^2\right] = B\left({M-B\over M-1}\right)\E\left[\|\xi_1\|^2\right].
    \label{eq:wosample:2}
\end{align}

Now we are ready to compute the stochastic variance for the ``without'' replacement sampling strategy. Let $\mathcal{B}$ be a batch of samples ``without'' replacement. Then we let $\xi_m := F'_m(\x) - f'(\x)$ and from \eqref{eq:wosample:2} obtain 
\begin{align*}
\E\left[\left\|\g^{\mathcal{B}}(\x) - f'(\x)]\right\|^2\right] 
 = & \E \left[ \|\bar{\xi}\|^2\right]
 \\ = & 
\left({M-B \over M-1}\right){\E[\|\xi_1\|^2]\over B}
\\ \leq &
\left({M-B \over M-1}\right){\sigma^2\over B}
\\ \leq &
{\sigma^2\over B}.
\end{align*}

To sum up, we have the stochastic variance bounded by ${\sigma^2 \over B}$ no matter ``with'' or ``without'' replacement sampling.
}

We can observe that the effect of using a minibatch stochastic gradient is nothing but reduced variance. All remaining analysis for the convergence rate remains the same. Therefore, it is quite easy to obtain the convergence rate of $\mSGD$
\begin{align}
{1\over T} \sum_{t=1}^T \E[\|f'(\x_t)\|^2] \lesssim {L\over T} + {\sqrt{L}\sigma \over \sqrt{TB}}.
\label{eq:mSGD_cr}
\end{align}
The iteration complexity and the query complexity are reported in Table~\ref{tab:cc}.

%\begin{align}
%\E[\g(\x)] = \E_{m} [F'_{m}(\x)] = f'(\x) \quad \forall \x.
%\end{align}

\section{A Simplified Distributed Communication Model}
\label{sec:commmodel}

When scaling up the stochastic gradient descent (SGD) algorithm
to a distributed setting, one often needs to develop 
{\em system relaxations} techniques to achieve
better performance and scalability. In this work,
we describe multiple popular system relaxation techniques
that have been developed in recent years. In this section,
we introduce a simple performance model of
a distributed system, which will be used in later \parts
to reason about the performance impact of different relaxation
techniques.

From a mathematical optimization perspective, all of the
system relaxations that we will describe {\em do not
make the convergence (loss vs. \# iterations / epochs) faster}.
\footnote{The reason
that we emphasize the ``mathematical optimization''
perspective is that some researchers
find that certain system relaxations can actually lead
to better generalization performance.
We do not consider generalization in this work.}
Then {\em why do we even want to introduce these 
relaxations into our system in the first place?}

One common theme of the techniques we cover in
this work is that their goal is not
to improve the convergence rate in terms
of \# iterations / epochs; rather, their
goal is to make each iteration finish faster
in terms of wall-clock time.
As a result, to reason about each system relaxation
technique in this work, we need to first
agree on a {\em performance model} of the
underlying distributed system. In this section,
we introduce a very simple performance model --- 
it ignores many (if not most) important system
characteristics, but it is just informative enough
for readers to understand why each system relaxation technique
in this work actually makes a system faster.

\subsection{Assumptions}

\begin{figure}[t!]
\centering
\includegraphics[width=0.3\textwidth]{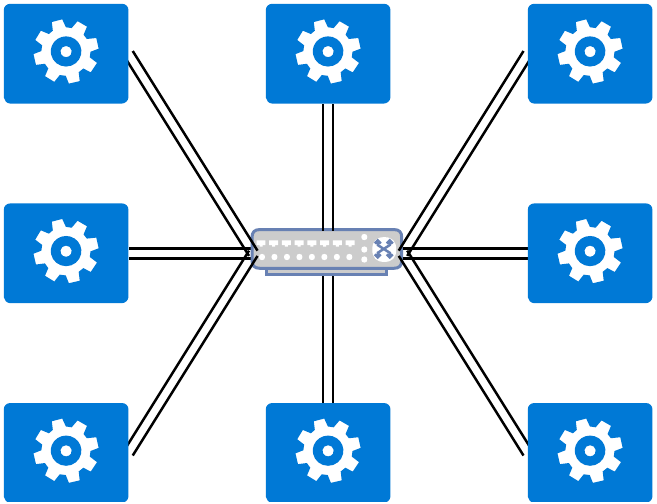}
\caption{An illustration of the distributed communication
model we use in this work. We assume that all devices
(worker, machine) are connected via a ``logical switch''
whose property is defined in Section~\ref{sec:commmodel}}.
\label{fig:commmodel}
\end{figure}

{\rv In practice, it is often the case that the bandwidth 
or latency of each worker's network connection is the dominating bottleneck in the communication cost.
As a result, in this work we focus on the following 
simplified communication model.}

Figure~\ref{fig:commmodel} illustrates our communication model. Each worker (blue rectangle)
corresponds to {\em one} computation device (worker),
and all workers are connected via a ``logical switch''
that has the following property:
\begin{enumerate}
\item The switch has infinitely large bandwidth.
{\rv We make this simplifying assumption to reflect 
the observation that, in practice, the bottleneck is often
the bandwidth 
or latency of each worker's network connection.}
\item For each message that ``passes through''
the switch (sent by worker $w_i$ and received
by worker $w_j$), the switch adds a constant
delay $t_{\text{latency}}$ independently of the number
of concurrent messages that this switch is 
serving. This delay is the timestamp difference
between the sender sending out the first bit
and the receiver receiving the first bit.
\end{enumerate}

For each worker, we also assume the following properties:
\begin{enumerate}
\item Each worker can only send one message 
at the same time.
\item Each worker can only receive one message
at the same time.
\item Each worker can concurrently receive one 
message and send one message at the same time.
\item Each worker has a fixed bandwidth, i.e.,
to send / receive one unit (e.g., MB) amount of data,
it requires $t_{\text{transfer1MB}}$ seconds.
\end{enumerate}

\begin{figure}[t!]
\centering
\includegraphics[width=1.0\textwidth]{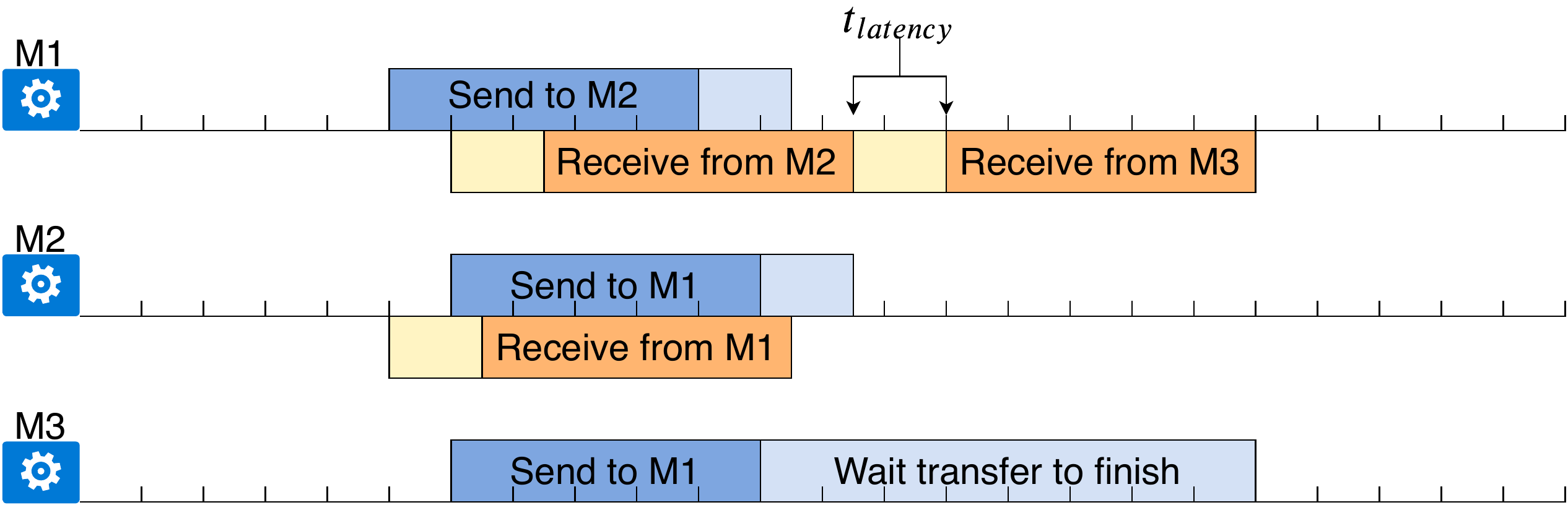}
\caption{Illustration of the communication pattern of Example~\ref{exp:communication_illustration}.}
\label{fig:communication_illustration}
\end{figure}

\begin{figure}[t!]
\centering
\includegraphics[width=1.0\textwidth]{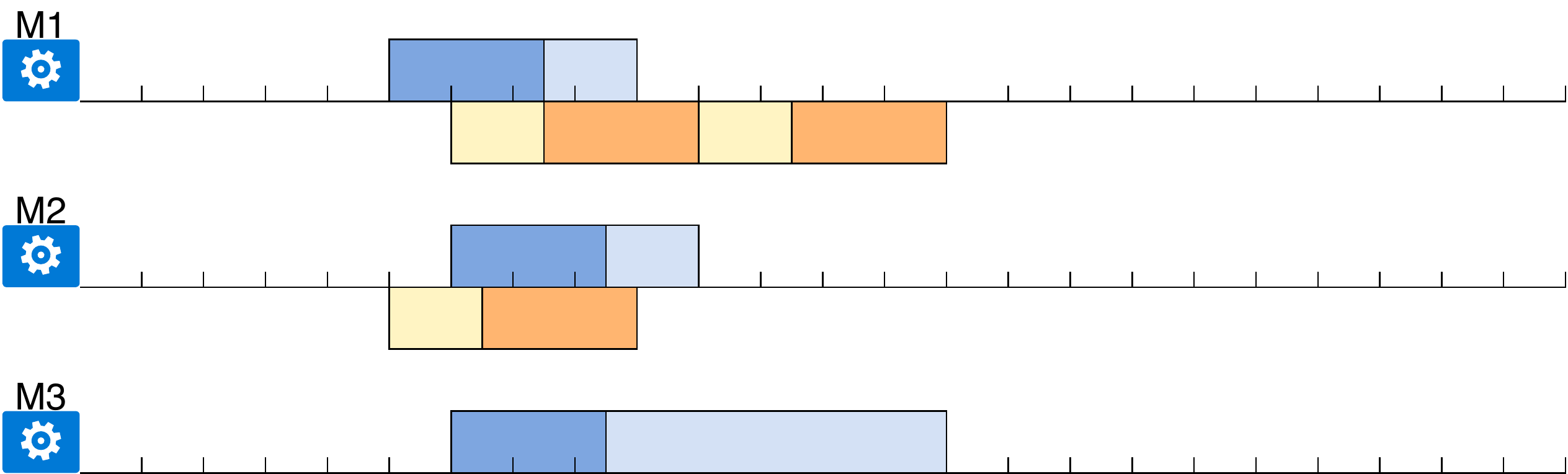}
\caption{Illustration of the communication pattern of Example~\ref{exp:communication_illustration}, with 2x data compression.}
\label{fig:communication_illustration_lowprecision}
\end{figure}

\begin{example} \label{exp:communication_illustration}
Under the above communication model,
consider the following three events:
\begin{verbatim}
            time      event
            0:05      M1 send 1MB to M2
            0:06      M2 send 1MB to M1
            0:06      M3 send 1MB to M2
\end{verbatim}
We assume that the latency added by the switch $t_{\text{latency}}$
is 1.5 units of time and it took 5 units of time to
transfer 1MB of data. 
Figure~\ref{fig:communication_illustration} illustrates 
the timeline on three machines under our communication 
model. The yellow block corresponds to the {\em latency}
added by the ``logical switch''. We also see that 
the machine \texttt{M1} can concurrently send (blue
block) and
receive (orange block) data at the same time; however,
when the machine \texttt{M3} tries to send data to
\texttt{M1}, because the machine \texttt{M2} is already sending data to \texttt{M1}, \texttt{M3}
needs to wait (the shallow blue block of \texttt{M3}).
\end{example}

\begin{example} \label{exp:communication_illustration}
Figure~\ref{fig:communication_illustration_lowprecision}
illustrates a hypothetical scenario in which 
all data sent in Example~\ref{exp:communication_illustration}
are ``magically'' compressed by 2$\times$ at the sender.
As we will see in later \parts, this is similar to what 
would happen if one were to compress the gradient by 2$\times$
during training.

We make multiple observations from Figure~\ref{fig:communication_illustration_lowprecision}.
\begin{enumerate}
\item First, compressing data does make the ``system'' faster.
Without compression, all three events finish in 14 units
of time (Figure~\ref{fig:communication_illustration})
whereas it finishes in 9 units of time after compression.
This is because the time used to {\em transfer}
the data is decreased by half in our communication model.
\item Second, even if the data are compressed by 2$\times$,
the speedup of the system is smaller than that; in fact, it
is only $14/9 = 1.55\times$. This is because, even 
though the transfer time is cut by half, the communication
latency does not decrease as a result of the data compression.
\end{enumerate}
\end{example}

We now use the above communication model to describe
the {\em communication patterns} of 
three popular ways to implement distributed stochastic 
gradient descent. These implementations will often serve
as the baseline from which we apply different system relaxations
to remove certain system bottlenecks that arise in 
different configurations of $(t_{\text{latency}}, t_{\text{transfer}})$
together with the relative computational cost on each machine.

\paragraph*{Workloads} We focus on one of the core building blocks 
to implement a distributed SGD system --- each worker $M_i$ holds a
parameter vector $w_i$, and they communicate to
compute the sum of all parameter vectors: $S = \sum_i w_i$.
At the end of communication, each worker holds one copy of $S$.

\begin{figure}[t!]
\centering
\includegraphics[width=0.3\textwidth]{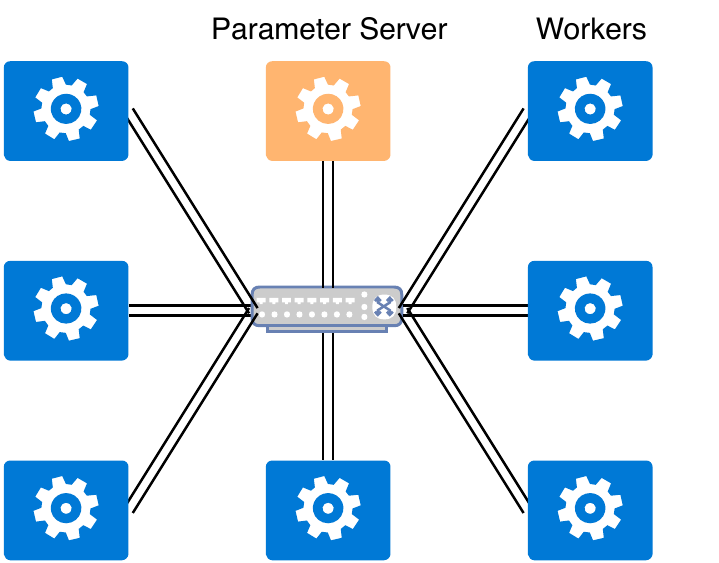}
\caption{Illustration of the parameter server architecture 
with a single dedicated parameter server.}
\label{fig:communication_illustration_ps1}
\end{figure}

\subsection{Synchronous Parameter Server}

The parameter server is not only one of the most popular system
architectures for distributed stochastic gradient descent; it is also
one of the most popular communication models that researchers
have in mind when they conduct theoretical analysis.
In a parameter server architecture, one or more machines
serve as the {\em parameter server(s)} and other machines
serve as the workers processing data. Periodically, workers
send updates to the parameters to the parameter servers and
the parameter servers send back the updated parameters.
Figure~\ref{fig:communication_illustration_ps1} illustrates this
architecture: the orange machine is the parameter
server and the blue machines are the workers.

A real-world implementation of a parameter server architecture
usually involves many system optimizations to speed up the communication.
In this section, we build our abstraction using the simplest 
implementation with only a single machine 
serving as the parameter server. We also scope ourselves and only
focus on the synchronous communication case.

When using this simplified parameter server architecture to calculate
the sum $S$, each worker $M_i$ sends their local parameter vector $w_i$
to the parameter server, and the parameter server collects all these 
local copies, sums them up, and sends back to each worker. In a
simple example with three workers and one parameter server, the
series of communication events looks like this:

\begin{verbatim}
        Time=0      Worker1 send w1 to PS
        Time=0      Worker2 send w2 to PS
        Time=0      Worker3 send w3 to PS
        Time=T      PS send S to Worker1
        Time=T      PS send S to Worker2
        Time=T      PS send S to Worker3
\end{verbatim}

\begin{figure}[t!]
	\centering
	\includegraphics[width=1.0\textwidth]{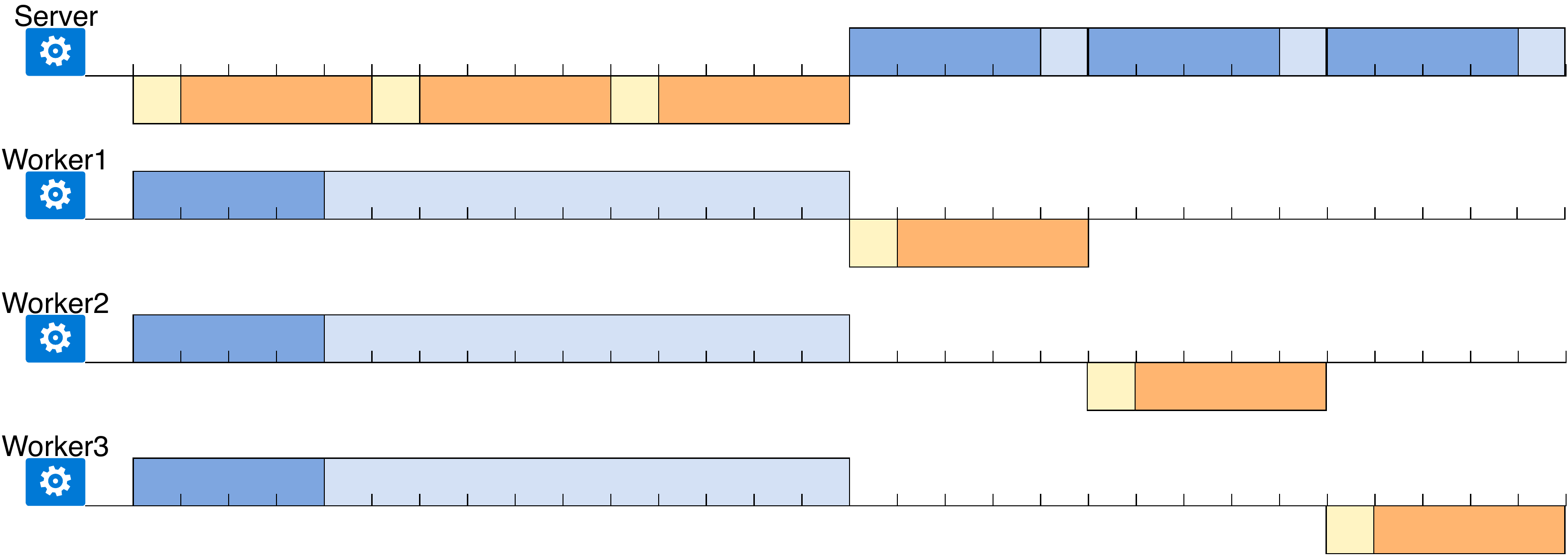}
	\caption{Illustration of the communication
		pattern of the parameter server architecture 
		with a single dedicated parameter server.}
	\label{fig:communication_illustration_timeline_ps1}
\end{figure}

Figure~\ref{fig:communication_illustration_timeline_ps1}
illustrates the communication timeline of these events.
We see that, in the first phase, all workers send their local
parameter vectors to the parameter server at the same time.
Because, in our communication model, the parameter server can receive data from only one worker at a time, it took
$3(t_{\text{latency}} + t_{\text{transfer}})$ for the aggregation phase to finish.
In the broadcast phase, because, in our communication model, the
parameter server can send data to only one worker at a time,
it took $3(t_{\text{latency}} + t_{\text{transfer}})$ for the broadcast phase to
finish. 

In general, when there are $N$ workers and $1$ parameter server,
{\em as under our communication model}, a parameter server
architecture in which all workers are {\em perfectly synchronized}
takes 
\[
2N(t_{\text{latency}} + t_{\text{transfer}})
\]
to compute and broadcast the sum $S$ over all local copies $\{w_i\}$.

\paragraph{Discussions} As we see, the communication cost of
the parameter server architecture grows linearly with respect
to the total number of workers we have in the system. As a result,
this architecture could be sensitive to both latency and 
transfer time. This motivates some system relaxations that could
alleviate potential system bottlenecks.
\begin{enumerate}
\item When the network has small latency $t_{\text{latency}}$ compared with
the transfer time $t_{\text{transfer}}$, one could conduct lossy compression
(e.g., via quantization, sparsification, or both) to decrease the
transfer time. Usually, this approach can lead to a linear speedup 
with respect to the compression rate, up to a point that 
$t_{\text{latency}}$ starts to dominate.
\item When the network has large latency $t_{\text{latency}}$, compression
on its own won't be the solution. In this case, one could adopt
a decentralized communication pattern, as we will discuss later
in this work. 
\end{enumerate}

\subsection{AllReduce}

Calculating the sum over distributed workers is a very common operator
used in distributed computing and high performance computing systems. 
In many communication frameworks, it can be achieved using the \texttt{AllReduce}
operator. Optimizing and implementing the \texttt{AllReduce} operator has
been studied by the HPC community for decades, and the implementation is
usually different for different numbers of machines, different sizes of messages, and different physical communication topologies.

In this work, we focus on the simplest case, in which all workers
form a {\em logical} ring and communicate only to their neighbors
(all communications still go through the single switch all workers
are connected to). We also assume that the local parameter vector
is large enough.

Under these assumptions, we can implement an \texttt{AllReduce} operator
in the following way. Each worker $w_n$ partitions their local parameter
vectors into $N$ partitions ($N$ is the number of workers): $w_n^{k}$ is the $k$th partition of the local model $w_n$. 
The communication happens in two phases:
\begin{enumerate}
\item {\bf Phase 1.}  At
the first iteration of Phase 1, each machine $n$ sends $w_n^{n}$ to its ``next''
worker in the logical ring, i.e., $w_j$ where $j = n + 1 \mod N$.
Once machine $j$ receives a partition $k$, it sums up the received
partition with its local partition, and sends the aggregated partition 
to the next worker in the next iteration. After $N-1$ communication iterations,
different workers now have the sum of different partitions. 
\item {\bf Phase 2.} Phase 2 is similar to Phase 1, with the difference
that when machine $n$ receives a partition $k$, it replaces
its local copy with the received partition, and passes it onto the next 
machine in the next iteration.
\end{enumerate}

At the end of communication, all workers have the sum $S$ of all partitions.

\begin{example} We walk through an example with four workers $M_1$, ... $M_4$. 
The communication pattern of the above implementation is as follows.
{\rv We use \texttt{w\{Mi,j\}} to denote the $j$th partition on 
machine \texttt{Mi}.}

\begin{small}
\begin{verbatim}
# For the first partition w{M1 (worker id), 1 (partition id)}
Time= 0   M1 sends w{M1,1}                               to M2
Time= t   M2 sends w{M1,1} + w{M2,1}                     to M3
Time=2t   M3 sends w{M1,1} + w{M2,1} + w{M3,1}           to M4
Time=3t   M4 sends w{M1,1} + w{M2,1} + w{M3,1} + w{M4,1} to M1  
Time=4t   M1 sends w{M1,1} + w{M2,1} + w{M3,1} + w{M4,1} to M2  
Time=5t   M2 sends w{M1,1} + w{M2,1} + w{M3,1} + w{M4,1} to M3

# For the second partition w{M2 (worker id), 2 (partition id)}
Time= 0   M2 sends w{M2,2}                               to M3
Time= t   M3 sends w{M2,2} + w{M3,2}                     to M4
Time=2t   M4 sends w{M2,2} + w{M3,2} + w{M4,2}           to M1
Time=3t   M1 sends w{M2,2} + w{M3,2} + w{M4,2} + w{M1,2} to M2  
Time=4t   M2 sends w{M2,2} + w{M3,2} + w{M4,2} + w{M1,2} to M3
Time=5t   M3 sends w{M2,2} + w{M3,2} + w{M4,2} + w{M1,2} to M4 

# For the third partition w{M3 (worker id), 3 (partition id)}
Time= 0   M3 sends w{M3,3}                               to M4
Time= t   M4 sends w{M3,3} + w{M4,3}                     to M1
Time=2t   M1 sends w{M3,3} + w{M4,3} + w{M1,3}           to M2
Time=3t   M2 sends w{M3,3} + w{M4,3} + w{M1,3} + w{M2,3} to M3  
Time=4t   M3 sends w{M3,3} + w{M4,3} + w{M1,3} + w{M2,3} to M4 
Time=5t   M4 sends w{M3,3} + w{M4,3} + w{M1,3} + w{M2,3} to M1 

# For the fourth partition w{M4 (worker id), 4 (partition id)}
Time= 0   M4 sends w{M4,4}                               to M1
Time= t   M1 sends w{M4,4} + w{M1,4}                     to M2
Time=2t   M2 sends w{M4,4} + w{M1,4} + w{M2,4}           to M3
Time=3t   M3 sends w{M4,4} + w{M1,4} + w{M2,4} + w{M3,4} to M4  
Time=4t   M4 sends w{M4,4} + w{M1,4} + w{M2,4} + w{M3,4} to M1
Time=5t   M1 sends w{M4,4} + w{M1,4} + w{M2,4} + w{M3,4} to M2 
\end{verbatim}
\end{small}
From the above pattern, it is not hard to see why, at the end,
each worker has a copy of $S = \sum_{n=1}^N w_n$. 
\end{example}

\begin{figure}[t!]
	\centering
	\includegraphics[width=1.0\textwidth]{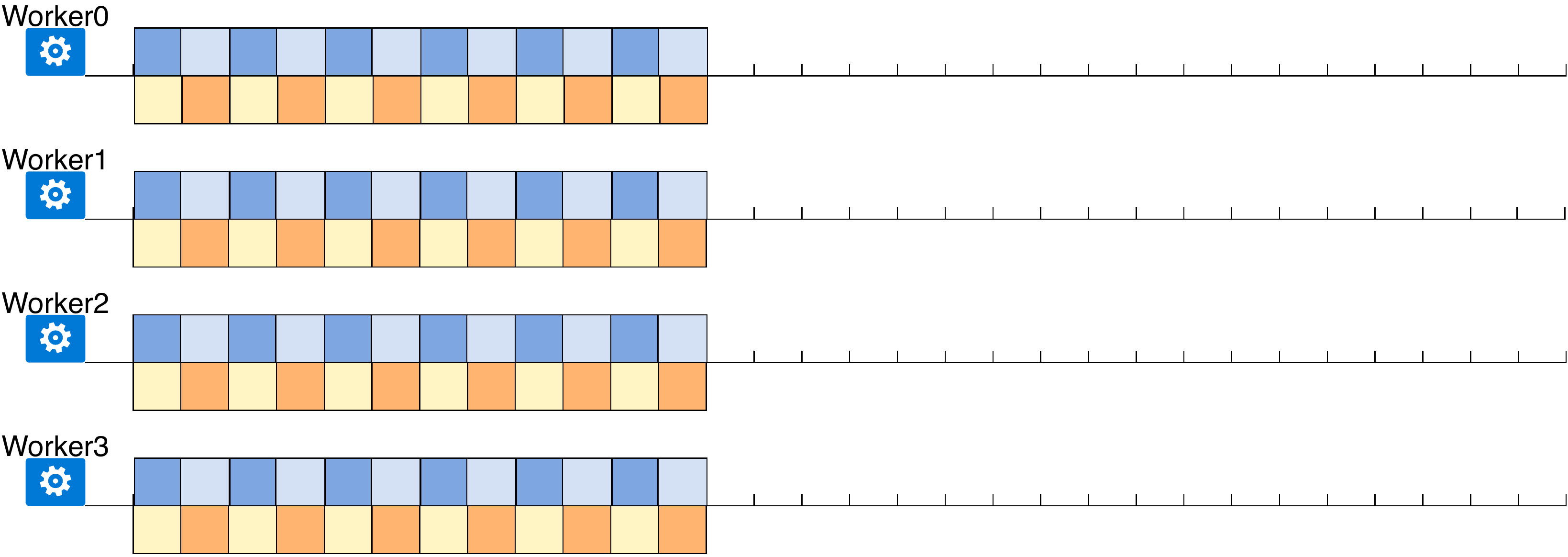}
	\caption{Illustration of the communication
		pattern of the AllReduce architecture with ring topology.}
	\label{fig:communication_illustration_timeline_allreduce}
\end{figure}

One interesting property of the above way of implementing the
\texttt{AllReduce} operator is that, at any timestep, each machine
concurrently sends and receives one partition of the data, which is
possible in our communication model. Figure~\ref{fig:communication_illustration_timeline_allreduce}
illustrates the communication timeline.

We make multiple observations.

\begin{enumerate}
\item Compared with a parameter server architecture with a single parameter server
(Figure~\ref{fig:communication_illustration_timeline_ps1}), the total
amount of data that each worker sends and receives is the same
in both cases --- in both cases, the amount of data sent and received
by each machine is equal to the size of the parameter vector.
\item At any given time, each worker sends and receives data concurrently.
At any given time, only the left neighbor $w_n$ sends data to $w_{n+1 \mod N}$
and $w_{n+1 \mod N}$ only sends data to $w_{n+2 \mod N}$. This allows the
system to take advantage of the {\em aggregated} bandwidth of $N$ machines
(which grows linearly with respect to $N$) instead of being bounded
by the bandwidth of a single central parameter server.
\end{enumerate}

In general, when there are $N+1$ workers,
{\em as under our communication model} and assuming that
the computation cost to sum up parameter vectors
is negligible, an \texttt{AllReduce} operator
in which all workers are {\em perfectly synchronized}
took 
\[
2N t_{\text{latency}} + 2 t_{\text{transfer}}
\]
to compute and broadcast the sum $S$ over all local copies $\{w_n\}$.

\paragraph{Discussions} As we see, the latency of
an \texttt{AllReduce} operator grows linearly with respect
to the total number of workers we have in the system. As a result,
this architecture could be sensitive to network latency.
This motivates some system relaxations that could
alleviate potential system bottlenecks.
\begin{enumerate}
	\item When the network has large latency $t_{\text{latency}}$, compression
	on its own won't be the solution. In this case, one could adopt
	a decentralized communication pattern, as we will discuss later
	in this work. 
	\item When the network has small latency $t_{\text{latency}}$ and the
	parameter vector is very large, the transfer time $t_{\text{transfer}}$ can still
	become the bottleneck. In this case, one could conduct lossy compression
(e.g., via quantization, sparsification, or both) to decrease the
transfer time. Usually, this approach can lead to a linear speedup 
with respect to the compression rate, up to a point that 
$t_{\text{latency}}$ starts to dominate.
\end{enumerate}

\paragraph{Caveats} We will discuss the case of asynchronous communication 
later in this work. Although it is quite natural to come up with an
asynchronous parameter server architecture, making the \texttt{AllReduce}
operator run in an asynchronous fashion is less natural. As a result,
when there are stragglers in the system (e.g., one worker is significantly slower than
all other workers), \texttt{AllReduce} can make it more difficult 
to implement a straggler avoidance strategy
if one simply 
uses the off-the-shelf implementation.

\paragraph{Why Do We Partition the Parameter Vector?} One interesting design choice
in implementing the \texttt{AllReduce} operator is setting each local
parameter vector to be partitioned into $N$ partitions. This decision is important if you want
to fully take advantage of the aggregated bandwidth of {\em all} workers.
Take the same four-worker example and assume that we do not partition the model.
In this case, the series of communication events will look like this:
\begin{verbatim}
Time= 0   M1 sends w{M1}                         to M2
Time= t   M2 sends w{M1} + w{M2}                 to M3
Time=2t   M3 sends w{M1} + w{M2} + w{M3}         to M4
Time=3t   M4 sends w{M1} + w{M2} + w{M3} + w{M4} to M1  
Time=4t   M1 sends w{M1} + w{M2} + w{M3} + w{M4} to M2  
Time=5t   M2 sends w{M1} + w{M2} + w{M3} + w{M4} to M3
\end{verbatim}
In general, with $N+1$ workers, the communication cost without
partitioning becomes
\[
2N (t_{\text{latency}} + t_{\text{transfer}}).
\]
Comparing this with the $2N t_{\text{latency}} + 2 t_{\text{transfer}}$
cost of \texttt{AllReduce} with model partition, we see
that model partition is the key reason for
taking advantage of the full aggregated
bandwidth provided by all machines.

\subsection{Multi-machine Parameter Server}

One can extend the single-server parameter server architecture and
use multiple machines serving as parameter servers instead. In this 
work, we focus on the scenario in which each worker
also serves as a parameter server.

Under this assumption, we can implement a multi-server parameter 
server architecture in the following way. 
Each worker $w_n$ partitions their local parameter vectors 
into $N$ partitions ($N$ is the number of workers): $w_n^{k}$. 
The communication happens in two phases:
\begin{enumerate}
\item Phase 1: All workers send their $n^{th}$ partition to worker $w_n$. Worker $w_n$ aggregates
all messages and calculates the $n^{th}$ partition of the sum $S$.
\item Phase 2: Worker $w_n$ sends the $n^{th}$ partition of the sum $S$ to all other workers.
\end{enumerate}

With careful arrangement of communication events, we can also take advantage of the full
aggregated bandwidth in this architecture, as illustrated in the following example.

\begin{example}
We walk through an example with four workers $M_1$, ... $M_4$. 
The communication pattern of the above implementation is as follows:
\begin{verbatim}
# First partition: w{M1 (worker id), 1 (partition id)}
# First partition of the result: S{1}
Time= 0   M2 sends w{M2,1} to M1
Time= t   M3 sends w{M3,1} to M1
Time=2t   M4 sends w{M4,1} to M1 
Time=3t   M1 sends S{1}    to M2
Time=4t   M1 sends S{1}    to M3
Time=5t   M1 sends S{1}    to M4

# Second partition: w{M2 (worker id), 2 (partition id)}
# Second partition of the result: S{2}
Time= 0   M1 sends w{M1,2} to M2
Time= t   M4 sends w{M4,2} to M2
Time=2t   M3 sends w{M3,2} to M2 
Time=3t   M2 sends S{2}    to M3
Time=4t   M2 sends S{2}    to M4
Time=5t   M2 sends S{2}    to M1

# Third partition w{M3 (worker id), 3 (partition id)}
# Third partition of the result: S{3}
Time= 0   M4 sends w{M4,3} to M3
Time= t   M1 sends w{M1,3} to M3
Time=2t   M2 sends w{M2,3} to M3 
Time=3t   M3 sends S{3}    to M4
Time=4t   M3 sends S{3}    to M1
Time=5t   M3 sends S{3}    to M2

# Fourth partition w{M4 (worker id), 4 (partition id)}
# Fourth partition of the result: S{4}
Time= 0   M3 sends w{M3,4} to M4
Time= t   M2 sends w{M2,4} to M4
Time=2t   M1 sends w{M1,4} to M4 
Time=3t   M4 sends S{4}    to M1
Time=4t   M4 sends S{4}    to M2
Time=5t   M4 sends S{4}    to M3
\end{verbatim}
For the above communication events, it is not hard to see that, 
at the end, each machine has access to the sum $S=\sum_{n=1}^N{w_n}$.
In terms of the communication pattern, under our communication model, the 
multi-server parameter server architecture has the same pattern as
\texttt{AllReduce}, illustrated in 
Figure~\ref{fig:communication_illustration_timeline_allreduce}.
\end{example}

In general, when there are $N+1$ workers,
{\em as under our communication model} and assuming that
the computation cost to sum up parameter vectors
is negligible, a multi-server parameter server
architecture
in which all workers are {\em perfectly synchronized}
took 
\[
2N t_{\text{latency}} + 2 t_{\text{transfer}}
\]
to compute and broadcast the sum $S$ over all local copies $\{w_n\}$.

\chapter{Distributed Stochastic Gradient Descent}

The previous \part provides us with the 
background of stochastic 
gradient descent and a simple communication model. This allows us to start analyzing the performance of
a simple, distributed stochastic gradient descent 
system, which will serve as the baseline
for the remaining part of this work.

\section{A Simplified Performance Model for Distributed Synchronous Data-Parallel SGD}

Recall the optimization problem that we hope
to solve:
\begin{align}
\min_{\x \in \mathbb{R}^d}\quad \left\{f(\x):={1\over M}\sum_{m=1}^M F_m(\x)\right\}
\label{eq:obj-gd-recall}
\end{align}

The stochastic gradient descent algorithm works
by sampling, uniformly randomly with
replacement, a term $m_t \in [M]$, and updating the
current model $\x_t$ with
\[
\x_{t+1} = \x_{t} - \gamma F'_{m_t}(\x_t)
\]
until convergence.

\begin{figure}[t!]
	\centering
	\includegraphics[width=1.0\textwidth]{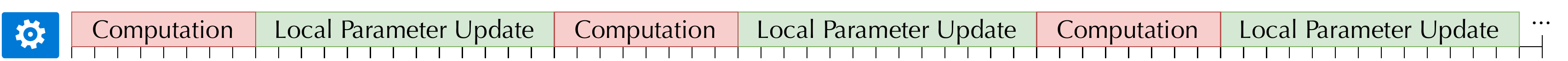}
	\caption{Illustration of SGD on a single machine.}
	\label{fig:pattern_SGD}
\end{figure}

\begin{figure}[t!]
	\centering
	\includegraphics[width=1.0\textwidth]{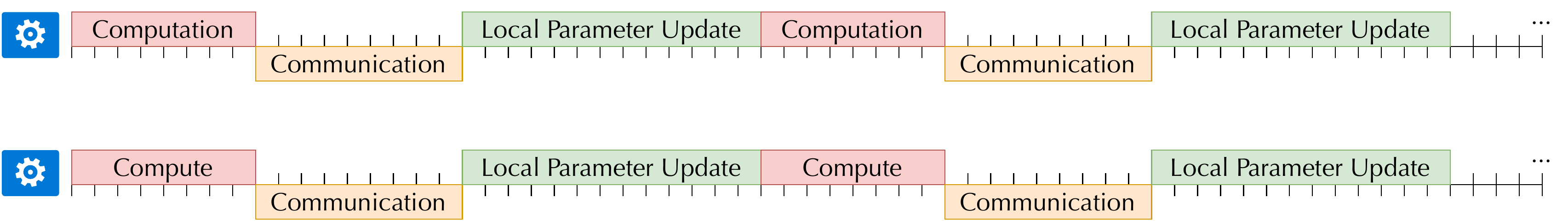}
	\caption{Illustration of the distributed synchronous data-parallel SGD.}
	\label{fig:pattern_DSGD}
\end{figure}

\subsection{SGD on a Single Machine}

Implementing SGD on a single machine, with
a single thread, is easy. The system 
stores the current model $\x_t$ in memory,
and repeats two stages, as illustrated
in Figure~\ref{fig:pattern_SGD}:
\begin{enumerate}
\item Computation: The system (1) fetches
$\x_t$ from the main memory; (2) fetches 
all information necessary to compute 
$F_{m_t}'(\x_t)$; and (3) computes
$F_{m_t}'(\x_t)$.
\item Update: The system updates
$\x_t$ with $F_{m_t}'(\x_t)$.
\end{enumerate}

\begin{figure}
\includegraphics[width=1.0\textwidth]{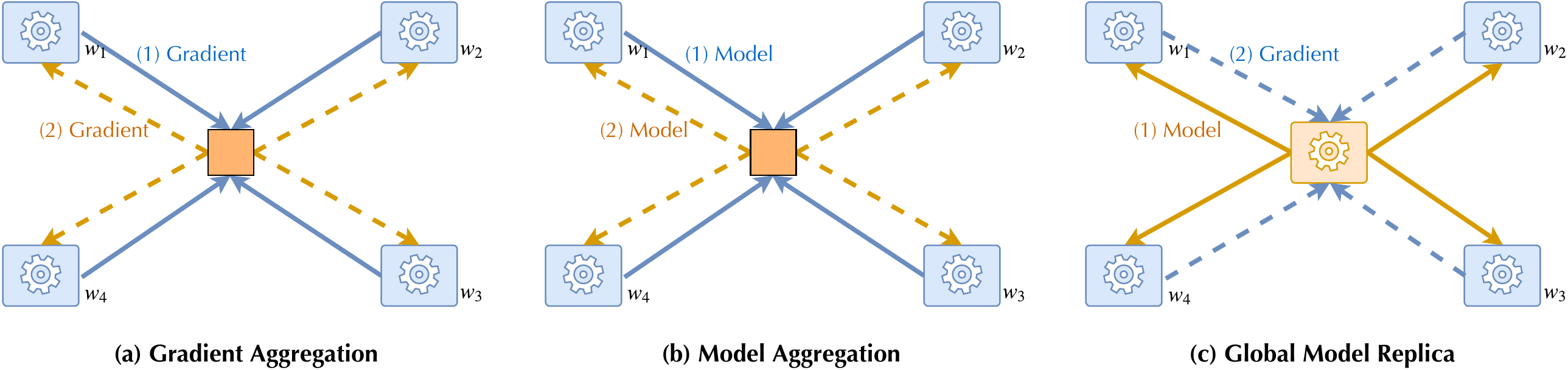}
\caption{Illustration of three possible implementations of data-parallel SGD
on multiple devices.}
\label{fig:threeimpl}
\end{figure}

\subsection{Data-Parallel SGD on Multiple Devices}

When distributing the above algorithm 
on multiple devices, there are multiple ways
of distributing the workload for different
system bottlenecks. For example,
when the model $\x_t$ is too large and does not
fit into the fast memory of a single device,
it can be partitioned onto different devices
for the computation phase. This strategy is
called {\em model parallelism}. In this work,
we focus on what is called {\em data parallelism} ---
each device has access to a partition of the
data set (i.e., $[M]$), and repeats a
three-stage process as illustrated in 
Figure~\ref{fig:pattern_DSGD}.

\begin{enumerate}
\item Computation: Each worker (say worker $n$)
samples, from its local partition, an
$F_{m_t^{(n)}}'(\x_t)$ to compute. $m_t^{(n)}$ denotes the index sampled by worker $n$ at iteration $t$.
\item Communication: Workers communicate
to compute the sum of all local gradients
$\sum_n F_{m_t^{(n)}}'(\x_t)$.
\item Update: The system updates $\x_t$
with $\sum_n F_{m_t^{(n)}}'(\x_t)$.
\end{enumerate}

The above algorithm can be implemented in three ways,
depending on whether the system aggregates the
gradient or the model, and the locality of 
model. Figure~\ref{fig:threeimpl} illustrates
these three implementations.

\subsubsection{Gradient Aggregation}

One simple strategy to implement the
above algorithm is for each worker 
$w_n$ to maintain a local model replica
$\x_t^{(n)}$. At the very beginning,
the replica on all workers is the same:
\[
\x_0^{(1)} = \cdots = \x_0^{(N)}.
\]
In the communication phase, the system 
aggregates the gradient using any one
of the three communication primitives 
we introduced before (e.g., \texttt{AllReduce}).
At the end of communication, each worker
sees the same aggregated gradient
\[
\sum_n F_{m_t^{(n)}}'(\x_t).
\]
In the update phase, each worker applies 
the update locally, independently. 
Because all workers see the same aggregated
gradient, their model replica stays equal after the updates.

\subsubsection{Model Aggregation}

It is possible to implement the system in a different way.
Each worker $w_n$ still maintains the
local replica $\x_t^{(n)}$ and makes sure they
are the same at the very beginning.

However, the system does the local update first
by applying the local gradient to each
local model:
\[
\x_{t+1/2}^{(n)} = \x_{t}^{(n)} - \gamma F_{m_t^{(n)}}'(\x_t^{(n)})
\]
In the communication phase, all workers
communicate the {\em model}, $x_{t+1/2}^{(n)}$
and calculate the sum
\[
\x_{t+1} = {1\over N}\sum_n \x_{t+1/2}^{(n)} = \x_{t} - \gamma {1\over N}\sum_n F_{m_t^{(n)}}'(\x_t^{(n)})
\]
The system then uses this sum $\x_{t+1}$ to update
its local model replica.

\subsubsection{Global Model Replica}

The third way of implementing the system is 
to maintain a {\em global}, instead of {\em local},
model replica. This global replica is
stored on one, or multiple, {\em parameter servers}.
At the very beginning, each worker 
$w_j$ fetches the global model replica $\x_t$
from the parameter servers, calculates the local
gradient, and sends the local gradient to 
the parameter servers. The parameter server
then calculates the sum of all local gradients,
and updates its global replica. 

\subsubsection{Discussion}

It is easy to see that all three implementations
logically implement the same algorithm. However,
the system tradeoff among these three different implementations can be quite delicate. 

\begin{enumerate}
\item {\bf Size of Model $=$ Size of Gradient}. 
When the size of the model is the same
as the size of the gradient, these three
implementations can have similar communication
cost. When we implement the gradient aggregation
and model aggregation approach using \texttt{AllReduce}
and the global model replica using a multi-server
parameter server, we see that all three implementations
have the communication cost
\[
2N t_{\text{latency}} + 2 t_{\text{transfer}}
\]
where the transfer time $t_{\text{transfer}}$ depends only
on the size of the model and the gradient.
In this case, we would expect the end-to-end
performance
of these three implementations to be 
similar.
\item {\bf Size of Model $\ne$ Size of Gradient}.
The tradeoff between these three implementations
becomes more complex when the size of the model
does not equal the size of the gradient.
For example, there could be a very dense 
model but a very sparse gradient, or vice versa. 
When this happens, these three implementations
can have very different performance. Specifically,
the communication cost can be summarized as follows:\footnote{This performance model assumes that
the communication cost does not 
change during the communication process. One example
that does not satisfy this assumption is when one
wants to use \texttt{AllReduce} to aggregate a set of 
{\em sparse} gradients~\citep{Alistarh2018-ct}. In this case,
the communication will
become denser and denser during the communication 
process (i.e., the sum of two sparse gradients
can only become denser).}
\begin{enumerate}
\item {\bf Gradient Aggregation (\texttt{AllReduce})}: 
\[
2N t_{\text{latency}} + 2 t_{\text{transfer}}^{(\text{grad})};
\]
\item {\bf Model Aggregation (\texttt{AllReduce})}: 
\[
2N t_{\text{latency}} + 2 t_{\text{transfer}}^{(\text{model})};
\]
\item {\bf Global Model Replica (Multi-server PS)}: 
\[
2N t_{\text{latency}} +  t_{\text{transfer}}^{(\text{model})} + t_{\text{transfer}}^{(\text{grad})}.
\]
\end{enumerate}
Obviously, different applications can have
different performance under these three implementations. 
\end{enumerate}

In this work, {\em we assume that 
the size of the model and the size of the gradient
are always the same.} Readers might
wonder why we bother to introduce all three
strategies, given that they all have similar performance
under this assumption. As we will
see later, different types of system relaxations
might be more suitable to be applied
to different implementations.
For example, asynchronous relaxation
might be easier to implement when there
is a global model replica. Moreover, 
some system relaxations might require different 
theoretical analysis under different 
implementations. For example, the 
lossy quantization technique applied to 
gradient aggregation and model aggregation 
lead to different convergence behavior;
decentralized relaxation might not
work at all if one uses gradient aggregation.

In this work, when introducing different
system relaxations, we assume that we have
the luxury of 
choosing one from these three implementations
without worrying about the delicate tradeoff
between them. In practice,
given a specific type of ML models, choosing
the right implementation and system relaxation
is often an engaged, task-specific problem.

\section{Theoretical Analysis}
The theoretical analysis is very similar to the $\mSGD$ we introduced in Section~\ref{Chap:sec:msgd}. One can simply imagine that the minibatch stochastic gradient in \eqref{eq:mSGD_mb} is computed by $N$ workers and that the minibatch size is $N=B$. Then, the convergence rate can be easily obtained from \eqref{eq:mSGD_cr}
\begin{align}
{1\over T} \sum_{t=1}^T \E[\|f'(\x_t)\|^2] \lesssim {L\over T} + {\sqrt{L}\sigma \over \sqrt{TN}}. \label{eq:pSGD}
\end{align}

%{\rc need to discuss with Ce about the data distribution.4}
%
%\textcolor{blue}{CZ: now I change the text to say each worker
%owns a partition of the data.}

\section{Caveats}

These are multiple aspects of data-parallel 
SGD implementations that we will not consider
in this work. These aspects are often 
critical in practice to achieving good
performance; however, including them into theoretical analysis is either 
impossible (because of the lack of an analytical
understanding) or more engaged (because of
the sophisticated performance model introduced
by these factors). The goal of this work is
{\em not} to cover all aspects of distributed learning; instead, providing a minimalist 
tutorial of the basics. As a result, we
briefly discuss these aspects.
We refer readers to \part~\ref{sec:further_readings}
for further reading about these topics.

\begin{enumerate}

\item {\bf Batch size.} One aspect that
we do not consider in our analysis is the 
impact of batch size on model accuracy.
The underlying assumption of this work
is that using a larger batch size, once
converged, will lead to the same
model accuracy as a smaller batch size. This 
assumption allows us to treat the impact 
of larger batch size as a way to 
reduce the variance of the stochastic gradient.
However, in practice, especially for models
such as deep neural networks, the impact of
batch size can be quite delicate and could
also have an impact on accuracy and generalization.

\item {\bf Hardware parallelism.} Another 
assumption that we are making on the system 
side is that the time that one device needs to
compute stochastic gradient over {\em one}
data point is $K$ times cheaper than
computing stochastic gradient over a 
batch of $K$ data points. This is true
in terms of the number of floating point
operations that one needs to conduct; however, it
might not necessarily be true with 
real-world hardware in terms of wall-clock
time. For hardware, especially those
optimizing throughput such as GPUs, 
batching might have a significant impact 
on performance. In some cases, calculating 
a single data point might be as expensive
as processing a batch of data points, in
terms of wall-clock time.
This aspect, together with the impact of
batch size on model accuracy, could make it
quite complicated to design distributed
learning systems. 

\item {\bf Model structure.} In this work,
we assume that the model ($\x_t$) is a dense
vector that does not have much structure.
In practice, the model might have more structures.
For example, if $\x_t$ corresponds to a 
deep neural network, it can be decomposed into
multiple layers. If one uses the standard
back-propagation algorithm, the communication
of layer can be overlapped with the computation
(e.g., one can calculate the gradient of
layer $l$ when communicating for layer $l+1$
in the backward pass). Recent research has tried to take advantage of
this structure (See \part~\ref{sec:further_readings}). 
As another example, when the data or model has
some sparsity structure, one could also take
advantage of it during (distributed)
training.

\item {\bf Communication Model.} Last but 
not least, we assume that communication 
via the ``logical switch'' is the
only way that different devices
can communicate. However, in practice, 
modern hardware is more complicated than this ---
different CPU cores might ``communicate''
via the shared L3 cache; different
CPU sockets might ``communicate'' via
QPI; a GPU/CPU hybrid system has even
more delicate ways of communicating. 
Designing a distributed learning system
when each of the workers is a 
multi-socket CPU system together with
multiple GPU devices is more 
complicated
than the simple performance model that 
we cover in this work.

\end{enumerate}

\chapter{System Relaxation 1: Lossy Communication Compression}
 
In distributed systems, 
data movement can often be
significantly slower than computation. This is
also often true
for distributed learning --- a forward and
backward pass to calculate the gradient on
a 152-layer ResNet requires roughly
3 $\times$ 11 GFLOPS and 
has a 230 MB model. In {\em theory}, with
16 GPUs, each of
which provides 10 TFLOPS, and
a 40 Gbit network, a 256-image batch
would require 0.05 seconds for gradient calculation, but would require 0.09 seconds for 
network transfer time. 
In practice, both processes are often slower
than the theoretical peak; however, 
their relative order remains similar.

In this \part, we focus on one system
relaxation technique to speed up the expensive
gradient exchange step in distributed
SGD --- instead of exchanging the gradient
as 32bit floating point numbers, the system
first quantizes it into a lower precision
representation before communication.
The impact of this relaxation is to 
decrease the time needed to transfer
a model --- if one quantizes the 
gradient into 8bit fixed point representation,
the transfer time would {\em theoretically} decrease by 4$\times$.

\section{System Implementation}

To implement this system relaxation, two aspects of the process need to be addressed. 
\begin{enumerate}
\item How can a 
floating point vector $\x$, representing either
the model or the gradient, be compressed,
using a function $Q(\cdot)$ such as the
result $Q({\rv \x})$, which can be stored more efficiently
(either more sparsely than $\x$ via sparsification,
or using fewer bits via quantization)?
\item How can the lossy compression function 
$Q(-)$ be used to implement distributed
SGD? As we will see later, compressing different
communication channels for different implementations
of distributed SGD (i.e., model aggregation, 
gradient aggregation, and a global model replica)
actually leads to different algorithms and needs
different analysis of convergence.
\end{enumerate}

\subsection{Lossy Communication Compression}

There are multiple ways of conducting 
lossy compression that are popularly used 
for distributed SGD. In this work, we focus
on {\em unbiased
lossy compression}, which ensures that
\[
\mathbb{E}_{\xi}\left[Q({\rv \x}; \xi) \right] = {\rv \x}.
\]
The term $\xi$ denotes the randomness of compression, and the
above equation ensures that the compressed 
vector $Q({\rv \x}; \xi)$ is 
 {\em unbiased} with respect to the
original vector $x$.

\begin{figure}[t!]
\includegraphics[width=1.0\textwidth]{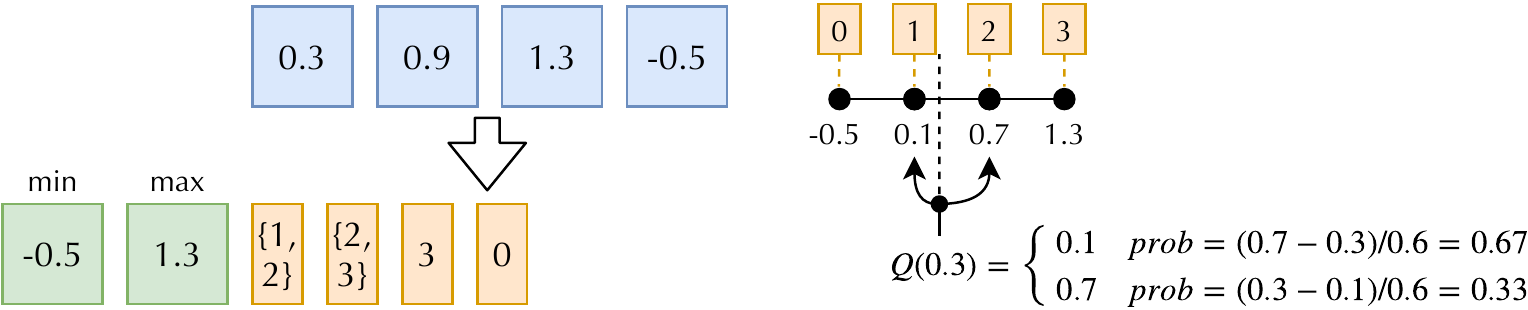}
\caption{Illustration of One Simple Quantization
Method.}
\label{fig:quantization}
\end{figure}

\paragraph*{Unbiased Compression via Quantization}
There are multiple ways of conducting unbiased
compression, and quantization is arguably the 
simplest approach. Figure~\ref{fig:quantization}
illustrates this process.

Given an original input vector $\x$, stored as 
floating points (e.g., 32 bits per element),
let $\min(\x)$ and $\max(\x)$ be the
smallest and the largest element of $\x$.
A simple way of quantizing the input vector
using $b$-bits per element is simply to 
quantize each element of $\x$, $\x_i$ independently.
With $b$-bits, one can represent $2^b$
``{\em knobs}'' partitioning the range
$[\min(\x), \max(\x)]$. Assume these
knobs are uniformly located as 
$\{c_0,...,c_{2^b - 1}\}$, i.e.,
\[
c_{i} = i \times \frac{\max(\x) - \min(\x)}{2^b - 1} + \min(\x).
\]
Assume that the $i^{th}$ element of $\x$,
$\x_i$ falls between $[c_{i}, c_{i+1})$;
we can use the following {\rv randomized quantization (RQ)} function:
\begin{align}
Q(\x_i; \xi_i) = \left \{
\begin{array}
cc_{i}  ~~~~~~~ \xi > \frac{\x_i - c_i}{c_{i+1} - c{i}}\\
c_{i+1}~~~~~ \text{otherwise}
\end{array}
\right.
\label{eq:RQ}
\end{align}
where $\xi_i$ is a uniform random variable 
on $[0, 1]$.

Figure~\ref{fig:quantization} illustrates
this process for $b = 2$ (i.e., 2-bit per element)
and $\min(\x) = -0.5, \max(\x) = 1.3$.
Take the first element $0.3$, for
example: it falls in the internal $[0.1, 0.7)$.
As a result, with probability $\frac{2}{3}$ the
system quantizes the element 0.3 to 0.1
and with probability $\frac{1}{3}$, the
system quantizes the element 0.3 to 0.7.
If one calculates the expectation, we have
\[
\frac{2}{3} \times 0.1 + \frac{1}{3} \times 0.7 = 0.3
\]

Because there are only $2^b$ knobs that
the quantization function $Q(-)$ can take
values in, the output can be encoded using
$b$ bits per element.

\paragraph*{Caveats} The above approach is
one of the simplest ways to construct an
unbiased compression function. There are 
multiple ways that it can be improved.
We briefly summarize these caveats below and refer
readers to \part~\ref{sec:further_readings}
for further readings.

\begin{enumerate}
\item In practice, more
sophisticated quantization functions can be constructed. For example,
instead of normalizing the vector using $l_\infty$
norm (i.e., $\min(\x), \max(\x)$), one can
normalize it using other norms, say $l_2$. 
One can also partition the vectors into different
``buckets'' and quantize each bucket 
independently. Moreover, the ``knobs'' do not
need to be uniformly distributed over 
$[\min(\x), \max(\x)]$, and one can even learn 
those knobs as part of the learning process.
All these approaches have been used 
for communication quantization and
can sometimes outperform the baseline
quantization strategy that we just introduced.
\item If the original input vector is stored
as 32-bit floating point numbers, quantization 
can only provide at most 32$\times$ compression,
because one cannot compress each element below
a single bit. There are other approaches that
do not have this limitation. One such approach
is called {\em sparsification}, which maps
the input vector $x$ to another vector $x'$ that
is more sparse. This mapping can also be made 
to be unbiased, a strategy that has
been popular in practice.
\item In this work, we focus on the scenario
in which the lossy compression scheme is {\em unbiased}. However, a
{\em biased} compression scheme can still be used while still achieving
convergence of the training process. For example,
a vector $\x$ can be compressed by taking
the top-$k$ element. This operation is clearly 
biased, but there is recent work proving convergence using this type of biased compression scheme.
\end{enumerate}

\subsection{SGD with Lossy Communication Compression ($\CSGD$)}

\begin{figure}
\includegraphics[width=1.0\textwidth]{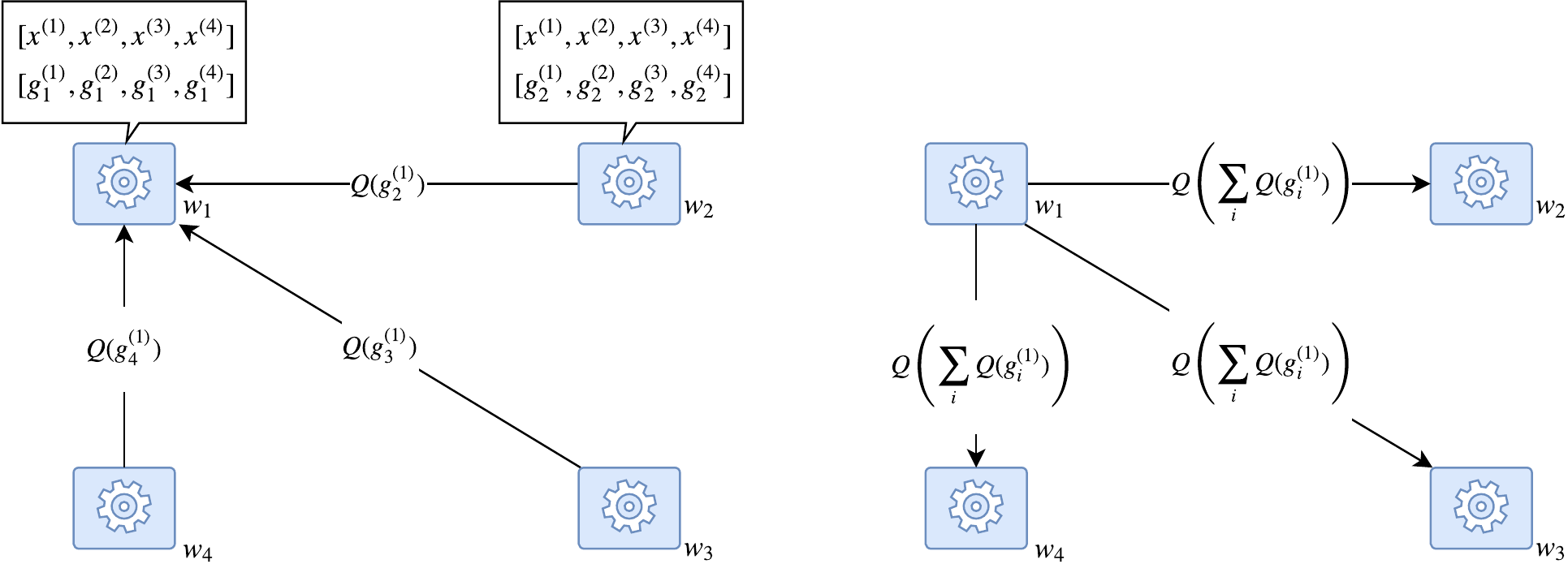}
\caption{Illustration of Multi-server Parameter
Servers with Lossy Communication Compression.
In this example,
there are four workers. Each worker is
the parameter server of one partition of
the model.
This figure illustrates the communication 
for the first partition of the model $x^{(1)}$
hosted by the first worker $w_1$.}
\label{fig:compression:ps}
\end{figure}

It is possible to use a quantization function $Q(-)$ 
to compress communications
for distributed training. In the previous \parts,
we described three different ways of
implementing distributed SGD and
different communication primitives; for each
of these implementations and primitives, compressing
the communication might actually lead to different
algorithms and convergence behaviors. 

\subsubsection{Multi-server Parameter Server
for Gradient Aggregation}

The easiest implementation for demonstrating the
impact of lossy compression is probably
gradient aggregation using a multi-server
parameter server. In this strategy, each machine
holds a replica of the full model. 
Given $N$ machines, the gradient vector is
partitioned into $N$ chunks --- each machine
$w_n$ is responsible for aggregating the 
$n^{th}$ partition, as illustrated in Figure~\ref{fig:compression:ps}.
%\footnote{\rc Ji's comment: Suggest using $N$ to denote the total number of workers and $n$ to denote the index.}

Specifically, all machines
send their $n^{th}$ partition of the local
gradient to the worker $w_n$, which 
aggregates all incoming vectors and broadcasts
back the sum. Both the
incoming and outcoming messages can be compressed by using
the quantization function. Let the 
local gradient on worker $n$ be $g_n$;
after aggregation, each worker receives
\begin{align}
Q\left({1\over N}\sum_{n=1}^N Q(g_n)\right).
\label{eq:Qg1}
\end{align}

\subsubsection{AllReduce for Gradient Aggregation}

\begin{figure}
\includegraphics[width=1.0\textwidth]{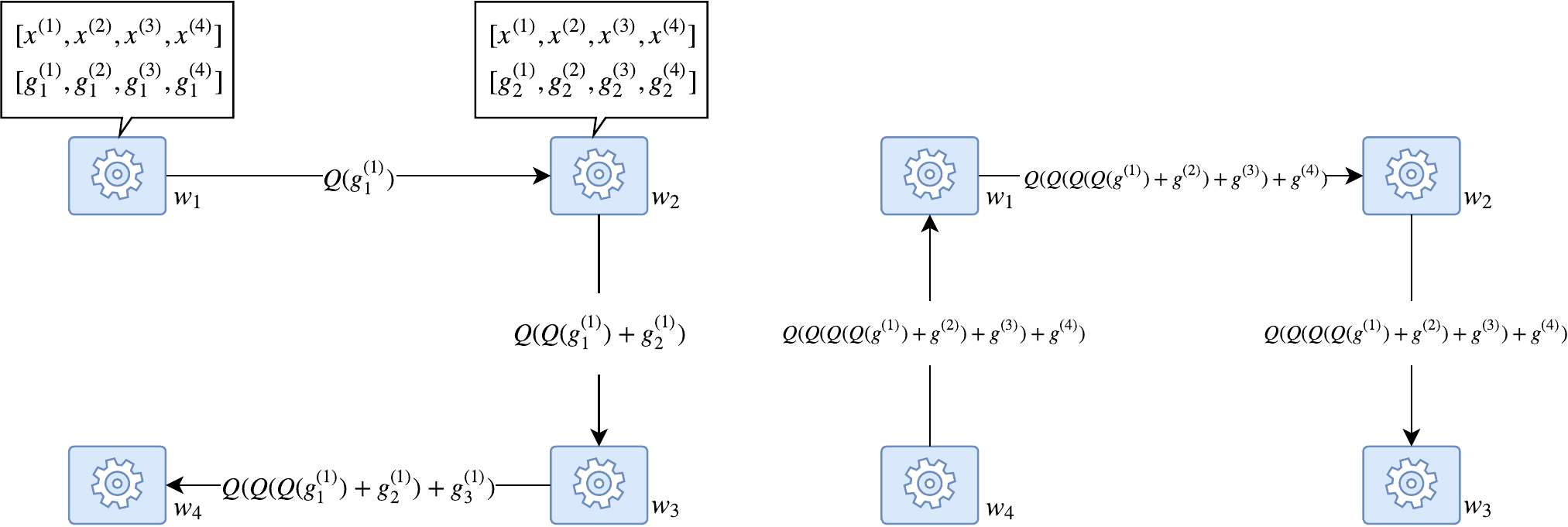}
\caption{Illustration of \texttt{AllReduce} with Lossy Communication Compression. In this example,
there are four workers. Each worker initiates
the communication of one partition of the model.
This figure illustrates the communication 
for the first partition of the model $x^{(1)}$
initiated by the first worker $w_1$.}
\label{fig:compression:allreduce}
\end{figure}

We can also apply lossy compression to \texttt{AllReduce},
which makes the system behavior more 
complicated to analyze. Figure~\ref{fig:compression:allreduce} illustrates the
communication pattern.

Specifically, given $n$ workers, the
system divides the local gradient into $n$
partitions. Each partition ``flows through''
a ring formed by all the machines, 
and keeps getting
aggregated with the local gradient
(Figure~\ref{fig:compression:allreduce}(left)).
At the end, one worker has the result of
the aggregation. In the second step,
the aggregated result ``flows through''
a ring of all the machines again --- whenever
a machine receives the aggregated result,
it makes a local copy, and sends the
aggregated result to the next machine in 
the ring (Figure~\ref{fig:compression:allreduce}(left)).

One caveat of this strategy is that, in order
to make sure all incoming and outcoming 
messages are compressed, the sum must be 
compressed $N$ times if there are $N$
machines in the ring. As a result,
as illustrated in Figure~\ref{fig:compression:allreduce}, after aggregation,
each worker receives
\begin{align}
Q(\cdots Q(Q(Q(Q(Q(g_1) + g_2) + g_3) + g_4) \cdots + g_N).
\label{eq:Qg2}
\end{align}

Compared with the aggregated result 
in the parameter server case ($
Q({1\over N}\sum_{n=1}^N Q(g_n))$), it is clear that
the \texttt{AllReduce} case requires more engaged analysis.

\subsubsection{Impact of Lossy Compression}

\begin{figure}
\includegraphics[width=1.0\textwidth]{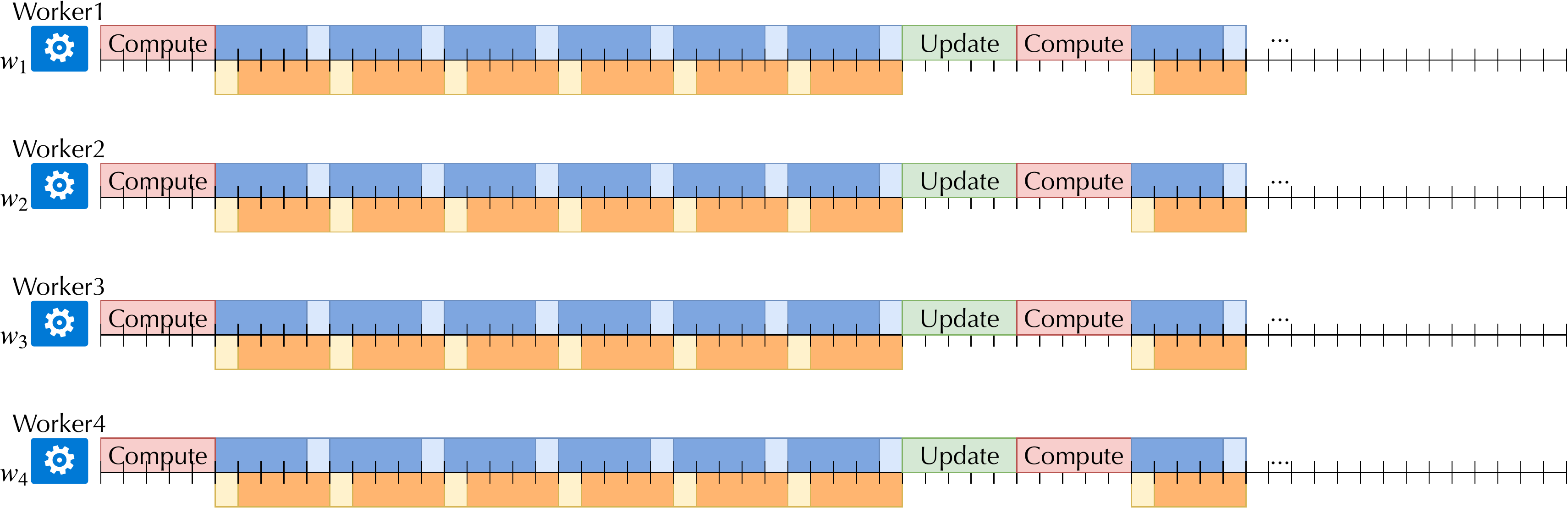}
\caption{Illustration of the Impact of Lossy 
Compression (Without Compression)}
\label{fig:compression:impact:orig0}
\end{figure}

\begin{figure}
\includegraphics[width=1.0\textwidth]{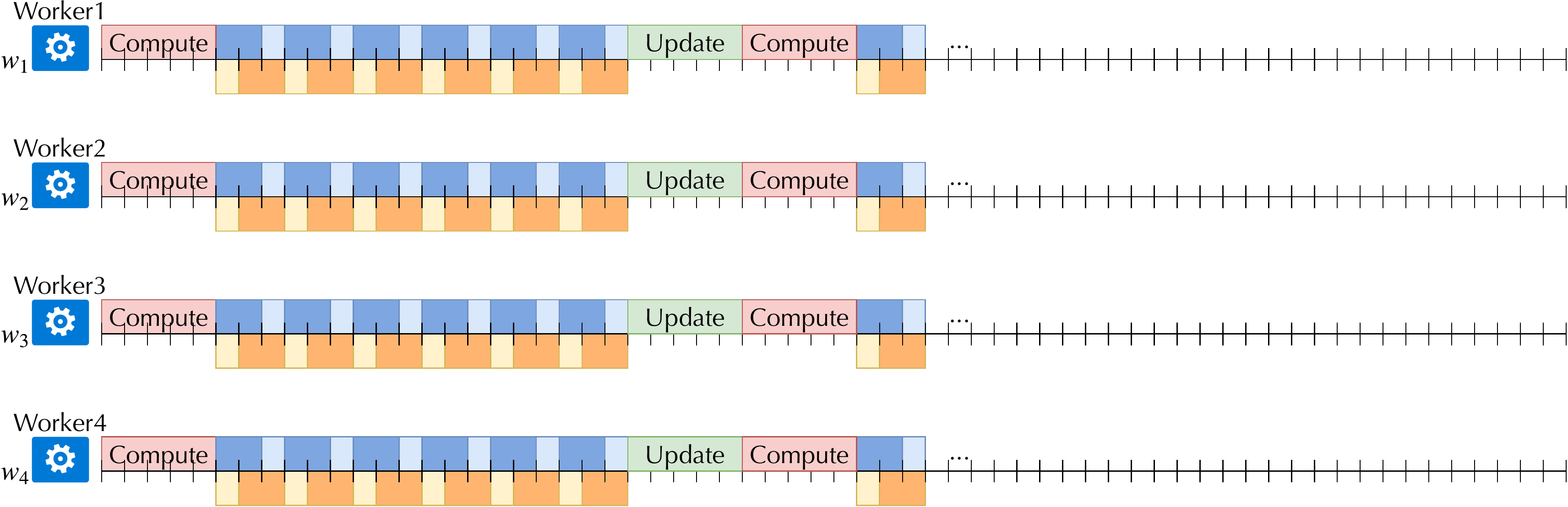}
\caption{Illustration of the Impact of Lossy 
Compression (With 2$\times$ Compression)}
\label{fig:compression:impact:orig}
\end{figure}

Both implementations have a similar impact
on the end-to-end performance. Figure~\ref{fig:compression:impact:orig0} and Figure~\ref{fig:compression:impact:orig} illustrate the
impact of a 2$\times$ compression. 

We see that lossy compression would decrease
the transfer time, as the communicated
messages are now smaller. It does not have
any impact on latency, as the number of
communications the system needs to
conduct stays the same. In this example,
we ignore the computation cost of 
conducting the compression, which is often
small compared with the computation time
of the gradient.

Specifically, for \texttt{AllReduce} and a multi-server
parameter server whose communication cost
is 
\[
2n t_{\text{latency}} + 2 t_{\text{transfer}},
\]
compressing the communication by $K$ times
leads to a communication cost of
\[
2n t_{\text{latency}} + \frac{2}{K} t_{\text{transfer}}.
\]
When system performance is bounded by
the communication cost, and the communication
cost is dominated by the transfer time,
we observe linear speedup with respect
to the compression ratio, in terms
of the time needed to finish a single
iteration.

\section{\rv Theoretical Analysis for $\CSGD$}

{\rv In this section, we analyze \CSGD,
an 
\SGD~ variant in which the stochastic gradient 
is compressed by some lossy compression scheme.
}
The analysis for the $\CSGD$ algorithm is very similar to the analysis for stochastic gradient based methods we have seen in previous \parts. First, we can see that the basic updating rule for $\CSGD$ is nothing but 
\begin{align}
\x_{t+1} = \x_t - \gamma \g_t,
\end{align} 
where $\g_t$ could take the form of either \eqref{eq:Qg1} or \eqref{eq:Qg2}. But it needs to admit the following two key assumptions, which essentially serve the same purpose as as Assumption~\ref{ass:sgd}:
%\begin{tcolorbox}[colback=pink!5!white,colframe=black!75!black]
%\begin{assumption} \label{ass:csgd}
%We make the following assumption:
\begin{itemize}
\item ({\bf Unbiased gradient}) The stochastic gradient is unbiased, that is, 
\[
\E[\g_t] = f'(\x_t);
\]
\item ({\bf Bounded stochastic variance}) The stochastic gradient has bounded variance, that is, there exists a constant $\sigma_{\text{c}}$ satisfying
\[
\E[\|\g_t - f'(\x_t)\|^2] \leq \sigma^2_{\text{c}}.
\]
\end{itemize}
%\end{assumption}
%\end{tcolorbox}
Following the same analysis procedure as for $\SGD$ {\rv in Section~\ref{sec:SGD:convergence}}, we can obtain the following convergence rate for $\CSGD$:
\begin{align}
\label{eq:csgd_convergence_0}
{1\over T}\sum_{t=1}^T\E[\|f'(\x_t)\|^2] \lesssim {L \over T} + {\sqrt{L}\sigma_{\text{c}} \over \sqrt{T}},
\end{align}
by choosing the learning rate $\gamma = {1\over L+\sigma_{\text{c}} \sqrt{TL}}$. The key difference is that the variance $\sigma_{\text{c}}$ is different from the $\sigma$ in Assumption~\ref{ass:sgd}. To take a closer look at $\sigma_{\text{c}}$, let us consider the form for $\g_t$ to be \eqref{eq:Qg1} and assume that each stochastic gradient $g_n$ is unbiased with bounded variance $\E[\|g_n - f'(\x_t)\|^2] \leq \sigma^2~\forall n$ and the compression is unbiased with bounded variance $\sigma'$, that is,

\begin{tcolorbox}[colback=pink!5!white,colframe=black!75!black]
\begin{assumption}\label{ass:unbias_compression}
{\bf (Unbiased compression)} The (probably randomized) compression operator $Q(\cdot)$ is assumed to be unbiased:
\[
\E[Q(\y)]= \y,\quad \forall \y.
\]
\end{assumption}
\end{tcolorbox}
\begin{tcolorbox}[colback=pink!5!white,colframe=black!75!black]
\begin{assumption}\label{ass:bounded_compression}
The (probably randomized) compression operator $Q(\cdot)$ is assumed to be bounded:
{\bf (Bounded compression)}
\[
\E \left[\left\|Q(\y) - \y\right\|^2 \right] \leq \sigma'^2,\quad \forall \y.
\]
\end{assumption}
\end{tcolorbox}
% \begin{equation}
%\begin{aligned}
%\E[Q(\y)]= &\y\quad \forall \y 
%\\
%\E \left[\left\|Q(\y) - \y\right\|^2 \right] \leq& \sigma'^2\quad \forall \y.
%\end{aligned}
%\label{ass:compression}
%\end{equation}
Then we can bound the variance for $\g_t$ by
\begin{align*}
& \E \left[\left\|\g_t - f'(\x_t) \right\|^2\right] 
\\ \leq & 
\E \left[ \left\|{1\over N}\sum_{n=1}^N Q(g_n) - f'(\x_t)\right\|^2 \right] +
\sigma'^{2}
\\ = &
{1\over N^2} \sum_{n=1}^N\E \left[ \left\|Q(g_n) - f'(\x_t)\right\|^2 \right] + \sigma'^{2}
\\ \leq &
{1\over N^2} \sum_{n=1}^N\left(\E \left[ \left\|Q(g_n) -g_n\right\|^2 \right] + \E \left[ \left\|g_n - f'(\x_t)\right\|^2 \right] \right) + \sigma'^{2}
\\ \leq &
\underbrace{{\sigma^2 \over N} + \left(1 + {1\over N} \right)\sigma'^2}_{=:\sigma_c^2}.
\end{align*}
Therefore, from~\eqref{eq:csgd_convergence_0} the convergence rate of $\CSGD$ can be summarized into
\begin{align}
{1\over T}\sum_{t=1}^T\E[\|f'(\x_t)\|^2] \lesssim {L \over T} + {\sqrt{L}\sigma \over \sqrt{NT}} + \underbrace{{\sqrt{L}\sigma' \over \sqrt{T}}}_{\text{caused by compression}}.
\label{eq:csgd_convergence}
\end{align}

Comparing this approach to the parallel $\SGD$'s convergence rate in~\eqref{eq:pSGD}, we can see that the last term is the addition caused by using compression. When the stochastic variance $\sigma^2 / N$ dominates the compression variance $\sigma'^2$, then the compression does not affect the convergence rate significantly. 

{\rv
The key to ensuring convergence is the unbiased assumption. One can verify that the randomized quantization compression strategy in \eqref{eq:RQ} satisfies this assumption. Randomized sparisification compression is another method satisfying the unbiased assumption:
\begin{itemize}
\item \textbf{Randomized Sparsification} \citep{Wangni2018-ux}: For any real number $z$, with probability $p$, set $z$ to $0$ and $\frac{z}{p}$ with probability $p$. This is also an unbiased compression operator. 
 \item \textbf{Randomized Quantization:} \citep{Alistarh2018-ct, Zhang2017-yx, wang2018atomo} For any real number $z \in [a, b]$ ($a$, $b$ are pre-designed low-bit numbers), with probability $\frac{b-z}{b-a}$, compress $p$ into $a$, and with probability $\frac{z-a}{b-a}$ compress $z$ into $b$. This compression operator is unbiased.
\end{itemize}
However, some popular compression methods do not really satisfy the requisite unbiased assumption, for example, 
 \begin{itemize}
 \item \textbf{1-Bit Quantization \citep{Bernstein2018-mv, Wen2017-ik}:} Compress a vector $\x$ into $\|\x\|\text{sign}(\x)$ or $x$, where $\text{sign}(\x)$ is a vector whose element takes the sign of the corresponding element in $\x$. This compression operator is biased.
 \item \textbf{Clipping:} For any real number $z$, directly set its lower $k$ bits to zero. For example, deterministically compress $1.23456$ into $1.2$ with its lower $4$ bits set to zero. This compression operator is biased.
 %\item \textbf{Top$-k$ Sparsification \citep{NIPS2018_7697}:} For any vector $\x$, compress $\x$ by retaining the top $k$ largest elements of this vector and setting the others to zero. This compression operator is biased.
 \end{itemize}
It is worth pointing that it is usually hard to ensure convergence using biased compression methods in theory, but they can still be used in practice and may yield positive results.

\section{Error Compensated Stochastic Gradient Descent ($\DS$) for Arbitrary Compression Strategies}
The unbiased compression assumption in Assumption~\ref{ass:unbias_compression} is relatively restrictive. To overcome this limitation, we introduce a very recent algorithm, called error-compensated stochastic gradient descent $\DS$ or \texttt{DoubleSqueeze} \citep{tang2019doublesqueeze}, which is compatible to any reasonable (potentially biased) compression methods.

To illustrate this algorithm, let us first formally define the objective:
\begin{align}
\min_{\x}:\quad \left\{f({\x}):={1\over N}\sum_{n=1}^N f_n(\x) \right\},
\label{eq:ecsgd_obj}
\end{align}
where $f_n({\x}):= \E_{\xi\sim \mathcal{D}_n} F_n(\x; \xi)$ and $\mathcal{D}_n$ denotes the distribution of local data at node $n$. We do not assume that all nodes can access the whole dataset. Apparently, this is a more general loss function than~\eqref{eq:obj-gd-recall}.

\paragraph{Algorithm description}
We now consider the parameter server architecture for simplicity. The key idea of $\DS$ is to record the error (or bias) caused by compression and compensate the error in the next round iteratively. More specifically, at the $t$th iteration, {\bf all workers} (indexed by $n$) compute their local gradients by
\begin{align}
    \v^{(n)}_t = &   F'_n(\x_t; \xi_t^{(n)}) + \bm{\delta}_{t-1}^{(n)}
    \label{eq:ecsgd_worker_v}
    \\
    \bm{\delta}^{(n)}_{t} = & \v_t^{(n)} - Q(\v_t^{(n)}),
    \label{eq:ecsgd_worker_delta}
\end{align}
where $  F'_n(\x_t; \xi_t^{(n)})$ is the local stochastic gradient, $\bm{\delta}_{t-1}^{(n)}$ is the compression error left by the previous iteration, $\v^{(n)}_t$ is the error-compensated gradient, and $\bm{\delta}_{t}^{(n)}$ is the new compression error. Note that $\x_t$ is the global model at the $t$th iteration, which is retrieved by each individual worker. To reduce the communication cost, all workers send the compressed error-compensated gradient $Q(\v_t^{(n)})$ to the parameter server. At the $t$th iteration, the {\bf parameter server} aggregates all received gradients plus the error $\bm{\delta}_t$ left in last iteration on the parameter server
\begin{align*}
    \v_t = \frac{1}{N}\sum_{n=1}^N Q\left(\v_t^{(n)}\right) + \bm{\delta}_{t-1}
    \numberthis \label{eq:ecsgd_v_t}
\end{align*}
and then sends the compressed $\v_t$ (that is, $Q(\v_t)$) to all workers recording the new error on the parameter server
\begin{align*}
    \bm{\delta}_{t} = \v_t - Q(\v_t).
    \numberthis \label{eq:ecsgd_delta_t}
\end{align*}
Note that the parameter server does not need to maintain a global model $\x_t$. Instead, the (virtual) global model $\x_{t}$ is retrieved by each worker through the received $Q(\v_t)$, that is,
\begin{align*}
\x_{t+1} = \x_t - \gamma Q(\v_t).
\numberthis \label{eq:ecsgd_x_t}
\end{align*}
One can see that all communicated information is compressed. 

\section{Theoretical Analysis for $\DS$}
A big advantage of the $\DS$ algorithm is that it can be compatible with any reasonable compression method and its convergence efficiency is quite robust to the compression. To understand the magic of $\DS$, we provide the essential mathematical iteration in the following lemma:
\begin{tcolorbox}[colback=yellow!5!white,colframe=black!75!black]
\begin{lemma}\label{lem:ec-sgd}
The $\DS$ algorithm with $N$ works by following the iteration rule shown below.
\begin{align}
%\tilde{\x}_{t+1} = \tilde{\x}_t - \gamma \tilde{\nabla}f(\x_t)
\tilde{\x}_{t+1} = \tilde{\x}_t - \gamma \frac{1}{N}\sum_{n=1}^N F'_n\left(\x_{t};\xi_t^{(n)} \right),
\label{eq:EC-SGD}
\end{align}
where
\begin{align*}
\tilde{\x}_t := & \x_t - \gamma \Omega_{t-1}
\\
%\tilde{\nabla}f(\x_t) := & f'(\x_t) - \Delta_t
%\\
    \Omega_t := & \bm{\delta}_t + \frac{1}{N} \sum_{n=1}^N \bm{\delta}_{t}^{(n)}.
    %\\
%\Delta_t := & \frac{1}{N}\sum_{n=1}^N \left(  f'(\x_t) -   F'_n\left(\x_{t};\Delta_t^{(n)}\right) \right).
\end{align*}
\end{lemma} 
\end{tcolorbox}
From \eqref{eq:EC-SGD} it is apparent that the $\DS$ algorithm essentially follows the principle of $\SGD$ or $\GD$, except that $\x_t$ gets a little bit of perturbation by $\Omega_t$, which aggregates all compression errors at round $t$.
\begin{proof}
This can be proved by straightforward linear algebra computation, by plugging \eqref{eq:ecsgd_v_t} into \eqref{eq:ecsgd_x_t}:
\begin{align*}
&{1\over \gamma}(\x_{t} - \x_{t+1})\\
= & Q\left(\bm{\delta}_{t-1} +\frac{1}{N} \sum_{n=1}^NQ\left(\v_t^{(n)}\right) \right) \\
= &  \bm{\delta}_{t-1} +\frac{1}{N} \sum_{n=1}^NQ\left(\v_t^{(n)}\right) - \bm{\delta}_t\quad\text{(from \eqref{eq:ecsgd_delta_t})}\\
%= &  \frac{1}{N} \sum_{n=1}^NQ\left(\v^{(n)}\right) + \bm{\delta}_{t-1} - \bm{\delta}_t\\
= &  \frac{1}{N} \sum_{n=1}^N\left(\v_t^{(n)} - \bm{\delta}_t^{(n)} \right) + \bm{\delta}_{t-1} -  \bm{\delta}_t\quad\text{(from \eqref{eq:ecsgd_worker_delta})}\\
= &  \frac{1}{N} \sum_{n=1}^N \left( F'(\x_t;{\xi}_t^{(n)}) + \bm{\delta}_{t-1}^{(n)} - \bm{\delta}_t^{(n)} \right) + \bm{\delta}_{t-1} - \bm{\delta}_t\quad\text{(from (\ref{eq:ecsgd_v_t}))}\\
= &  \frac{1}{N}\sum_{n=1}^N  F'(\x_{t};{\xi}_t^{(n)})  + \Omega_{t-1} - \Omega_t,
%\\ = &  
% f'(\x_{t}) - \Delta_t + \Omega_{t-1} -\Omega_t.
\end{align*}
This completes the proof.
\end{proof}

Due to a similar updating form to $\CSGD$, we can apply a similar proof strategy with particular consideration to the perturbation of $\x$.

\begin{tcolorbox}[colback=blue!5!white,colframe=black!75!black]
\begin{theorem} \label{thm:dsgd_1}
Under Assumptions \ref{ass:gd} ($L$-Liptchitz gradient assumption for $f(\cdot)$) and \ref{ass:bounded_compression}, and the commonly used bounded stochastic gradient assumption,
\[
\E \left[\left\| f'(\x) - {1\over N} \sum_{n=1}^N F'_n(\x_t; \xi_t^{(n)}) \right\|^2\right] \leq \sigma^2,
\]
choose the learning rate to be
\begin{align*}
\gamma = \left(2L + \sqrt{T\over N}\sigma + T^{1/3}\sigma'^{2/3}\right)^{-1}.
\end{align*}
If $T$ is sufficiently large such that the learning rate satisfies the following condition
\begin{align*}
\gamma \leq \min\left\{\frac{1}{4L},~\sqrt{\frac{N}{T}}{1\over \sigma},~\frac{1}{T^{1/3}\sigma'^{2/3}} \right\},
\end{align*}
then the $\DS$ algorithm admits the following convergence rate:
\begin{align*}
{1\over T} \sum_{t=1}^T \E \left\| f'({\x}_t) \right\|^2 \lesssim{1\over T} + {\sigma\over \sqrt{TN}} + {\sigma'^{2/3} \over T^{2/3}},
\end{align*}
where $f(\x_0)$ and $L$ are treated to be constants.
\end{theorem}
\end{tcolorbox}

\begin{proof}
Denote by for short %Editor query: is there some text missing here? 
\[
\Delta_t:= \frac{1}{N}\sum_{n=1}^N   F'_n\left(\x_{t};\xi_t^{(n)}\right) -  f'(\x_t).
\]
To apply the analysis strategy of $\GD$, we first prove some preliminary results to estimate the perturbation. For $\Delta_t$ --- the difference between the stochastic gradient and the true gradient --- we have the following two properties:
\begin{align*}
\mathbb E[\Delta_t] =& \frac{1}{N}\sum_{n=1}^N \left(  f'(\x_t) - \mathbb E \left[  F'_n\left(\x_{t};\xi_t^{(n)}\right)\right] \right) =   \bm{0},
\numberthis \label{eq:ecsgd_proof:delta_1}
\\
\mathbb E[\|\Delta_t\|^2] =& \frac{1}{N^2}\mathbb E \left\| \sum_{n=1}^N \left(  f'(\x_t) -   F'_n\left(\x_{t};\xi_t^{(n)}\right) \right) \right\|^2
%\\ = & 
%\frac{1}{N^2}\sum_{n=1}^N\mathbb E \left\|   f'(\x_t) -   F'_n\left(\x_{t};\xi_t^{(n)}\right)\right\|^2 + %\frac{1}{N^2}\sum_{i\neq i'}^n\left\langle  f'(\x_t) -   F'_n\left(\x_{t};\xi_t^{(n)}\right),  f'(\x_t) - %  F'_n\left(\x_{t};\bm{\zeta}_t^{(i')}\right) \right\rangle,
\\ = & \frac{1}{N^2}\sum_{n=1}^N\mathbb E \left\|   f'(\x_t) -   F'_n\left(\x_{t};\xi_t^{(n)}\right)\right\|^2
\\ \leq & 
\frac{\sigma^2}{N}.\quad \text{(from the bounded SG Assumption)}
\numberthis \label{eq:ecsgd_proof:delta_2}
\end{align*}
Then we estimate the upper bound of $\Omega_t$, which measures the distance between $\x_t$ and $\tilde{\x}_t$:  
\begin{align*}
\mathbb E[\|\Omega_t\|^2] =&\mathbb E\left\| \bm{\delta}_t + \frac{1}{N} \sum_{n=1}^N \bm{\delta}_{t}^{(n)}\right\|^2\\
\leq & 2\mathbb E\left\| \bm{\delta}_t \right\|^2 + 2\mathbb E\left\| \frac{1}{N} \sum_{n=1}^N \bm{\delta}_{t}^{(n)}\right\|^2\\
\leq & 2\sigma'^2 + \frac{2}{N}\sum_{n=1}^N\mathbb E\left\| \bm{\delta}_{t}^{(n)}\right\|^2\\
 \leq & 4\sigma'^2.
\end{align*} %\footnote{\rc Ji's comment: this can be a lemma.}
% \begin{align*}
%    & \E \left[\|\bm{\delta}_t\|^2 \right] 
%    \\ = & 
%    \E\left[\left\| {1\over N}\sum_{n=1}^N Q(\v_t^{(n)}) + \bm{\delta}_{t-1} \right\|^2 \right] 
%    \\ = & 
%    \E \left[\left\| {1\over N}\sum_{n=1}^N Q\left(  F'_n(\x_t; \xi_t^{(n)}) +  \bm{\delta}_{t-1}^{(n)}\right) + \bm{\delta}_{t-1} \right\|^2 \right].
% \end{align*}

From the assumption that $f(\x)$ has the L-Lipschitz gradient, we obtain
\begin{align*}
\mathbb E\|f'(\tilde{\x}_t) - f'(\x_t) \| \leq & L^2\mathbb E\|\tilde{\x}_t - \x_t\|^2
\\ = & 
L^2\gamma^2\mathbb E\|\Omega_{t-1}\|^2 
\\ \leq &
2L^2\gamma^2\sigma'^2.
\numberthis\label{proof:syn_eq1}
\end{align*}

From Lemma~\eqref{lem:ec-sgd}, we have the updating rule
\[
\tilde{\x}_{t+1} = \tilde{\x}_t - \gamma (f'(\x_t) + \Delta_t).
\]
Next we can follow the standard analysis pipeline by considering $\mathbb E f(\tilde{\x}_{t+1}) - \mathbb Ef(\tilde{\x}_t)$:
\begin{align*}
&\mathbb E f(\tilde{\x}_{t+1}) - \mathbb Ef(\tilde{\x}_t)
\\ \leq &  
\mathbb E\left\langle\tilde{\x}_{t+1} - \tilde{\x}_t,f'(\tilde{\x}_t)\right\rangle + \frac{L}{2}\mathbb E\|\tilde{\x}_{t+1} - \tilde{\x}_t\|^2
\\ = & 
-\gamma\mathbb E \left\langle f'(\x_t) , f'(\tilde{\x}_t)\right\rangle  + \gamma\mathbb E \left\langle\Delta_t , f'(\tilde{\x}_t)\right\rangle +  \frac{L\gamma^2}{2}\mathbb E\| f'(\x_t)  - \Delta_t\|^2
\\ = & 
-\gamma\mathbb E \left\langle f'(\x_t), f'(\tilde{\x}_t)\right\rangle +  \frac{L\gamma^2}{2}\mathbb E\| f'(\x_t)\|^2 + \frac{L\gamma^2}{2}\mathbb E\|\Delta_t\|^2 
\\ &
\quad\text{(due to $\mathbb E \Delta_t = \bm{0}$ in \eqref{eq:ecsgd_proof:delta_1})}
\\ \leq & 
-\gamma\mathbb E \left\langle f'(\x_t) , f'(\tilde{\x}_t)\right\rangle   +  \frac{L\gamma^2}{2}\mathbb E\| f'(\x_t) \|^2 + \frac{L\gamma^2\sigma^2}{2N}.
\\ &
\quad\text{(due to \eqref{eq:ecsgd_proof:delta_2})}
\\ = &
- \left(\gamma - {L\gamma^2 \over 2} \right)\mathbb E \left\|f'(\x_t)\right\|^2- \gamma\mathbb E \left\langle f'(\x_t) , f'(\tilde{\x}_t) -  f'(\x_t)\right\rangle 
\\ & + \frac{L\gamma^2\sigma^2}{2N}.
\end{align*}
Next, using the following relaxation, 
\begin{align*}
& |\mathbb E \left\langle f'(\x_t) , f'(\tilde{\x}_t) - f'(\x_t)\right\rangle| 
\\ \leq &
2\mathbb E \left\|f'(\tilde{\x}_t) -  f'(\x_t)\right\|^2 + \frac{1}{2}\mathbb E\| f'(\x_t) \|^2
\\ \leq &
2L^2\mathbb E \left\|\tilde{\x}_t -  \x_t\right\|^2 + \frac{1}{2}\mathbb E\| f'(\x_t) \|^2
\\ \leq &
2L^2\gamma^2\mathbb E \left\|\Omega_t\right\|^2 + \frac{1}{2}\mathbb E\| f'(\x_t) \|^2
\end{align*}
and \eqref{proof:syn_eq1} yields
\begin{align*} 
& \mathbb E f(\tilde{\x}_{t+1}) - \mathbb Ef(\tilde{\x}_t) 
\\ \leq & 
%- \gamma\mathbb E \left\|\nabla  f(\x_t)\right\|^2 + \left(\frac{\gamma}{2}\mathbb E \left\| f'(\x_t)\right\|^2 +  2\gamma\mathbb E \left\|  f'(\tilde{\x}_t) -  f'(\x_t)\right\|^2 \right)   +  \frac{L\gamma^2}{2}\mathbb E\| f'(\x_t) \|^2 + \frac{L\gamma^2\sigma^2}{2n}  
%\\ \leq&  
\left(- \frac{\gamma}{2} +  \frac{L\gamma^2}{2}\right) \mathbb E \left\|\nabla  f(\x_t)\right\|^2 + \frac{L\gamma^2\sigma^2}{2N}  + 4L^2\sigma'^2\gamma^3. %\quad\text{(from \eqref{proof:syn_eq1})}.
\end{align*}
Summing up the inequality above from $t=0$ to $t = T-1$, we get
\begin{align*}
 & \mathbb E f(\tilde{\x}_{T}) - \mathbb Ef(\tilde{\x}_0) \\ 
 \leq & - \left(\frac{\gamma}{2} - \frac{L\gamma^2}{2}\right)\sum_{t=0}^{T-1} \mathbb E \left\|  f'(\x_t)\right\|^2 + \frac{L\gamma^2\sigma^2T}{2N}  + 4L^2\sigma'^2\gamma^3T,
\end{align*}
which can be also written as
\begin{align*}
& \left(\frac{\gamma}{2} - \frac{L\gamma^2}{2}\right)\sum_{t=0}^{T-1} \mathbb E \left\|\nabla  f(\x_t)\right\|^2 
\\ \leq &
\mathbb E f(\tilde{\x}_{0}) - \mathbb Ef(\tilde{\x}_{T}) + \frac{L\gamma^2\sigma^2T}{2N}  + 4L^2\sigma'^2\gamma^3T\\
\leq & \mathbb E f(\x_{0}) - \mathbb Ef(\x^*) + \frac{L\gamma^2\sigma^2T}{2N}  + 4L^2\sigma'^2\gamma^3T.
\end{align*}
Since $\gamma \leq {1\over 4L}$, we can rewrite the above inequality as 
\begin{align*}
{1\over T} \sum_{t=0}^{T-1} \mathbb E \left\|\nabla  f(\x_t)\right\|^2 
 \lesssim &
{1\over \gamma T} + \frac{\gamma \sigma^2}{N}  + \sigma'^2\gamma^2.
\end{align*}
Based on the choice of $\gamma$, one can verify that
\begin{align*}
    {1\over \gamma T} \lesssim & {1\over T} +   {\sigma \over \sqrt{NT}} + {\sigma'^{2/3} \over T^{2/3}}
    \\
    {\gamma \sigma^2 \over N} \lesssim & {\sigma \over \sqrt{NT}}
    \\ 
    {\sigma'^2 \gamma^2} \lesssim & {\sigma'^{2/3} \over T^{2/3}},
\end{align*}
which implies the claim.
\end{proof}
 Comparing this approach to \eqref{eq:csgd_convergence}, we can now see one more advantage of $\DS$ over $\CSGD$ with respect to the convergence efficiency, in addition to the fact that $\DS$ does not require unbiased compression, as does $\CSGD$.
 
\begin{remark}
A compression assumption that is probably more realistic than Assumption~\ref{ass:bounded_compression} could be 
\[
\E\left[ \left\| Q(\y) - \y \right\|^2 \right] \leq \alpha \|\y\|^2\quad \forall \y.
\]
However, this will involve more complicated analysis. Readers can refer to \citet{liu2019double}. We can still see similar advantages over $\CSGD$, even under this new assumption.
\end{remark} 

}

\chapter{System Relaxation 2: Asynchronous Training}

In implementations that we
described in previous \parts, the communications
among all workers are perfectly synchronized ---
all workers conduct computation at the same time, 
and are all blocked until the communication 
among all machines is finished. 
However, when the number of machines in a distributed
system grows, such a {\em synchronous}
strategy might have several limitations.
First, as the communication costs 
increase, the amount of time that each machine
can spend on computation decreases, as they
cannot conduct any computation during
the communication phase.
Second, if some machines are slower than
other machines (i.e., there are stragglers), all machines
need to wait for the slowest machine to
finish computation. 

In this \part, we describe one system
relaxation to accommodate these problems. 
In this relaxation, we remove the 
synchronization barriers among all the machines.
This decreases the synchronization overhead, with
the consequence that each machine now has access to
a {\em staled} model.

\section{System Implementation}

There can be multiple ways of implementing an 
asynchronous communication strategy, each of
which might have slight differences in 
their convergence behavior. In this \part,
we describe one of the simplest implementations, 
which is easiest in terms of both theoretical 
analysis and system implementation.

We focus on a scenario in which the system
is implemented using a {\em single-server}
parameter server. 

\begin{enumerate}
\item The parameter server holds the global 
replica of the model.
\item Each worker $w_i$ works in four phases. (1) At
the beginning of each iteration, $w_i$
asks the parameter server for the
global replica of the model. (2) Upon receiving
this model, the worker $w_i$ uses it 
to compute the local gradient. (3) The 
worker then sends the local gradient to the 
parameter server, which then
applies it to update the global model replica.
In this step, we assume that the update
of the global model is {\em atomic} --- that is, the parameter
won't ``mix'' the updates from different 
workers and these updates won't overwrite
each other.
(4) Worker $w_i$ waits until the transfer
is finished, and repeats from step (1) 
immediately.
\end{enumerate}

In practice, the above process can be implemented
in slightly different ways.
\begin{enumerate}
\item A worker $w_i$ can start processing
the next data point without waiting for
the transfer to finish.
\item The parameter server does not need to
conduct atomic updates on the global model
replica.
\end{enumerate}

Some of these implementations can make 
theoretical analysis more engaged. We 
choose to focus on the simple asynchronous
communication scheme described above, 
as it is one of the simplest implementations
that can demonstrate the advantage of
asynchronous communication under
our performance model.

\subsubsection{Impact of Asynchronous Communication}

\begin{figure}
\includegraphics[width=1.0\textwidth]{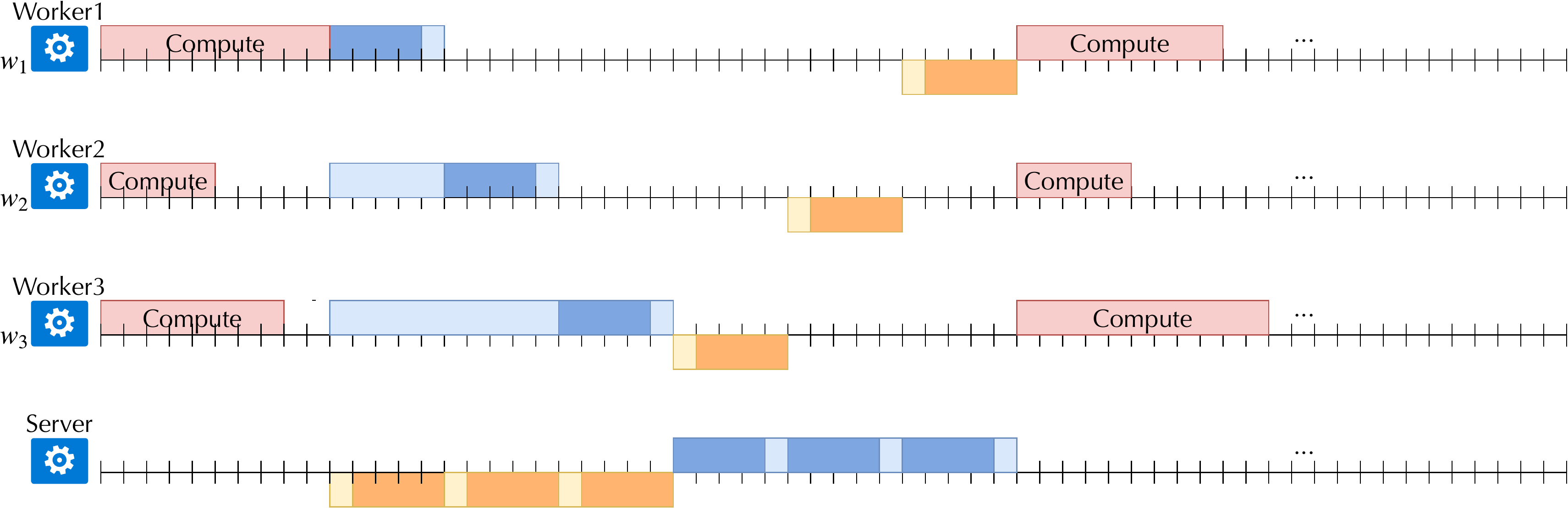}
\caption{Illustration of the Impact of Asynchronous Communication (without asynchrony)}
\label{fig:async:impact:orig}
\end{figure}

\begin{figure}
\includegraphics[width=1.0\textwidth]{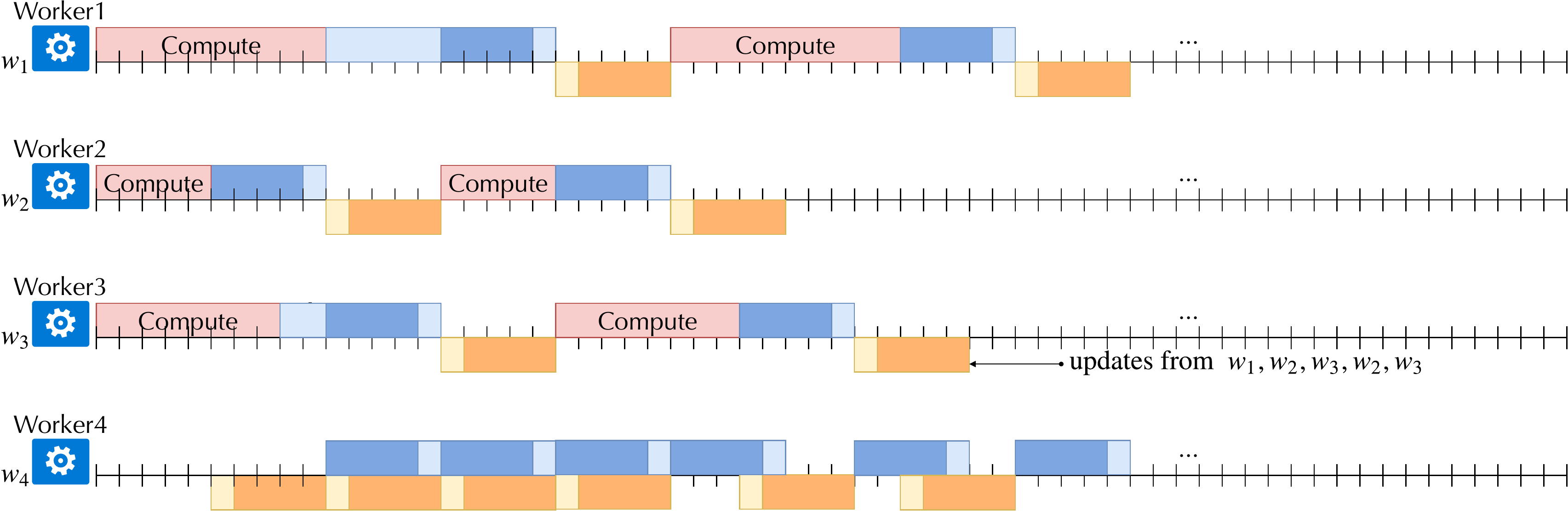}
\caption{Illustration of the Impact of Asynchronous Communication (with asynchrony)}
\label{fig:async:impact:opt}
\end{figure}

Figure~\ref{fig:async:impact:orig} and Figure~\ref{fig:async:impact:opt} illustrate the
impact of asynchronous communication, 
{\em under our simplified performance model for
communication.}
We see that, under our current performance
model, asynchronous communication can make
each iteration faster --- instead of
waiting for all workers to finish processing,
a worker can start the
next iteration of computation
immediately after its communication
with the parameter server is finished.
On the other hand, we also observe that
the model used by each machine for
computation might be {\em \rv staled}. 
Take, for example, the last model 
of worker 3 $w_3$ in 
Figure~\ref{fig:async:impact:opt} ---
this model misses one update
from $w_1$ because $w_3$ does not
wait for $w_1$ before it starts
the next iteration of computation.

As we can see from 
Figure~\ref{fig:async:impact:opt},
one potential system bottleneck 
is the parameter server: when the server
is saturated in terms of
network bandwidth, adding more 
workers cannot make processing
more data (i.e., calculating more
gradients) faster without making
the staleness larger. One way to 
accommodate this is to use
multi-server parameter server
architecture, in which each parameter
server takes charge of one partition
of the model. In this case, 
different partition of the model might
have different {\em staleness}, which
will, not surprisingly, make the
theoretical analysis more engaged.

\subsubsection{Caveats}

In practice, there are other considerations
that require careful attention when
using asynchronous communication for 
distributed training. We briefly summarize these caveats and refer
readers to \part~\ref{sec:further_readings}
for further readings.

\begin{enumerate}
\item {\bf Bounded Staleness.} If not
careful, the above implementation does
not necessarily provide bounded staleness ---
it is possible that one machine is always
waiting for its communication to finish.
In this case, some updates on the 
global replica might be using a gradient
that is calculated based on a very
old model, which could slow down the convergence.
Some research have tried to 
accommodate this by enforcing 
bounded staleness in their implementation.
\item {\bf Straggler.} When there
are stragglers in the system (i.e., some
machines are significantly slower than other machines), the staleness is systematic --- one
partition of the data is always more staled 
than another. In this case,
staleness-aware / heterogeneous-aware algorithms could be designed to stabilize 
the convergence behavior.
\end{enumerate}

One motivation for introducing 
asynchronous communication is to accommodate
stragglers. There are other approaches
to achieve the same objective without 
introducing asynchrony, however. For example, there
has been research that attempts to use error correction
code to make the synchronous approach more
robust to stragglers. In this strategy,
each worker conducts a small amount of redundant 
computation, such that 
the exact (or approximate) gradient can be recovered 
without waiting for the slowest $k$ machines.

\section{Theoretical Analysis}

We now analyze $\ASGD$, a $\SGD$ variant
with asynchronous communications.
The updating rule of $\ASGD$ can be cast into the following form:
\begin{align*}
\x_{t+1} = \x_t - \gamma g_t(\x_{\dd(t)}),
\end{align*}
where $\dd(t) \in \{1,2,\cdots, t\}$ and $\x_{\dd(t)}$ denote some early iterations of the model $\x_t$.

One important property for $g_t(\x_{\dd(t)})$ is
\begin{align}
\E[g_t(\x_{\dd(t)})] = f'(\x_{\dd(t)}).
\end{align}

To ensure the convergence rate, let us make an additional assumption about staleness:
\begin{tcolorbox}[colback=pink!5!white,colframe=black!75!black]
\begin{assumption} \label{ass:asgd}
%=======
%An clear advantage of $\SGD$ is that the computational complexity reduces to $O(d)$ per iteration. It is worth pointing out that the $\SGD$ algorithm is NOT a descent algorithm due to randomness.
%
%The next question is whether it converges and, if so, how fast. We first make some typical assumptions.
%\begin{tcolorbox}[colback=pink!5!white,colframe=black!75!black]
%\begin{assumption}
%>>>>>>> b94b8679f310fa566039184f25fe4da4b1ce56ff
We make the following assumption:
\begin{itemize}
\item ({\bf Bounded staleness}) Assume that the staleness is bounded, that is,  
\[
0 \leq t - \dd(t) \leq \tau\quad \forall t,
\]
where $\tau$ is a global staleness upper bound.
\end{itemize}
\end{assumption}
\end{tcolorbox}
Intuitively, to ensure convergence, the staleness cannot be overlarge. In an extreme case, if $\dd(t)=1\;\forall t$, then the model is always updated from the true stochastic gradient in the first step, which has no hope of converging. Therefore, it is necessary to restrict the bound of $t - \dd(t)$. In practice, the staleness parameter $t-\dd(t)$ is proportional to the number of workers.

Next we start to show the convergence rate proof of $\ASGD$, step by step.
\begin{align*}
f(\x_{t+1})- f(\x_t) \leq -\gamma \langle f'(\x_t),  g_t(\x_{\dd(t)})\rangle + {L\gamma^2 \over 2} \|g_t(\x_{\dd(t)})\|^2 
\end{align*}

Take expectation on both sides:
\begin{align*}
\E[f(\x_{t+1})]- \E[f(\x_t)] \leq & -\gamma \E[\langle f'(\x_t),  g_t(\x_{\dd(t)})\rangle + 
\\ &
{L\gamma^2 \over 2} \E[\|g_t(\x_{\dd(t)})\|^2] 
\numberthis \label{eq:asgd_proof_0}
%\\ = &
%\gamma \E[\langle f'(\x_t),  g(\x_{\dd(t)})\rangle] + {L\gamma^2 \over 2} \E[\|g_t{\x_{\dd(t)}}\|^2] 
% \\ = &
% \gamma \E[\langle f'(\x_t), f'(\x_{t_{\tau_t}})]
\end{align*}
We consider the items on the right-hand side respectively. For $\E[\|g_t(\x_{\dd(t)})\|^2]$, we have
\begin{align*}
\E_{\dd(t)}[\|g_t(\x_{\dd(t)})\|^2] =&  \E_{\dd(t)}[\|\g_t(\x_{\dd(t)})-f'(\x_{\dd(t)})\|^2] + \|\f_t(\x_{\dd(t)})\|^2
\\ \leq &
\sigma^2 + \|\f_t(\x_{\dd(t)})\|^2,
\numberthis \label{eq:asgd_proof_1}
\end{align*}
where the first equality is due to the fact \eqref{eq:mean-var}, and the second is due to the bounded variance assumption, that is, Assumption~\ref{ass:sgd}.

For $\E[\langle f'(\x_t),  g(\x_{\dd(t)})\rangle]$, we have
\begin{align*}
& \E[\langle f'(\x_t),  g(\x_{\dd(t)})\rangle] 
\\= & 
\E[\langle f'(\x_t),  \E_{\dd(t)}[g(\x_{\dd(t)})]\rangle]
\\= & 
\E[\langle f'(\x_t), f'(\x_{\dd(t)})\rangle]
\\ = &
 \E\left[{1\over 2}\|f'(\x_t) - f'(\x_{\dd(t)})\|^2 - {1\over 2}\|f'(\x_t)\|^2 - {1\over 2}\|f'(\x_{\dd(t)})\|^2 \right]
 \numberthis \label{eq:asgd_proof_2}
%\E[\langle f'(\x_t),  f'(\x_t)\rangle] + \E[\langle f'(\x_t), f'(\x_{\dd(t)}) - f'(\x_t)\rangle]
%\\ = &
%\E [\|f'(\x_t)\|^2] + \E [\|f'(\x_t)]
\end{align*}
The last equality uses an important property,
\begin{align*}
\langle \a, \b \rangle = {1\over 2}\|\a-\b\|^2 - {1\over 2}\|\a\|^2 - {1\over 2}\|\b\|^2 ,
\end{align*}
which can be verified by a straightforward linear algebra computation. Next we take expectation on both sides of \eqref{eq:asgd_proof_0} and plug \eqref{eq:asgd_proof_1} and \eqref{eq:asgd_proof_2} into that:
\begin{align*}
&\E[f(\x_{t+1})] - \E[f(\x_t)]  
\\ \leq &
-{\gamma \over 2} \E[\|f'(\x_t)\|^2] - {\gamma \over 2} \E[\|f'(\x_{\dd(t)})\|^2] + {\gamma \over 2} \E[\|f'(\x_t) - f'(\x_{\dd(t)})\|^2]
\\ & 
+ {\gamma^2 L \sigma^2 \over 2} + {\gamma^2L \over 2} \E[\|f'(\x_{\dd(t)})\|^2]
\\ \leq &
-{\gamma \over 2} \E[\|f'(\x_t)\|^2] + {\gamma^2 L \sigma^2 \over 2} - {\gamma \over 2}\left(1- {\gamma L}\right) \E[\|f'(\x_{\dd(t)})\|^2]
\\ & + {\gamma \over 2}\E[\|f'(\x_t) - f'(\x_{\dd(t)})\|^2]
\\ \leq &
-{\gamma \over 2} \E[\|f'(\x_t)\|^2] + {\gamma^2 L \sigma^2 \over 2} + {\gamma \over 2}\E[\|f'(\x_t) - f'(\x_{\dd(t)})\|^2] ,
\numberthis \label{eq:asgd_proof_3}
\end{align*}
where the last inequality is obtained by choosing a sufficiently small learning rate $\gamma$ satisfying 
\begin{align}
1 \geq \gamma L. \label{eq:asgd_proof_lr1}
\end{align}

We have seen the first two terms on the right-hand side of \eqref{eq:asgd_proof_3} (if ignoring the constant parts) in the proof to $\SGD$. %The third term of \eqref{eq:asgd_proof_3} will be equal to $\E\|f'(\x_{t})\|^2$ (ignoring the constant part) if the staleness is zero, that is, $\tau_t = 0$. 
We can roughly treat this term as $\E\|f'(\x_{t})\|^2$ plus some perturbation, depending on the staleness $\tau$. The major term we need to treat very seriously is the last term $\E[\|f'(\x_t) - f'(\x_{\dd(t)})\|^2]$, whose upper bound is given by the following lemma. It basically shows that this item is bounded by a higher order of $\gamma$: $O(\gamma^3)$.
\begin{tcolorbox}[colback=yellow!5!white,colframe=black!75!black]
\begin{lemma}\label{lem:asgd_1}
Under Assumptions~\ref{ass:gd}, \ref{ass:sgd}, and \ref{ass:asgd}, we have
\begin{align*}
\E[\|f'(\x_t) - f'(\x_{\dd(t)})\|^2] \leq & 2\gamma^2L^2 \tau \sigma^2 +
%\\ &
2\gamma^2L^2 \tau \sum_{s=\dd(t)}^{t-1} \E\left[ \left\| f'_s(\x_{\dd(t)}) \right\|^2 \right].
\end{align*}
\end{lemma} 
\end{tcolorbox}
\begin{proof}
We start by bounding the difference between $f'(\x_t)$ and $f'(\x_{\dd(t)})$
\begin{align*}
 & \E[\|f'(\x_t) - f'(\x_{\dd(t)})\|^2]
\\ \leq &
L^2 \E[\|\x_t - \x_{t_{\tau_t}}\|^2]\quad(\text{Due to Assumption~\ref{ass:gd}})
\\ = &
\gamma^2L^2 \E\left[\left\|\sum_{s=\dd(t)}^{t-1} g_s(\x_{\dd(s)})\right\|^2\right].
\\ = & 
\gamma^2L^2 \E\left[\left\|\sum_{s=\dd(t)}^{t-1} g_s(\x_{\dd(s)}) - \sum_{s=\dd(t)}^{t-1} f'_s(\x_{\dd(s)}) + \sum_{s=\dd(t)}^{t-1} f'_s(\x_{\dd(s)}) \right\|^2\right]
\\ \leq &
2 \gamma^2 L^2 \E\left[\left\|\sum_{s=\dd(t)}^{t-1} \underbrace{\left(g_s(\x_{\dd(s)}) -  f'_s(\x_{\dd(s)}) \right)}_{=:\xi_{s-\dd(t)}}\right\|^2\right]
\\ & + 
2\gamma^2 L^2 \E\left[\left\|\sum_{s=\dd(t)}^{t-1} f'_s(\x_{\dd(s)})\right\|^2\right],
\numberthis \label{eq:asgd_proof_4}
\end{align*} 
where the last inequality uses a variant of triangle inequality $\|\a+\b\|^2 \leq 2\|\a\|^2 + 2\|\b\|^2$.
Note that the sequence of $\{\xi_0, \xi_1, \cdots, \xi_{\tau_t-\dd(t)-1}\}$ 
%\[
%\{g_s(\x_{\dd(s)}), g_{s+1}(\x_{s+1-\tau_{s+1}}),\cdots, g_{t-1}(\x_{t-1-\tau_{t-1}})\}
%\]
is a martingale sequence with 
\[
\xi_{l} := g_{s+l}(\x_{\dd(s+l)}) -  f'_{s+l}(\x_{\dd(s+l)}).
\]
A martingale sequence satisfies that
\begin{align*}
\E(\xi_{l+1}~|~\xi_0, \xi_1, \cdots, \xi_l) = 0.
\end{align*} 
Therefore, it is easy to verify that
\begin{align*}
\E\left[\left\|\sum_{l=0}^{t-1-\dd(t)} \xi_l \right\|^2 \right] = & \E\left[\left\|\sum_{l=0}^{t-2-\dd(t)} \xi_l \right\|^2 \right] + \E\left[\left\|\xi_{t-1} \right\|^2\right] \\
= &
\sum_{l=0}^{t-1-\dd(t)} \E\left[ \|\xi_l\|^2 \right].
\end{align*}
In other words, we have
\[
\E\left[\left\|\sum_{s=\dd(t)}^{t-1} \left(g_s(\x_{\dd(s)}) -  f'_s(\x_{\dd(s)}) \right) \right\|^2\right] = \sum_{l=0}^{t-1-\dd(t)} \E\left[ \|\xi_l\|^2 \right] \leq \tau \sigma^2.
\]

Applying this property to \eqref{eq:asgd_proof_4} yields
\begin{align*}
 & \E[\|f'(\x_t) - f'(\x_{\dd(t)})\|^2] 
 \\ \leq & 
 2\gamma^2L^2 \tau \sigma^2 + 2\gamma^2 L^2 \E\left[\left\|\sum_{s=\dd(t)}^{t-1} f'_s(\x_{\dd(s)})\right\|^2\right]
 \\ \leq &
 2\gamma^2L^2 \tau \sigma^2 +  2\gamma^2L^2 \tau \sum_{s=\dd(t)}^{t-1} \E\left[ \left\| f'_s(\x_{\dd(s)}) \right\|^2 \right],
\end{align*}
where the last inequality uses Assumption~\ref{ass:asgd} the following property: for any vectors $\a_1, \a_2, \cdots, \a_n$, we have
\[
\left\|\sum_{i=1}^n \a_i\right\|^2 \leq n \sum_{i=1}^n\|\a_i\|^2.
\]
This completes the proof.
\end{proof}
We choose the learning rate to be sufficiently small. In particular, let 
\begin{align}
\gamma L \tau \leq 1/2. \label{eq:asgd_proof_lr2}
\end{align}
Using the following short notations:
\begin{align*}
a_t := & \E[\|f'(\x_t)\|^2]
\\
b_t := & \E[f(\x_t)],
\end{align*}
and apply Lemma~\ref{lem:asgd_1} to \eqref{eq:asgd_proof_3}:
\begin{align}
\nonumber
b_{t+1} - b_t \leq & -{\gamma \over 2} a_t + {\gamma^2 L \over 2}\left(1+2\gamma L \tau\right)\sigma^2 + {\gamma^3L^2\tau} \sum_{s=\dd(t)}^{t-1} a_s
\\ \nonumber
\leq & -{\gamma \over 2} a_t + {\gamma^2 L \over 2}\left(1+2\gamma L \tau\right)\sigma^2 + {\gamma^3L^2\tau} \sum_{s=\dd(t)}^{t-1} a_s
\\
\leq & 
-{\gamma \over 2} a_t + {\gamma^2 L}\sigma^2 + {\gamma \over 4\tau} \sum_{s=\dd(t)}^{t-1} a_s.
\label{eq:asgd_proof_5}
\end{align}
Next we take sum of \eqref{eq:asgd_proof_5} over $t$ from $t=1$ to $t=T$
\begin{align*}
b_{T+1} - b_1 \leq & -{\gamma \over 2} \sum_{t=1}^Ta_t + \gamma^2 LT \sigma^2 + {\gamma \over 4\tau} \sum_{t=1}^T \sum_{s=\dd(t)}^{t-1}a_s
\\ \leq &
-{\gamma \over 2} \sum_{t=1}^Ta_t + \gamma^2 LT \sigma^2 + {\gamma \over 4} \sum_{t=1}^{T} a_t
\\ = &
- {\gamma \over 4} \sum_{t=1}^T a_t + \gamma^2 LT\sigma^2.
\end{align*}
Therefore, we have the following convergence rate:
\begin{align*}
{1\over T}\sum_{t=1}^T a_t \leq {4(b_1 - b_{T+1}) \over \gamma T} + {4\gamma L\sigma^2}.
\end{align*}
If we treat $f(\x_1) - f^\star$ to be constant, then we have
\begin{align*}
{1\over T}\sum_{t=1}^T\E \left\| f'(\x_{t}) \right\|^2 \lesssim & \frac{f(\x_1) - \E\left[f(\x_{T+1})\right]}{\gamma T} + \gamma L\sigma^2
\\ \lesssim &  
\frac{f(\x_1) - f^\star}{\gamma T} + \gamma L\sigma^2
\\ \lesssim &  
\frac{1}{\gamma T} + \gamma L\sigma^2.
\end{align*}
We choose the learning rate $\gamma$ to be
\begin{align}
\gamma = \frac{1}{L(\tau+1) + \sqrt{TL}\sigma} \label{eq:asgd_lr}
\end{align}
to satisfy the requirements of $\gamma$ in \eqref{eq:asgd_proof_lr1} and \eqref{eq:asgd_proof_lr2},
which leads to
\begin{align*}
{1\over \gamma T} \leq & {L(\tau+1)  \over T} + {\sqrt{L}\sigma \over \sqrt{T}}
\\
\gamma L \sigma^2 \leq & {\sqrt{L}\sigma \over \sqrt{T}}.
\end{align*}
Therefore, we can summarize the convergence rate of $\ASGD$ in the following theorem.
%We choose the learning rate $\gamma$, satisfying 
%\[
%1 \geq \gamma L
%\]
%and obtain a simpler form:

\begin{tcolorbox}[colback=blue!5!white,colframe=black!75!black]
\begin{theorem} \label{thm:asgd_1}
Choose the learning rate in \eqref{eq:asgd_lr} for $\ASGD$. Under Assumptions~\ref{ass:gd}, \ref{ass:sgd}, and \ref{ass:asgd}, $\ASGD$ admits the following convergence rate:
\begin{align*}
{1\over T} \sum_{t=1}^T \E \left\| f'({\x}_t) \right\|^2 \lesssim{L\over T} + {\sqrt{L}\sigma\over \sqrt{T}} + {L\tau \over T},
\end{align*}
where we treat $f(\x_1)-f^\star$ to be constant.
\end{theorem}
\end{tcolorbox}
We highlight the following observations from this convergence analysis:
\begin{itemize}
    \item ({\bf Consistency to $\SGD$}) If there is only one worker, then $\tau = 0$ and the $\ASGD$ algorithm reduces the $\SGD$ algorithm. We can see that the convergence rate in Theorem~\ref{thm:asgd_1} is consistent with $\SGD$.
    \item ({\bf Linear speedup}) The additional term in the rate is the last term, as compared to $\SGD$. It is caused by the asynchronous parallelism. Recall that $\tau$ is proportional to the total number of workers. If $\tau$ satisfies $\tau \leq {\sqrt{T}\sigma \over \sqrt{L}}$, then the convergence efficiency would not be affected by using asynchronous updating. As a result, linear speedup can be achieved. Here, the linear speedup is in the sense that the average computational complexity per worker is asymptotically the computational complexity using one single worker divided by $N$ -- the total number of workers.
%Editor comment: This last sentence is a bit unclear. I think the problem is "speedup is in the sense that" -- is there a word missing after "is", like "achieved"? Or could "is" be replaced by "takes place"? Or should it be something like "the linear speedup is due to the fact that the average computational complexity..."?
\end{itemize}

\chapter{System Relaxation 3: Decentralized Communication}

Lossy communication compression is designed
to alleviate system bottlenecks caused
by network bandwidth. Another type of
network bottleneck is caused by {\em latency}. In that case, when there are $N$
workers, both the \texttt{AllReduce} and the multi-server
parameter server have an $O(N)$ dependency
on the network latency. The fundamental 
reason for this is that all these approaches 
insist that the information on each
worker be propagated to all other
workers in a single round of communication.

There are multiple standard ways to get rid
of the latency bottleneck, e.g., using
a textbook reduction tree. In this \part, we 
describe an alternative approach to improving the latency overhead. We call this method
 decentralized communication.
Specifically, we form a logical
ring among $N$ workers (all workers
are still connected via the same ``logical
switch''). At every single iteration, 
a worker sends one message to the neighbor on its immediate left and one message to the neighbor on its immediate right. With this method, the latency overhead
becomes $O(1)$. On the downside, however, the
information on a single worker only
reaches its two adjacent neighbors in one round of communication.

\section{System Implementation}

\begin{figure}
\includegraphics[width=1.0\textwidth]{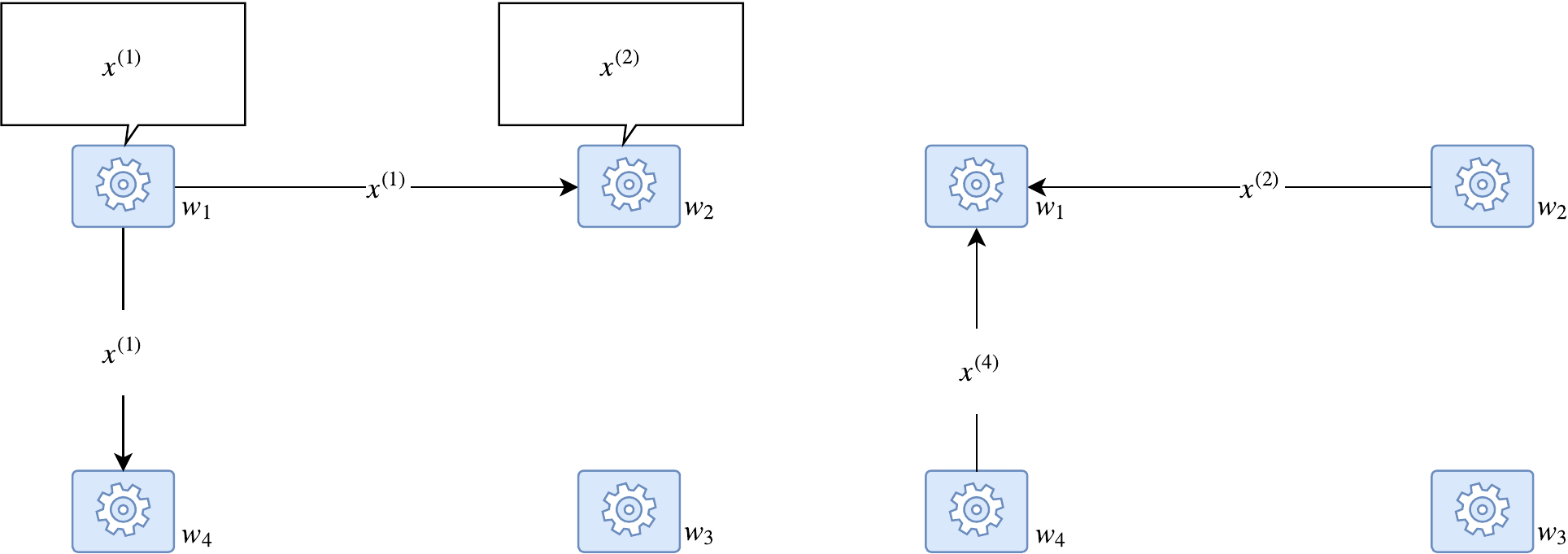}
\caption{Illustration of Decentralized Communication
Pattern on the first worker $w_1$ (other
workers are similar). Each
worker holds a local model replica and only
sends this model to the neighbors on its immediate right and left.}
\label{fig:decent}
\end{figure}

To implement this system relaxation, we rely
on a communication pattern like that of the
{\em model aggregation} approach for
implementing distributed SGD instead
of exchanging {\em gradients} as we
did in the previous \part on 
system exchange {\em models}.
Specifically,
\begin{enumerate}
\item At step $t$, each worker $w_n$ holds a model replica $\x_t^{(n)}$. The system first calculates the stochastic
gradient using its local replica $f'(x^{(n)})$
and then updates its local model using this standard
SGD rule:
\[
\x_{t + 1/2}^{(n)} = \x_{t}^{(n)} - \gamma f'(\x^{(n)}).
\]
\item Each worker $w_n$ sends its locally 
updated model $x_{t + 1/2}^{(n)}$ to the neighbor on its immediate right, $w_{(n+1) \mod N}$, and
to the neighbor on its immediate left, $w_{(n - 1) \mod N}$.
Symmetrically, the worker will also receive models
from the neighbor on its immediate right, $w_{(n+1) \mod N}$,
and from the neighbor on its immediate left, $w_{(n - 1) \mod N}$.
Upon receiving the neighbors' models,
the worker updates its local model as the
average between its local model and its neighbors' models:
\[
\x_{t+1}^{(n)} = \frac{1}{3}\left(
	\x_{t + 1/2}^{(n)} +
	\x_{t + 1/2}^{((n + 1) \mod N)} +
	\x_{t + 1/2}^{((n - 1) \mod N)} 
 \right).
\]
\end{enumerate}

\subsubsection{Impact of Decentralized Communication}

\begin{figure}[t!]
\includegraphics[width=1.0\textwidth]{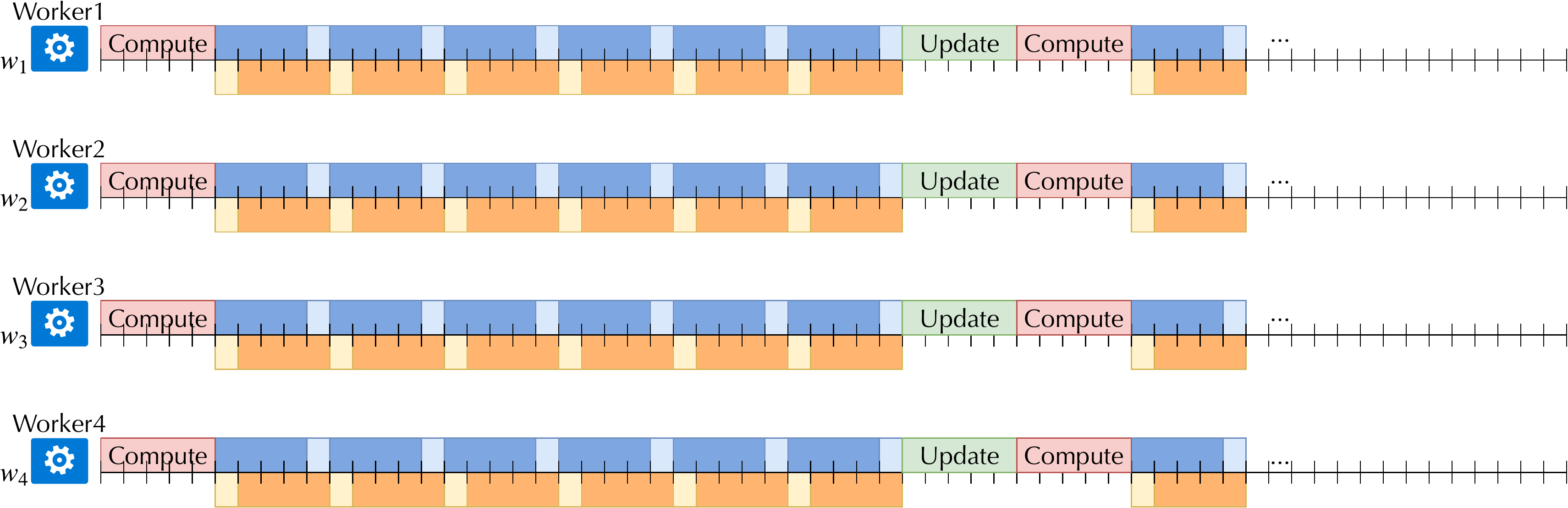}
\caption{Illustration of the Impact of Decentralized Communication (without decentralization)}
\label{fig:decent:impact:orig}
\end{figure}

\begin{figure}[t!]
\includegraphics[width=1.0\textwidth]{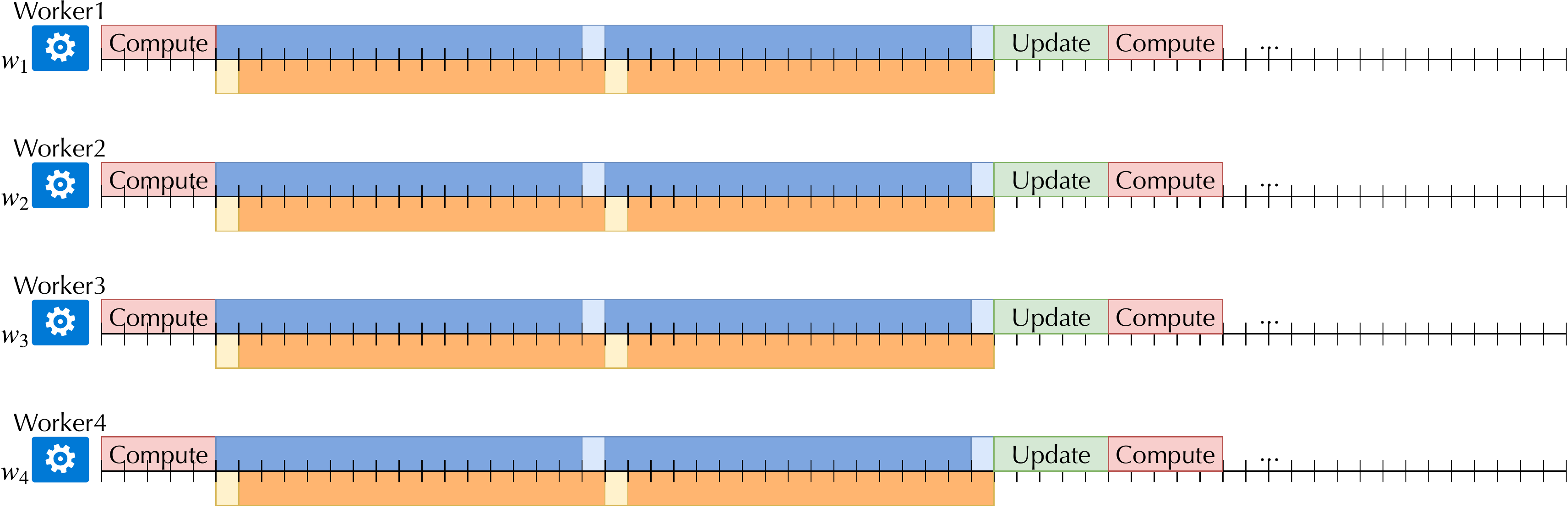}
\caption{Illustration of the Impact of Decentralized Communication (with decentralization)}
\label{fig:decent:impact:opt}
\end{figure}

The communication cost 
for one round of communication in the
decentralized setting is 
\[
2 t_{\text{latency}} + 2 t_{\text{transfer}}.
\]
Recall that the communication cost of the 
multi-server parameter server or the \texttt{AllReduce}
is
\[
2(N-1) t_{\text{latency}} + 2 \frac{N-1}{N} t_{\text{transfer}}.
\]
With our method, we see a clear improvement with the decentralized
communication in terms of communication 
latency. When the underlying 
network has a high latency, the decentralized
strategy can be significantly faster.

Figure~\ref{fig:decent:impact:orig}
and Figure~\ref{fig:decent:impact:opt} illustrate
the communication patterns. Interestingly,
in this specific example, 
the decentralized
approach communication might seem to take longer.
This happens because, when there are
$n$ workers, each worker only needs to send
$\frac{N-1}{N}$ of its model in the
centralized case instead
of the full model. In the decentralized
case, workers need to exchange their full models.
As a result, the transfer time of the decentralized
approach might be slightly higher. However,
as illustrated in Figure~\ref{fig:decent:impact:opt},
the decentralized approach has an
advantage in terms of latency, and this
improvement will be more significant when
there are more workers and the underlying
network has a higher latency.

More generally, one can extend the above example
beyond a simple ring topology. We consider this below
 in the theoretical analysis.

\section{Theoretical Analysis}

We analyze $\DSGD$, a $\SGD$ variant with
decentralized communication. {\rv
Let us consider the same objective as \eqref{eq:ecsgd_obj}. For convenience of reference, we repeat it again here}
\begin{align}
\min_{\x}:\quad \left\{f({\x}):={1\over N}\sum_{n=1}^N f_n(\x) \right\}
\label{eq:dsgd_obj}
\end{align}
where $f_n({\x}):= \E_{\xi\sim \mathcal{D}_n} F_n(\x; \xi)$ and $\mathcal{D}_n$ denotes the distribution of local data at node $n$. We do not assume that all nodes can access the whole dataset. 

Based on the algorithm description in the previous section, the $\DSGD$ algorithm's updating rule can be cast in the following form:
\begin{align}
\X_{t+1} = \left(\X_t - \gamma \G\(\X_t; \{\xi_t^{(n)}\}_{n=1}^N\) \right)W.
\label{eq:DSGD}
\end{align}
Here we use the following notations:
\begin{align*}
\X_{t} := & \left[
\x^{(1)}_t, \x^{(2)}_t, \cdots, \x^{(N)}_t
\right] 
\\
\G\(\X_t; \{\xi_t^{(n)}\}_{n=1}^N\) := & \left[
F'_1\(\x^{(1)}_t; \xi_t^{(1)}\),
\cdots,
F'_N\(\x^{(N)}_t; \xi_t^{(N)}\) 
\right]
\end{align*} 
where $\x_t^{(n)}$ denotes the local model on node $n$ at time $t$. We also use $\G_t$ to denote $\G\(\X_t; \{\xi_t^{(n)}\}_{n=1}^N\)$ for short. $W \in \R^{N\times N}$ is called the confusion matrix.

\subsection{Assumptions}
To show the convergence rate of $\DSGD$, we need to make a few important assumptions about the objective function:
\begin{tcolorbox}[colback=pink!5!white,colframe=black!75!black]
\begin{assumption} \label{ass:dsgd_1}
We make the following assumptions:
\begin{itemize}
\item ({\bf Smoothness and Lipschitzian gradient}) All functions $F_n(\cdot; \xi)$'s are smooth and all $f_n(\x)$'s have an $L-$Lipschitzian gradient, that is,
$\forall \x, \; \forall \y\; \forall n\in [N]$
\begin{align*}
\|f_n'(\x) - f_n'(\y)\| \leq & L\|\x- \y\| \quad \label{eq:ass:dsgd_1}
%\\
%f(\y) - f(\x) \leq & \langle f'(\x), \y-\x \rangle + {L\over 2} \|\y - \x\|^2  \label{eq:ass:gd_2}
\end{align*}
\item ({\bf Unbiased sampling}) For all workers $n\in [N]$, the stochastic gradient is unbiased, that is,
\[
\E_{\xi\sim \mathcal{D}_n}[F'_n(\x; \xi)] = f'_n(\x)\quad \forall \x
\]
\item ({\bf Bounded inner variance}). All local stochastic gradients have a bounded variance, that is,
\[
\E_{\xi \sim \mathcal{D}_n}\left\|f'_n(\x) - F'_n(\x; \xi)\right\|^2 \leq \sigma^2 \quad \forall \x
\]
\item ({\bf Bounded outer variance}). The global gradient variance is bounded, that is,
\[
{1\over N}\sum_{n=1}^N \left\|f'_n(\x) - f'(\x) \right\|^2 \leq \varsigma^2\quad \forall \x.
\]
\end{itemize}
\end{assumption}
\end{tcolorbox}
The first assumption on the smoothness and the Lipschitz gradient is essentially the same as Assumption~\ref{ass:gd}. The second assumption is essentially that all local stochastic gradients are locally unbiased. The bounded inner variance assumption assumes that all local stochastic gradients have a bounded local variance that is similar to Assumption~\ref{ass:sgd}. The last assumption 
is that the total gradient difference among workers is bounded. If all workers access the same dataset, then $\varsigma = 0$.

We next make the necessary assumptions for the confusion matrix $W$:
\begin{tcolorbox}[colback=pink!5!white,colframe=black!75!black]
\begin{assumption} \label{ass:dsgd_2}
We also make the following assumptions for the confusion matrix $W$:
\begin{itemize}
\item ({\bf Symmetric and doubly stochastic matrix}) $W$ is a symmetric and doubly stochastic matrix, that is,
\[
W^\top \1 = \1\quad\text{and}\quad W=W^\top;
\]
\item ({\bf Spectral gap}) denoted by $\rho$ the second largest eigenvalue of $W$ in term of the absolute value, that is,
\[
\rho:= \max_{n=2,3,\cdots, N}|\lambda_n(W)|
\]
where $\lambda_n(W)$ denotes the $n$ largest eigenvalue of $W$. We assume that the spectral gap $1-\rho$ is greater than $0$.
\end{itemize}
\end{assumption}
\end{tcolorbox}
$W^\top \1 = 1$ is to ensure that the weighted sum is $1$ when take the average of models, and the symmetry ensures that all eigenvalues of $W$ are real numbers. The spectral gap $1-\rho$ roughly measures how fast the information can be spread over the network. The smaller $\rho$ is, the faster the communication is. Note that since $W$ is doubly stochastic, the largest eigenvalue is always $1$. Let us look at a few examples to get some sense of the value of $\rho$:
\begin{align*}
W_1 =& {\1 \1^\top \over N}, && \rho = 0
\\
W_2 = & 
\left[\begin{matrix}
{1\over 3} & {1\over 3} & 0 & \cdots & 0 & {1\over 3} 
\\
{1 \over 3} & {1\over 3} & {1\over 3} & 0 & \cdots & 0
\\
0 & {1 \over 3} & {1\over 3} & {1\over 3} &  \cdots & 0
\\
\vdots & \vdots & \vdots & \vdots & \vdots & \vdots 
\\
0 & \cdots & 0 &  {1\over 3} & {1\over 3} & {1\over 3}
\\
{1\over 3} & 0 & \cdots & 0 &  {1\over 3} & {1\over 3} 
\end{matrix}\right]
&& \rho \approx 1- {16 \pi^2\over 3N^2}
\\
W_3 = & \left[\begin{matrix}
 \text{any doubly stochastic matrix} & {\bf 0}
 \\
 {\bf 0}^\top & 1
\end{matrix}\right] && \rho=1.
\end{align*}
$W_1$ (with $\rho=1$) corresponds to the fully connected network, which is the fastest network for spreading information; $W_2$ (with a $\rho$ value close to but smaller than $1$) corresponds to the ring network; $W_3$ (with $\rho=1$) corresponds to a disconnected network, which does work for $\DSGD$.

The last assumption we make is about the initial local models, which we assume to be identical. This assumption is not necessary (we can obtain a similar result without it) but can simplify notations and derivations in the proof.
\begin{tcolorbox}[colback=pink!5!white,colframe=black!75!black]
\begin{assumption} \label{ass:dsgd_3}
We make the following assumption for the algorithm initialization:
\begin{itemize}
\item The initial values for all workers are identical, that is,
\[
\x^{(1)}_1 = \x^{(2)}_1 = \cdots = \x^{(N)}_1.
\]
\end{itemize}
\end{assumption}
\end{tcolorbox}

\subsection{Convergence rate}
The theoretical analysis for $\DSGD$ is relatively complicated. Readers can jump directly to Theorem~\ref{thm:dsgd_1} if they are not interested in the convergence proof.

To simplify the notations used in our proof, let us define some new notations for convenience. 
Denote by 
\begin{align*}
f'\(\X_t\) := & \left[
f'_1\(\x^{(1)}_t\),
f'_2\(\x^{(2)}_t\),
\cdots,
f'_N\(\x^{(N)}_t\) 
\right]
\\
\bar{f}'(\X_t) := & f'(\X_t){\1 \over N},
\\
\bar{\x}_t := & \X_t{{\bf 1} \over N},
\\
\bar{\X}_t := & \X_t{{\bf 11}^\top \over N},
\\
\bar{\g}_t := & \G_t{{\bf 1} \over N}.
\end{align*} 
Then it is not hard to see that
\begin{align}
\E\left[\G\(\X_t; \{\xi_t^{(n)}\}_{n=1}^N\) \right] = f'\(\X_t\).
\end{align}

%=======
%A clear advantage of $\SGD$ is that the computational complexity reduces to $O(d)$ per iteration. It is worth pointing out that the $\SGD$ algorithm is NOT a descent algorithm due to its randomness.
%
%The next question is whether [it [what is ìitî here?]] converges and, if it does, how quickly. We first make some typical assumptions:
%\begin{tcolorbox}[colback=pink!5!white,colframe=black!75!black]
%\begin{assumption}
%>>>>>>> b94b8679f310fa566039184f25fe4da4b1ce56ff

We next show some preliminary results that illustrate our final proof:

\begin{tcolorbox}[colback=yellow!5!white,colframe=black!75!black]
\begin{lemma}\label{lem:asgd_0}
Under Assumption~\ref{ass:dsgd_1}, we have
\begin{align}
\left\|\bar{f'}(\X_t) - f'(\bar{\x}_t)\right\|^2 \leq & {L^2\over N} \|\X_t - \bar{\X}_t\|^2_F \label{eq:dsgd_proof_8}
\\
\left\| f'(\X_t) - f'(\bar{\x}_t){\bf 1}^\top \right\|^2_F \leq &  L^2 \|\X_t - \bar{\X}_t\|^2_F
+
N\varsigma^2
\\
\E\left[ \left\| \G_t\right\|^2\right] \leq & 2N\left\| f'(\bar{\x}_t) \right\|^2 + 
\\ & \nonumber
4L^2 \|\X_t - \bar{\X}_t\|^2_F + 4N\varsigma^2 + N\sigma^2
\end{align}
\end{lemma} 
\end{tcolorbox}
\begin{proof}
We first bound the difference between $\bar{f}'(\X_t)$ and $f'(\bar{\x}_t)$:
\begin{align*}
\left\|\bar{f'}(\X_t) - f'(\bar{\x}_t)\right\|^2 = & \left\| {1\over N}\sum_{n=1}^N f'_n(\x^{(n)}_t) -  {1\over N}\sum_{n=1}^N f'_{n}\( \bar{\x}_t\)\right\|^2
\\ = &
{1\over N^2}\left\| \sum_{n=1}^N f'_n(\x^{(n)}_t) -  \sum_{n=1}^N f'_{n}\( \bar{\x}_t\)\right\|^2
\\ \leq &
{1\over N}\sum_{n=1}^N\left\|  f'_n(\x^{(n)}_t) - f'_{n}\(\bar{\x}_t\)\right\|^2
\\ \leq &
{L^2\over N} \sum_{n=1}^N \|\x_t^{(n)} - \bar{\x}_t\|^2
\\ = &
{L^2\over N} \|\X_t - \bar{\X}_t\|^2_F 
\end{align*}
where the first inequality uses the property 
\[
\left\|\sum_{n=1}^N \z^{(n)} \right\|^2 \leq N \sum_{n=1}^N \left\|\z^{(n)}\right\|^2.
\]

Next, we prove the second inequality by using Assumption~\ref{ass:dsgd_1} (smoothness):
\begin{align*}
& {1\over 2}\left\| f'(\X_t) - f'(\bar{\x}_t){\bf 1}^\top \right\|^2_F 
\\ = &
{1\over 2}\left\| f'(\X_t) - f'(\bar{\X}_t) + f'(\bar{\X}_t) - f'(\bar{\x}_t){\bf 1}^\top \right\|^2_F
\\ \leq &
\left\| f'(\X_t) - f'(\bar{\X}_t) \right\|^2_F +
% \\ &
 \left\|f'(\bar{\X}_t) - f'(\bar{\x}_t){\bf 1}^\top \right\|^2_F
 \\ \leq &
 L^2 \|\X_t - \bar{\X}_t\|^2_F + 
 %\left\| f'(\bar{\X}_t) - \bar{f}'(\bar{\X}_t)\1^\top \right\|^2_F +
 %\\ &
 \left\|{f}'(\bar{\X}_t)- f'(\bar{\x}_t){\bf 1}^\top \right\|^2_F
 \\ = &
  L^2 \|\X_t - \bar{\X}_t\|^2_F + 
\sum_{n=1}^N \left\| f'_n(\bar{\x}_t) - f'(\bar{\x}_t) \right\|^2 
\\ \leq & 
  L^2 \|\X_t - \bar{\X}_t\|^2_F
+
N\varsigma^2.
% \\ \leq & 
%  L^2 \|\X_t - \bar{\X}_t\|^2_F + N\varsigma^2 +
% \left\| \bar{f}'(\bar{\X}_t)\1^\top - f'(\bar{\x}_t){\bf 1}^\top \right\|^2_F
\end{align*}

Next, we prove the last inequality:
\begin{align*}
\E\left[ \left\| \G_t\right\|^2\right] = & \left\| \E\left[ \G_t\right] \right\|^2_F + \E \left\| \G_t - \E\left[ \G_t\right] \right\|^2_F
\\ \leq &
\left\| f'(\X_t) \right\|^2_F + N\sigma^2
\\ \leq &
2\left\| f'(\X_t) - f'(\bar{\x}_t)\1^\top \right\|^2_F + 2\left\| f'(\bar{\x}_t)\1^\top \right\|^2_F + N\sigma^2
\\ = &
2\left\| f'(\X_t) - f'(\bar{\x}_t)\1^\top \right\|^2_F + 2N\left\| f'(\bar{\x}_t) \right\|^2 + N\sigma^2.
\end{align*}
%Together with the proved second inequality, we prove the last inequality [Please clarify this text. Wasnít the last inequality just proven above? What is happening with these two inequalities in the description below?]]:
%\begin{align*}
%& {1\over 3}\left\| f'(\X_t) - f'(\bar{\x}_t){\bf 1}^\top \right\|^2_F 
%\\ = &
%{1\over 3}\left\| f'(\X_t) - f'(\bar{\X}_t) + f'(\bar{\X}_t) - \bar{f}'(\bar{\X}_t)\1^\top + \bar{f}'(\bar{\X}_t)\1^\top - f'(\bar{\x}_t){\bf 1}^\top \right\|^2_F
%\\ \leq &
%\left\| f'(\X_t) - f'(\bar{\X}_t) \right\|^2_F +
%% \\ &
% \left\| f'(\bar{\X}_t) - \bar{f}'(\bar{\X}_t)\1^\top \right\|^2_F +
% \\ &
% \left\| \bar{f}'(\bar{\X}_t)\1^\top - f'(\bar{\x}_t){\bf 1}^\top \right\|^2_F
% \\ \leq &
% L^2 \|\X_t - \bar{\X}_t\|^2_F + \left\| f'(\bar{\X}_t) - \bar{f}'(\bar{\X}_t)\1^\top \right\|^2_F +
% %\\ &
% \left\| \bar{f}'(\bar{\X}_t)\1^\top - f'(\bar{\x}_t){\bf 1}^\top \right\|^2_F
%% \\ \leq & 
%%  L^2 \|\X_t - \bar{\X}_t\|^2_F + N\varsigma^2 +
%% \left\| \bar{f}'(\bar{\X}_t)\1^\top - f'(\bar{\x}_t){\bf 1}^\top \right\|^2_F
%\end{align*}
%\begin{align*}
%\left\| f'(\bar{\X}_t) - \bar{f}'(\bar{\X}_t)\1^\top \right\|^2_F = & \sum_{n=1}^N \left\| f'_n(\bar{\x}_t) - f'(\bar{\x}_t) \right\|^2 \leq N\varsigma^2
%\end{align*}
%\begin{align*}
%\left\| \bar{f}'(\bar{\X}_t)\1^\top - f'(\bar{\x}_t){\bf 1}^\top \right\|^2_F = \left\| x \right\|^2_F
%\end{align*}
\end{proof}

Next, we prove a useful inequality:

\begin{tcolorbox}[colback=yellow!5!white,colframe=black!75!black]
\begin{lemma}\label{lem:dsgd:seq}
Given two non-negative sequences $\{a_t\}_{t=1}^{\infty}$ and $\{b_t\}_{t=1}^{\infty}$ that satisfying
\begin{equation*}
a_t =  \sum_{s=1}^t\rho^{t-s}b_{s}, \numberthis \label{eqn1}
\end{equation*}
with $\rho\in[0,1)$, we have
\begin{align*}
\sum_{t=1}^{k}a_t^2 \leq & 
\frac{1}{(1-\rho)^2} \sum_{s=1}^kb_s^2.
\end{align*}
\end{lemma} 
\end{tcolorbox}

\begin{proof}
Consider the left-hand side:
\begin{align*}
\sum_{t=1}^{k}a_t^2  = & \sum_{t=1}^{k}\sum_{s=1}^t\rho^{t-s}b_{s}\sum_{r=1}
^t\rho^{t-r}b_{r} 
\\ = & 
\sum_{t=1}^{k}\sum_{s=1}^t\sum_{r=1}^t\rho^{2t-s-r}b_{s}b_{r} \nonumber
\\ \leq & 
\sum_{t=1}^{k}\sum_{s=1}^t\sum_{r=1}^t\rho^{2t-s-r}{b_{s}^2+b_{r}^2\over2} 
\\ = & 
\sum_{t=1}^{k}\sum_{s=1}^t\sum_{r=1}^t\rho^{2t-s-r}b_{s}^2 \nonumber
\\ \leq & 
{1\over 1-\rho}\sum_{t=1}^{k}\sum_{s=1}^t\rho^{t-s}b_{s}^2 
\\ \leq & {1\over (1-\rho)^2}\sum_{s=1}^{k}b_{s}^2.
\end{align*}
This completes the proof.
\end{proof}

Given these preliminary results, we can show the first inequality for the upper bound of the total model variation, that is, the distance between the local models $\X_t$ and the average model $\bar{\X}_t$. In attempting to understand the decentralized algorithm, the first question people may ask is whether all local models will achieve consensus eventually as the centralized counterpart always satisfies the consensus over all iterations. Lemma~\ref{lem:asgd_1} essentially shows that $\DSGD$ can ensure that all $\x_t^{(n)}$s converge to $\bar{\x}_t$. This happens because the total variation $\sum_{t=1}^T\E \left[\|\bar{\X}_t - \X_t\|^2_F\right]$ is roughly bounded by $O(\gamma^2T)$. Based on our experiences of analyzing stochastic algorithms, we know that the learning rate is usually proportional to $1/\sqrt{T}$. Therefore, the total variation is bounded by a constant, which indicates that all $\x_t^{(n)}$ converges to $\bar{\x}_t$. 

\begin{tcolorbox}[colback=yellow!5!white,colframe=black!75!black]
\begin{lemma}\label{lem:dsgd_1}
Under Assumptions~\ref{ass:dsgd_2} and \ref{ass:dsgd_3}, we have
%\begin{align*}
%\E \left[\|\bar{\X}_t - \X_t\|^2_F\right] \leq \frac{\gamma^2 }{1-\rho}  \E\left[ \|\G_t\|^2 \right]
%\end{align*}
%{\rc Should this be??:
\begin{align*}
\sum_{t=1}^T\E \left[\|\bar{\X}_t - \X_t\|^2_F\right] \leq \sum_{t=1}^T\frac{\gamma^2\rho^2 }{(1-\rho)^2}  \E\left[ \|\G_t\|^2 \right].
\end{align*}
\end{lemma} 
\end{tcolorbox}

\begin{proof}
From the updating rule, we obtain the following closed form for $\X_t$ and $\bar{\X}_t$:
\begin{align*}
\X_t = & \gamma \sum_{s=1}^{t-1} \G_s W^{t-s}
\\
\bar{\X}_t = & \X_t {{\bf 1 1}^{\top} \over N}
\end{align*}
Next, we bound the difference between $\bar{\X}_t$ and $\X_t$ as follows:
\begin{align*}
& {1\over \gamma}\left\| \X_t - \bar{\X}_t \right\|_F
\\ = &
{1\over \gamma}\left\| \X_t \(I -  {{\bf 1 1}^{\top} \over N}\) \right\|_F
\\ = &
\left\| \sum_{s=1}^{t-1} \G_s W^{t-s} \(I -  {{\bf 1 1}^{\top} \over N}\) \right\|_F
\\ = &
\left\| \sum_{s=1}^{t-1} \G_s  \(W^{t-s} -  {{\bf 1 1}^{\top} \over N}\) \right\|_F \quad \(\text{due to}~W{\bf 1} = {\bf 1}\) 
\\ \leq &
 \sum_{s=1}^{t-1} \left\| \G_s  \(W^{t-s} -  {{\bf 1 1}^{\top} \over N}\) \right\|_F \quad \(\text{due to the triangle inequality}\)
 \\ \leq &
  \sum_{s=1}^{t-1} \left\| \G_s \right\|_F \left\|\(W^{t-s} -  {{\bf 1 1}^{\top} \over N}\) \right\| \quad \(\text{due to $\|AB\|_F \leq \|A\|_F\|B\|$}\) 
  \\ \leq &
  \sum_{s=1}^{t-1} \rho^{t-s} \left\| \G_s \right\|_F.
%  \\ \leq &
%  \sqrt{\sum_{s=1}^{t-1} \rho^{\rc 2(t-s-1)}} \sqrt{\sum_{s=1}^{t-1} \| \G_s\|_F^2}
%  \\ \leq &
%  \sqrt{(1-\rho)}  \sqrt{\sum_{s=1}^{t-1} \| \G_s\|_F^2}.
%  \\ \leq & 
%  \frac{1}{\sqrt{(1-{\rc\rho^2})}}\sqrt{\sum_{s=1}^{t-1} \| \G_s\|_F^2}
\end{align*}
Taking the square and the sum over $t$, the equation above becomes
\begin{align*}
&\sum_{t=1}^T\frac{1}{\gamma^2}\left\| \X_t - \bar{\X}_t \right\|_F^2\\
\leq & \sum_{t=1}^T\rho^2\left( \sum_{s=1}^{t-1} \rho^{t-s-1} \left\| \G_s \right\|_F \right)^2\\
\leq & \frac{\rho^2}{(1-\rho)^2}\sum_{t=1}^T\left\|\G_t\right\|_F^2\quad\text{(due to Lemma~\ref{lem:dsgd:seq})}.
\end{align*}
This completes the proof.
\end{proof}

%One important observation is that the total variation is roughly bounded by $O(\gamma^2T)$. Based on our experiences of analyzing stochastic algorithms, we know that the learning rate is usually proportional to $1/\sqrt{T}$. Therefore, the total variation is bounded by a constant, which indicates that all $\x_t^{(n)}$ converges to $\bar{\x}_t^{(n)}$. This result roughly solves our concern for the decentralized algorithm since the decentralized algorithm does not take the full average of all local models. 

\begin{tcolorbox}[colback=yellow!5!white,colframe=black!75!black]
\begin{lemma}\label{lem:asgd_1}
Under Assumptions~\ref{ass:gd}, \ref{ass:sgd}, and \ref{ass:asgd}, if the learning rate is chosen to satisfy 
\[
\gamma\rho \leq \frac{{1-\rho}}{4L}
\] 
we have
%\begin{align*}
%\E \left[\|\bar{\X}_t - \X_t\|^2_F\right] \leq \frac{8N(\sigma^2 + \varsigma^2)\gamma^2 }{1-\rho} + \frac{2N\gamma^2}{1-\rho} \E\left[ \|f'(\bar{\x}_t)\|^2 \right].
%\end{align*}
%{Should this be?: 
\begin{align*}
\sum_{t=1}^T\E \left[\|\bar{\X}_t - \X_t\|^2_F\right] \leq & \sum_{t=1}^T\frac{8N(\sigma^2 + 
\varsigma^2)\gamma^2\rho^2 }{(1-\rho)^2} +
\\ &
\sum_{t=1}^T\frac{2N\gamma^2\rho^2}{(1-\rho)^2} \E\left[ \|f'(\bar{\x}_t)\|^2 \right].
\end{align*}
\end{lemma} 
\end{tcolorbox}
\begin{proof}
Applying Lemma~\ref{lem:asgd_0} to Lemma~\ref{lem:dsgd_1} gives
\begin{align*}
\frac{(1-\rho)^2}{\gamma^2\rho^2}\E \left[\|\bar{\X}_t - \X_t\|^2_F\right] \leq & 2N \E\left[ \|f'(\bar{\x}_t)\|^2\right] + 
\\ &
4L^2 \E\left\|\bar{\X}_t - {\X}_t\right\|^2_F + 4N\varsigma^2 + N\sigma^2.
\end{align*}
Given the restriction on the learning rate $\gamma$, we have $4L^2 \leq (1-\rho)^2/(2\gamma^2\rho^2)$ and thus obtain
\begin{align*}
\frac{(1-\rho)^2}{\gamma^2\rho^2}\E \left[\|\bar{\X}_t - \X_t\|^2_F\right] \leq & 2N \E\left[ \|f'(\bar{\x}_t)\|^2\right] + 
\\ &
+ 8N(\varsigma^2 + \sigma^2),
\end{align*}
which completes the proof.
\end{proof}

\begin{tcolorbox}[colback=yellow!5!white,colframe=black!75!black]
\begin{lemma}\label{lem:asgd_5}
Under Assumption~\ref{ass:dsgd_1}, if $\gamma \leq {1\over L}$, we have
\begin{align*}
\E\left[\|f'(\bar{\x}_t)\|^2\right] \leq & {2\over \gamma}\E\left[f(\bar{\x}_t)\right] - \E\left[f(\bar{\x}_{t+1})\right] + 
\\ &
{L^2 \over N} \E\left[\|\bar{\X}_t - \X_t\|^2_F\right] + {L\gamma \sigma^2\over N}.
\end{align*}
\end{lemma} 
\end{tcolorbox}
\begin{proof}
Let us start with a basic property:
\begin{align}
\E[\bar{\g}_t] = {1\over N}\sum_{n=1}^N f'_n(\x^{(n)}_t)=:\bar{f'}(\X_t)
\label{eq:dsgd_proof1}
\end{align}
We consider the improvement of $\E\left[f(\bar{\x}_{t+1})\right]$ over $\E \left[ f(\bar{\x}_t)\right]$ in the following:
\begin{align*}
&
\E\left[f(\bar{\x}_{t+1})\right] - \E \left[ f(\bar{\x}_t)\right] 
\\ \leq &
\E \langle f'(\bar{\x}_t), -\gamma  \bar{\g}_t \rangle + {L \gamma^2 \over 2} \E \left[\|\bar{\g}_t\|^2 \right] 
\\ = &
-\gamma \E\langle f'(\bar{\x}_t), \bar{f'}(\X_t) \rangle + {L\gamma^2 \over 2} \E\left[\|\bar{\g_t}\|^2\right]
\\ = &
-{\gamma\over 2} \E\|f'(\bar{\x}_t)\|^2 - {\gamma\over 2} \E\|\bar{f'}({\X}_t)\|^2 + {\gamma \over 2}\E \|\bar{f'}(\X_t)-f'(\bar{\x}_t) \|^2 +
\\ &
+ {L \gamma^2 \over 2} \E \left[\|\bar{\g}_t\|^2 \right]. \numberthis \label{eq:dsgd_proof_6}
\end{align*}
We look at the last term first in the above inequality:
\begin{align*}
\E \left[ \|\bar{\g}_t\|^2 \right] =& 
\left\|\E\left[\bar{\g}_t\right] \right\|^2 + 
\E \left[ \left\| \bar{\g}_t - \E\left[\bar{\g}_t \right]\right\|^2 \right]
\\ \leq &
\|\bar{f'}(\X_t)\|^2 + {\sigma^2\over N} \numberthis \label{eq:dsgd_proof_7}
\end{align*}
where the inequality uses the property of $\bar{\g}_t$ in \eqref{eq:dsgd_proof1} and the property for any i.i.d. random variables $z^{(1)}, \cdots, z^{(n)}$
\[
{\bf Var}\({1\over N}\sum_{n=1}^N z^{(n)}\) = {1\over N}{\bf Var}(z^{(n)}), 
\]
together with the boundedness in Assumption~\ref{ass:dsgd_1}.

Plugging \eqref{eq:dsgd_proof_7} and \eqref{eq:dsgd_proof_8} into \eqref{eq:dsgd_proof_6} yields 
\begin{align*}
&
\E\left[f(\bar{\x}_{t+1})\right] - \E \left[ f(\bar{\x}_t)\right] 
\\ \leq &
-{\gamma\over 2} \E\|f'(\bar{\x}_t)\|^2 - \({\gamma\over 2} - {L\gamma^2\over 2} \) \E\|\bar{f'}({\X}_t)\|^2 + 
\\ &
{L^2 \gamma \over 2N}\E \left\| {\X}_t- \bar{\X}_t \right\|^2_F
+ {L \gamma^2 \sigma^2 \over 2N} 
\\ \leq &
-{\gamma\over 2} \E\|f'(\bar{\x}_t)\|^2 +
{L^2 \gamma \over 2N}\E \left\| {\X}_t- \bar{\X}_t \right\|^2_F
+ {L \gamma^2 \sigma^2 \over 2N}. 
\end{align*}
This completes the proof.
\end{proof}

Now we are ready to show the main result.

\begin{tcolorbox}[colback=blue!5!white,colframe=black!75!black]
\begin{theorem} \label{thm:dsgd_1}
Under Assumptions~\ref{ass:dsgd_1} and \ref{ass:dsgd_2}, we choose the learning rate to be
\begin{align*}
\gamma = \(1+\sqrt{TN}\sigma + T^{1/3}\varsigma^{2/3}\rho^{2/3}(1-\rho)^{-2/3}\)^{-1}.
\end{align*}
If $T$ is sufficiently large such that the learning rate satisfies the following condition
\begin{align}
\gamma \leq \min\left\{\frac{1-\rho}{4L},~\frac{(1-\rho)^2N}{L} \right\},
\label{eq:DSGD_lr_cond}
\end{align}
then the $\DSGD$ algorithm admits the following convergence rate:
\begin{align*}
{1\over T} \sum_{t=1}^T \E \left\| f'(\bar{\x}_t) \right\|^2 \lesssim{1\over T} + {\sigma\over \sqrt{NT}} + \(\frac{\varsigma\rho}{ T(1-\rho)}\)^{2/3}.
\end{align*}
\end{theorem}
\end{tcolorbox}
\begin{proof}
Taking the summarization over $t$ from $t=1$ to $t=T$ for Lemma~\ref{lem:asgd_5} yields
\begin{align*}
& {1\over T} \sum_{t=1}^T \E \left\| f'(\bar{\x}_t) \right\|^2 
\\ \leq & {2\over T\gamma} \(f(\bar{\x}_0) - f^\star\) + \frac{L\sigma^2 \gamma}{N} + {L^2\over NT} \sum_{t=1}^T \left\|\bar{\X}_t - \X_t\right\|^2_F 
\\ \leq &
{2\over T\gamma} \(f(\bar{\x}_0) - f^\star\) + \frac{L\sigma^2 \gamma}{N} +
\\ &
\frac{8L^2\varsigma^2\gamma^2 }{(1-\rho)^2} 
+ \frac{8L^2\sigma^2\gamma^2\rho^2 }{(1-\rho)^2} 
+ \frac{2L^2\gamma^2\rho^2}{(1-\rho)^2T}\sum_{t=1}^T\E \|f'(\bar{\x}_t)\|^2.
\end{align*}
From the restriction on the learning rate, we have
\[
\frac{8L^2\sigma^2\gamma^2\rho^2 }{(1-\rho)^2}  \leq \frac{L\sigma^2 \gamma}{N}
\]
and
\[
{2L^2\gamma^2\rho^2 \over (1-\rho)^2} \leq {1\over 2}.
\]
It follows that
\begin{align*}
&{1\over 2T} \sum_{t=1}^T \E \left\| f'(\bar{\x}_t) \right\|^2
\\ \leq & {2\over T\gamma} \(f(\bar{\x}_0) - f^\star\) + \frac{L\sigma^2 \gamma}{N} + {L^2\over NT} \sum_{t=1}^T \left\|\bar{\X}_t - \X_t\right\|^2_F 
\\ \leq &
{2\over T\gamma} \(f(\bar{\x}_0) - f^\star\) + \frac{2L\sigma^2 \gamma}{N} +
\frac{8L^2\varsigma^2\gamma^2\rho^2 }{(1-\rho)^2} 
\end{align*}
For simplicity, we treat $f(\bar{\x}_0) - f^\star$ and $L$ as constants and obtain the following simplified inequality:
\begin{align*}
{1\over T} \sum_{t=1}^T \E \left\| f'(\bar{\x}_t) \right\|^2 \lesssim & {1\over T\gamma}  + \frac{\sigma^2 \gamma}{N} + 
\frac{\varsigma^2\gamma^2\rho^2 }{(1-\rho)^2}.
\end{align*} 
Due to the form of the learning rate in \eqref{eq:DSGD_lr_cond}, we have
\begin{align*}
\frac{1}{T\gamma} \leq & {1\over T} + {\sigma\over \sqrt{NT}} + {\varsigma^{2/3} \over T^{2/3}(1-\rho)^{2/3}}
\\
\frac{\sigma^2\gamma}{N} \leq & {\sigma \over \sqrt{NT}}
\\
\frac{\varsigma^2\gamma^2\rho^2}{1-\rho} \leq & {\varsigma^{2/3}\rho^{2/3} \over T^{2/3}(1-\rho)^{2/3}},
\end{align*}
which completes the proof.
\end{proof}
We highlight the following observations from this convergence analysis:
\begin{itemize}
    \item ({\bf Consistency to $\mSGD$}) If we use the fully connected network, then $\rho = 0$ and the $\DSGD$ algorithm reduces the (centralized) $\mSGD$ algorithm. We can see that the convergence rate in Theorem \eqref{thm:dsgd_1} is consistent with $\mSGD$, since the last term becomes zero.
    \item ({\bf Linear speedup}) The additional term in the rate is the the last term, comparing to $\mSGD$. It is caused by using decentralized updating. As long as $T$ is sufficiently large, then the last term will be dominated by the second term. As a result, the linear speedup can be achieved. 
\end{itemize}

\chapter{Further Readings} 
\label{sec:further_readings}

Developing efficient distributed
learning systems is an emerging topic that
has received intensive interests in recent
years. The goal of this paper is by no
means to provide a complete summary of the
recent development in this area --- instead,
our goal is merely to provide an
``overly simplified'' overview. 

In this \part, we assemble a {\em best-effort} reading 
list to provide readers pointers for
further reading. This list is incomplete --- it is
more like a set of ``{\em pointers to pointers}'' whose
goal is to provide a bird-eye view on 
the trend of research and to provide 
a starting point for readers to start their
own navigation. To assemble this list, we went through {\em most} papers published in ICML, N(eur)IPS, VLDB, SIGMOD, SysML, SOSP and OSDI since 2015 (up to 2019), 
and {\em tried our best} to summarize relevant papers into multiple categories.  
This selection method means that this list 
inevitably misses many early seminal work
on this topic. However, we believe that the  
union of 
all papers cite by papers in this list
should provide a reasonable coverage.

\section{Communication and Data Compression}

Data movement during training can be one of the largest
system bottleneck, especially
when there are many workers in the system or the computation
device is significantly more powerful than the peak throughput
of data movements. One collection of work focus on 
optimizing data movements via compression, and popular compression
strategy includes quantization, sparsification, sketching,
and other noise-corrupted transformation. Some examples of
recent work in this direction include \citet{Acharya2019-zk,
Zhu2018-xf,
Wu2018-xt,
Cohen2018-bl,
Bernstein2018-mv,
Zhang2017-yx,
Chen2017-mo,
Zhang2016-hi,
Gupta2015-og,
Zhu2015-ua,
Jiang2018-zp,
Elgohary2016-gf,
Kara2018-oc,
Wang2019-lb,
Lim2019-hk,
Wang2018-aa,
Wang2018-sl,
Agarwal2018-hg,
Alistarh2018-ct,
Banner2018-yh,
Yu2018-io,
Stich2018-sa,
Jiang2018-bn,
Alistarh2017-yh,
Wen2017-ik,
De_Sa2015-hu}.

\section{Decentralization and Approximate Averaging}

Another line of work focuses on the scenario in which
calculating the exact average among all workers is difficult
in a single round of communication. One example is in
peer-to-peer networks in which each worker can only
communicate with its neighbor. One collection of work
focus on analyzing the system behavior and designing
novel algorithms in such a scenario. Some examples
of recent work in this direction include \citet{
sun2020improving,
hsieh2020non,
lu2020moniqua,
richards2020decentralised,
koloskova2020unified,
Yu2019-xd,
Assran2019-iq,
li2019decentralized,
li2019linear,
Tang2018-fl,
li2017decentralized,
He2018-fu,
Lian2017-ni,
li2017primal,
simonetto2017decentralized,
Colin2016-xj,
yuan2016convergence,
shi2015extra,
shi2015proximal,
ling2013decentralized,
nedic2009distributed}.
It is also worth pointing out that the researchers in federated learning also borrow the idea of decentralization for the data privacy purpose \citep{he2019central}.

There are also work which try to combine both
communication compression and decentralization, for example
\citet{
beznosikov2020biased,
taheri2020quantized,
tang2019texttt,
Koloskova2019-kl,
Tang2018-ec}.

\section{Asynchronous Communication}

The synchronization barrier among all workers is often 
a system bottleneck, especially when there are many workers
or there are stragglers (i.e., some workers are slower
than other workers). One collection of work focus on
removing the synchronization barrier by allowing the
workers to proceed in an {\em asynchronous} fashion.
Some examples of recent work in this direction include
\citet{Zhou2018-fw,
Simsekli2018-bb,
Nguyen2018-vi,
Gu2018-tk,
lian2016comprehensive,
Zheng2017-au,
Peng2017-kv,
Aybat2015-gk,
Hsieh2015-ny,
Jiang2017-uu,
Wangni2018-ux,
Sun2017-zf,
You2016-pj,
Chen2016-za,
Pan2016-py,
Lian2015-vz,
Chaturapruek2015-vt,
LiuWright14siopt,
Liu14icml,
Sridhar2013nips,
AgarwalD12,
Hogwild11nips}.

There are also work which try to combine both decentralization
and asynchronous communication, for example \citep{Lian2019-ig}.

\section{Optimizing for Communication Rounds}

There is a collection of work that tries to optimize for 
the number of communication rounds during training --- instead
of communicating every iteration, these methods allow 
each worker to run longer {\em locally} for multiple
iterations. Some examples of recent work in this direction
include \citet{
Garber2017-pn,
Ma2015-js,
Wang2017-ua,
Kamp2017-iq,
Zhang2015-og}.

\section{System Optimization and Automatic Tradeoff Management}

On the system side, one line of work is to
further optimize the system communication 
primitives and communication strategies 
to take advantage of the property
of the underlying ML workload. Some examples of recent work in this direction include
\citep{Hashemi2019-hq,
Jayarajan2019-tf,
Cho2019-ko,
Jia2019-hu,
Wang2018-nq}.
Another line of work tries to automatically 
optimize in the tradeoff introduced by
these system relaxation techniques (e.g.,
the communication frequency which is often
a hyperparameter). Some examples of recent work in this direction include \citep{Kaoudi2017-gf,
Xin2018-vl,
Mahajan2018-ad,
Wang2019-gp,
Li2018-gy,
Dunner2018-gy}.

\section{Other Topics}

One line of work focuses on designing variance reduction
techniques for stochastic first-order methods, e.g.,
\citet{
Ji2019-ea,
Horvath2019-ll,
Zou2018-pw,
Zhou2018-rl,
Zhou2018-sy,
Hazan2016-hp,
Li2016-ry,
Allen-Zhu2016-ss,
Allen-Zhu2016-ne,
Reddi2016-wp,
Jothimurugesan2018-uh,
Zhou2018-le,
Liu2018-jq,
Hanzely2018-ud,
Arjevani2017-fw,
Bietti2017-ny,
Palaniappan2016-wi,
Hofmann2015-nb}. One interesting way of achieving
this is to change the sample distribution during training
(i.e., sample more informative data points more frequently
than non-informative data points), e.g.,
\citet{
Katharopoulos2018-pk,
Namkoong2017-ns,
Gopal2016-af,
Zhao2015-ev,
Johnson2018-da,
Cutkosky2018-yj,
Zhu2016-pg,
Qu2015-wp}.

Another line of work focuses on designing distributed 
learning algorithms that is robust to failures, e.g.,
by making the system Byzantine-resilient or 
tolerant to stragglers via Gradient coding.
Examples of recent work in this direction include
\citet{
Xie2019-wq,
Yin2018-qv,
Damaskinos2018-ti,
Chen2018-lu,
Ye2018-mf,
Raviv2018-sx,
Tandon2017-zt,
Alistarh2018-sc,
Karakus2017-sw,
Blanchard2017-lw}.

Many, if not most, theoretical analysis of system
relaxations of distributed learning systems assumes
that the system samples data points with replacement.
However, this is different from how most learning
systems are implemented in practice. There have been
efforts in trying to close this gap by analyzing different
strategies of scan order. Examples of recent work in
this direction include
\citep{
Nagaraj2019-ur,
Haochen2019-dg,
Shamir2016-fe}.

{\rv 
In this work, we mainly focus on techniques for distributed
learning that optimize for the communication among workers.
However, there are other, orthogonal directions, in
optimizing for the performance and scalablity of
distributed learning systems. For example,
when one focuses on distributed learning on the edge~\citep{DBLP:journals/corr/abs-1803-04311}
or in geo-distributed setting~\citep{201564},
additional considerations are often 
necessary. Another line of work tries to further 
take advantage
of specific structures of the given task, e.g.,
deep neural networks. One example is a very
interesting line of work that uses large batch size
for training deep learning models~\citep{DBLP:journals/corr/GoyalDGNWKTJH17,ghadimi2013minibatch,DBLP:journals/corr/abs-1904-00962}.
For more techniques in this direction, we 
refer the reader to a comprehensive survey paper~\citep{DBLP:journals/corr/abs-1802-09941} on this 
topic.
}

%Another related line of work is {\em federated learning}.

%\textcolor{red}{SAY SOMETHING ABOUT FEDRATED LEARNING}

\section*{Acknowledgment}
We thank Hanlin Tang for providing a neater proof for the convergence of $\DSGD$ which is used in this book. We also thank
Shaoduo Gan, Jiawei Jiang, and Binhang Yuan for adding latest citations (after 2020) in Chapter 6.

\backmatter  
\printbibliography

\end{document}